\documentclass[sigconf]{acmart}

\usepackage{enumitem}
\usepackage{multirow}
\usepackage{soul}
\usepackage{xcolor}
\usepackage{makecell}

\AtBeginDocument{%
  }

\copyrightyear{2026}
\acmYear{2026}
\setcopyright{cc}
\setcctype{by}
\acmConference[IUI '26]{31st International Conference on Intelligent User Interfaces}{March 23--26, 2026}{Paphos, Cyprus}
\acmBooktitle{31st International Conference on Intelligent User Interfaces (IUI '26), March 23--26, 2026, Paphos, Cyprus}
\acmPrice{}
\acmDOI{10.1145/3742413.3789147}
\acmISBN{979-8-4007-1984-4/2026/03}

\usepackage{soul}
\soulregister\cite7 
\soulregister\ref7
\soulregister\todo7



\newcommand{\todo}[1]{{\color{red} [#1]}}

\definecolor{revision_color}{HTML}{ACE8FF}

\acmSubmissionID{5230}

\usepackage{svg}
\usepackage{bbm}
\begin{document}

\title{Rules or Weights? Comparing User Understanding of Explainable AI Techniques with the Cognitive XAI-Adaptive Model}

\author{Louth Bin Rawshan}
\affiliation{%
  \department{Department of Computer Science}
  \institution{National University of Singapore}
  \city{Singapore}
  \country{Singapore}
}
\email{e0546303@u.nus.edu}

\author{Zhuoyu Wang}
\affiliation{%
  \department{Department of Computer Science}
  \institution{National University of Singapore}
  \city{Singapore}
  \country{Singapore}
}
\email{wang.zhuoyu@u.nus.edu}

\author{Brian Y Lim}
\authornote{Corresponding author}
\affiliation{%
  \department{Department of Computer Science}
  \institution{National University of Singapore}
  \city{Singapore}
  \country{Singapore}
}
\email{brianlim@nus.edu.sg}

\renewcommand{\shortauthors}{Rawshan, Wang, and Lim}

\begin{abstract}
Rules and Weights are popular XAI techniques for explaining AI decisions.
Yet, it remains unclear how to choose between them, lacking a cognitive framework to compare their interpretability. 
In an elicitation user study on forward and counterfactual decision tasks, we identified 7 reasoning strategies of interpreting three XAI Schemas—weights, rules, and their hybrid. 
To analyze their capabilities, we propose CoXAM, a Cognitive XAI-Adaptive Model with shared memory representation to encode instance attributes, linear weights, and decision rules.
CoXAM employs computational rationality to choose among reasoning processes based on the trade-off in utility and reasoning time, separately for forward or counterfactual decision tasks.
In a validation study, CoXAM demonstrated a stronger alignment with human decision-making compared to baseline machine learning proxy models.
The model successfully replicated and explained several key empirical findings, including that counterfactual tasks are inherently harder than forward tasks, decision tree rules are harder to recall and apply than linear weights, and the helpfulness of XAI depends on the application data context, alongside identifying which underlying reasoning strategies were most effective.
With CoXAM, we contribute a cognitive basis to accelerate debugging and benchmarking disparate XAI techniques.
\end{abstract}



\maketitle

\section{Introduction}

AI systems increasingly influence high-stakes decisions in finance, law, healthcare, and criminal justice, making explainable AI (XAI) essential for exposing a model’s decision process to end users. 
Although numerous XAI techniques have been developed~\cite{mersha2024explainable, schwalbe2024comprehensive}, popular explanations take the form of linear models~\cite{ribeiro2016should, lundberg2017unified, poursabzi2021manipulating} or rules~\cite{breiman1984cart, ribeiro2018anchors, lakkaraju2016interpretable}.
Despite their simplicity, these methods remain effective in critical domains such as healthcare~\cite{caruana2015intelligible} and finance~\cite{yeo2025comprehensive, caterson2024application}.
Perhaps this is because these structures align well with human reasoning schemas of factor weighting~\cite{memelink2013intentional}, and rule-based categorization~\cite{goodman2008rational}.
Several user studies have compared the effectiveness of these explanation methods, but a consensus remains elusive, reporting conflicting results that linear models~\cite{hase2020evaluating} or rules~\cite{ribeiro2018anchors} are more helpful.

We argue that a deeper examination of human cognition can help to elucidate when and why different XAI Schemas may be useful or misused.
We draw inspiration from the successes of cognitive modeling to understand people and improve user experience and performance.
Applications include: 
intelligent tutoring systems~\cite{corbett1994knowledge,koedinger1997intelligent}, programming tutors that infer novice error patterns~\cite{piech2015learning,rivers2017automated} and touchscreen typing~\cite{jokinen2021touchscreen,shi2024crtypist}. 
We argue that cognitive modeling of how users understand XAI can help developers to analyze the users' reasoning strategies, and ultimately serve as a platform for rapid prototyping and early evaluation of XAI on simulated users.

In this work, we focus on tabular data, where attributes are structured and semantically meaningful. 
Tabular data underpins many real-world applications in healthcare, finance, and operations, and remains a core setting for machine learning in practice~\cite{caruana2015intelligible,borisov2022tabular,grinsztajn2022tree}.
We use global explanations instead of instance-based explanations, since people use the same rules globally rather different rules per instance~\cite{bruner1956study, shepard1961learning, gigerenzer1996fast, maddox2004rulebased}.
In this context, we study how users interpret two XAI Schemas---weight-based factors (Weights) and rule-based decision trees (Rules)---to understand the AI model’s decisions via forward and counterfactual simulation tasks\footnote{These tasks assess whether participants can anticipate an AI prediction (forward) and identify which attribute to change to reach an alternative target prediction (counterfactual)~\cite{doshi2017towards}.}. From a formative user study with 24 participants, we elicited reasoning strategies,  
three for forward simulation (approximate calculation (Weights), feature attribution (Weights), attribute-threshold-attribute traversal (Rules)), and
four for counterfactual simulation (inverse calculation (Weights), inverse feature attribution (Weights), root/leaf node threshold crossing (Rules), availability heuristic (any)).

We propose \textbf{CoXAM}, a Cognitive XAI-Adaptive Model to represent a user who can learn and use linear-factor and rule-based explanations for AI simulation tasks.
It is XAI-Adaptive by using a shared memory framework to encode attribute values, linear factors and rule nodes as chunks.
Reasoning strategies for forward simulation were implemented with drift-diffusion decision dynamics~\cite{ratcliff1993methods, ratcliff2016diffusion} to model the decision probability and execution time, explicit heuristics determined from user reasoning steps, and several cognitive parameters to model human bounded rationality (e.g., diffusion noise, retrieval threshold).
Assuming that each user could use any reasoning strategy, CoXAM chooses the "best" strategy with computational rationality~\cite{oulasvirta2022computational}, based on the trade-off between the utility of the decision likely being correct and decision speed.

We conducted a summative user study, across two application scenarios (Wine Quality, Mushroom Edibility), with 340 participants performing forward and counterfactual simulation tasks with Weights explanation, Rules explanation, or their Hybrid\footnote{We further introduce a Hybrid XAI condition that interleaves Weights and Rules  explanations to test users’ ability to integrate or switch between reasoning styles.} over 27600 trials.
We evaluated CoXAM with the study data.
We found that Weights explanation was most helpful for the Wine Quality scenario which had linear attributes, and Rules explanations for Mushroom which was more nonlinear.
Unlike Weights explanation, Rules explanation was harder to recall than read, resulting in significantly lower forward simulation accuracy.
User accuracy on counterfactual tasks was worse than on forward tasks, due to the inverse reasoning needed.
These results were mostly replicated by CoXAM, which had the best fit to human decision responses compared to baseline proxies (KNN, Decision Tree, Linear Regression, SHAP) with the lowest NLL and BIC scores.
Deeper analysis of reasoning strategies found that approximate calculation performed worse than feature attribution for forward tasks when values were larger, and availability heuristic was least effective for counterfactual tasks.

Our \textbf{contributions} are:
\begin{enumerate}[topsep=0pt]
    \item Elicitation of user reasoning strategies for Rules and Weights explanations.
    \item Evaluation of their (in)effectiveness for forward and counterfactual simulation tasks.
    \item XAI-adaptive cognitive model of different XAI schemas for different decision tasks.
\end{enumerate}
CoXAM serves as a testbed for user behavior in tabular settings, allowing comparisons of Rule vs. Weights explanations while identifying the strategies users rely on. We include discussions on extensions to broader XAI families and tasks. 

\section{Related Work} 

Our main focus is to investigate and model human reasoning in Rule-based and Weight-based XAI. We explore related work that evaluate and contrast rule- and weight-based XAI Schemas, examine user modeling in XAI domains, and draw upon cognitive modeling in more general domains.

\subsection{Rules vs. Weights: A Core Debate}

A central question in tabular XAI is whether \emph{rule} explanations (e.g., decision trees, decision sets, anchors) or \emph{weight} explanations (e.g., linear models, local attributions) better support users. This debate predates XAI: classic judgment research documents strong performance for additive cue combination—including simple or equal-weight schemes—relative to informal rule formulations~\cite{meehl1954clinical, dawes1979robust, einhorn1975unit, dawes1974linear}. In parallel, work on compact, condition-based decision procedures showed that short, explicitly stated rules can rival regression under limited time or information~\cite{gigerenzer1996reasoning,czerlinski1999good}. Modern XAI inherits this split and yields mixed head-to-head findings: Anchors (rules) improved precision and reduced effort in some forward simulation tasks~\cite{ribeiro2018anchors}, whereas broader studies reported LIME (weights) improving simulatability over rule-based alternatives~\cite{hase2020evaluating}. For counterfactual prediction, one study found no clear advantage for either family~\cite{hase2020evaluating} while another observed benefits for linear/weight information in a text-editing setting~\cite{arora2022explain}. A large-$N$ comparison reported decision trees (rules) as more locally interpretable than logistic regression (weights) under their task representation~\cite{slack2019assessing}. 

Overall, the literature does not support a universal winner. Outcomes appear sensitive to task framing (forward vs.\ counterfactual) and dataset properties. This background motivates examining, within concrete tasks, \emph{what information each form conveys} and \emph{how users apply that information} in different tasks.

\subsection{Modeling User Performance on XAI}
Several works have modeled human decisions in XAI, but typically with the aim of capturing behavioral trends, often neglecting underlying cognitive processes. This behavioral modeling is commonly achieved by training supervised machine learning models with user decision data to simulate human-like responses~\cite{wang2022will, virgolin2020formulaOfInterpretability,lage2018human, hilgard2021learningByHumansForHuman, reichman2024machine}. Recent works have further differentiated these modeling techniques: Ma et al.~\cite{ma2023should} trained a personalized and refinable decision tree based on user choices, while Mozannar et al.~\cite{mozannar2022teach} utilized user feedback to model human rejection of AI erroneous suggestions. Conversely, Chen et al. \cite{chen2022use} demonstrated that a decision proxy trained exclusively on AI explanations, without direct human input or labels, could still achieve reasonable alignment with human decisions. Fiori et al.~\cite{fiori2024using} describe that human understanding of natural language explanations could be loosely approximated by using large-language models as a proxy for the human.

Unlike prior work that focuses on matching behavioral data, our work seeks to model the underlying human cognitive processes in consuming XAI. This would enable deeper insights into which strategies users adopt and how their reasoning is affected by XAI schema. 
We focus on AI-assisted decision making~\cite{ma2023should} where the user decides whether to delegate decisions to the AI or decide for themselves, instead of human-AI collaboration~\cite{lai2022human, vaccaro2024combinations} where the AI complements the user's knowledge to support effective teaming.

\subsection{Cognitive Modeling for Human Behavior}

Cognitive modeling has long been used to understand and predict human decision-making across domains such as user interface performance~\cite{john2004predictive}, robotics~\cite{baron1994exploring}, autonomous driving~\cite{kolekar2021behavior,choi2021drogon,bhattacharyya2022modeling}, the study of psychological disorders~\cite{zeng297concurrent}, and agent-based simulations of collective behavior~\cite{dobson2019integrating,kennedy2011modelling}.  
Traditional approaches often rely on data-driven imitation learning~\cite{le2022survey} or machine theory of mind~\cite{rabinowitz2018machine}, which learn to reproduce human behavior from large datasets but typically lack explicit representations of underlying cognitive processes. Reinforcement learning~\cite{sutton2018reinforcement,cao2019overview} can model sequential decision-making and adaptation, yet, by itself, it rarely accounts for intrinsic human limitations such as memory decay, attentional bottlenecks, or time–effort trade-offs~\cite{fuchs2023modeling}.

Recent work in cognitive science and HCI has instead moved toward \emph{computational rationality}~\cite{oulasvirta2022computational,lieder2020resource}, which extends the classic notion of bounded rationality~\cite{simon1957models} by describing human cognition as an optimization process constrained by limited computational resources. Under this view, people act approximately rationally—not to maximize objective accuracy or reward, but to optimize performance relative to their internal costs of computation, memory, and attention. Deviations from normative rationality are thus interpreted as adaptive trade-offs rather than biases or errors.

Computational rationality provides a unifying lens for modeling cognition in interactive and decision-making contexts. In HCI, it has been used to explain how users allocate attention and choose interaction strategies that balance accuracy and effort in tasks such as touchscreen typing~\cite{shi2024crtypist}, menu search~\cite{oulasvirta2022computational}, and pedestrian decision-making under uncertainty~\cite{wang2025pedestrian}. Within cognitive science, the framework has informed rational models of strategy selection and meta-reasoning~\cite{lieder2017strategy}, demonstrating how people decide which cognitive strategy to deploy given resource constraints.

In the context of XAI, computational rationality provides a principled way to describe user behaviors as adaptive: users aim to make sense of explanations efficiently, balancing the cognitive effort of understanding against the potential gain in decision quality. This framing connects variations in user performance with the underlying constraints of human cognition, without assuming that one explanation style is inherently superior.

\section{Study Approach and Background}

To understand why explanations are effective (or not), we aim to simulate human reasoning when learning from and using two common XAI Schemas---Rules and Weights \footnote{Note that the underlying surrogate model is logistic regression, but we call it "Weights", to illustrate the type of information shown to users.}---by developing a cognitive model of user reasoning strategies.

We focus on tabular data because it provides structured, discrete attributes and remains an integral component of many real-world applications in healthcare, finance, and operations~\cite{caruana2015intelligible,borisov2022tabular,grinsztajn2022tree}.
Our choice of using global explanations instead of instance-based ones is motivated by the natural process of reasoning with rules, where people tend to apply the same rules broadly rather than forming different rules for each instance~\cite{bruner1956study,shepard1961learning,gigerenzer1996fast,maddox2004rulebased}.

We describe our overall approach, and introduce background concepts and context.

\subsection{User-Centered Design for Cognitive Modeling}

\begin{figure}[t]
    \centering
    \includegraphics[width=6.0cm]{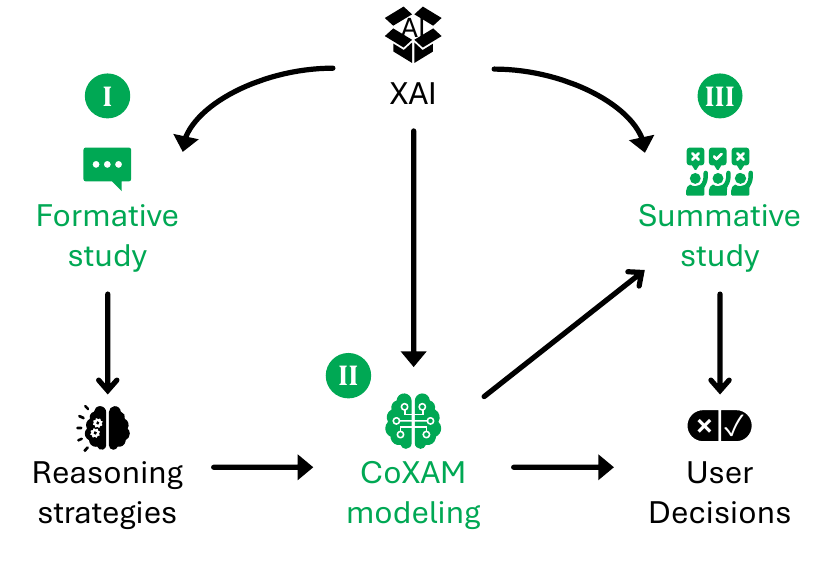}
    \caption{Overall approach to model users in XAI understanding across three studies: 
    I) Formative study to elicit reasoning strategies, 
    II) Modeling study to examine user reasoning, and
    III) Summative study to observe decisions.
    }
    \Description{This figure illustrates the overall workflow of the research approach. The XAI Schemas inform two empirical components: a Formative Study that identifies participants’ reasoning strategies, and a Modeling Study (CoXAM) that formalizes these strategies to simulate user reasoning. The resulting model is then evaluated in a Summative Study using user decision data from forward and counterfactual tasks.}
    \label{fig:overall-approach}
\end{figure}

To simulate human interpretation of different XAI Schemas, we have to identify cognitive processes and reasoning strategies from actual users, encode this in a cognitive model, and evaluate against real user decision data. Hence, we conducted three studies (illustrated in Fig.~\ref{fig:overall-approach}):

\begin{enumerate}
    \item[I)]\textit{Formative user study} to elicit and characterize various reasoning strategies for Rule-based and Weight-based XAI interpretation.
    \item[II)]\textit{Modeling study} to develop a cognitive model and evaluate it with baseline models for predictions of user (mis)interpretations, which may result in good or poor decisions and performances.
    \item[III)] \textit{Summative user study} to collect data on user decisions on a large scale to associate with each reasoning strategy, and calibrate and evaluate our modeling.
\end{enumerate}

\subsection{XAI Schemas: Rule-based Decision Tree and Weight-based Linear Regression}

Weight-based and rule-based explanations remain among the most popular explanation schemas of XAI methods for lay users.
With weight-based explanations, 
users are informed which attributes are more influential toward a decision.
However, the explanation is not actionable to know how to change a decision.
Instead, rules (if-then) indicate explicit thresholds or antecedents when the result will change.
Yet, it remains unsettled which XAI Schema is more interpretable for users.

We focus on applying these XAI Schemas for surrogate global explanation
and study the use of linear regression for Weights~\cite{bo2024incremental, poursabzi2021manipulating}, and decision tree for Rules~\cite{kozielski2025towards, angelino2018corels}.
Furthermore, we study the combined use of both Weights and Rules to investigate if the consolidated learning is beneficial.
Through our formative study, we will identify reasoning strategies that users employ for each XAI Schema, and examine their efficacy and investigate why.

\begin{figure*}[t]
    \centering
    \includegraphics[width=12.5cm]{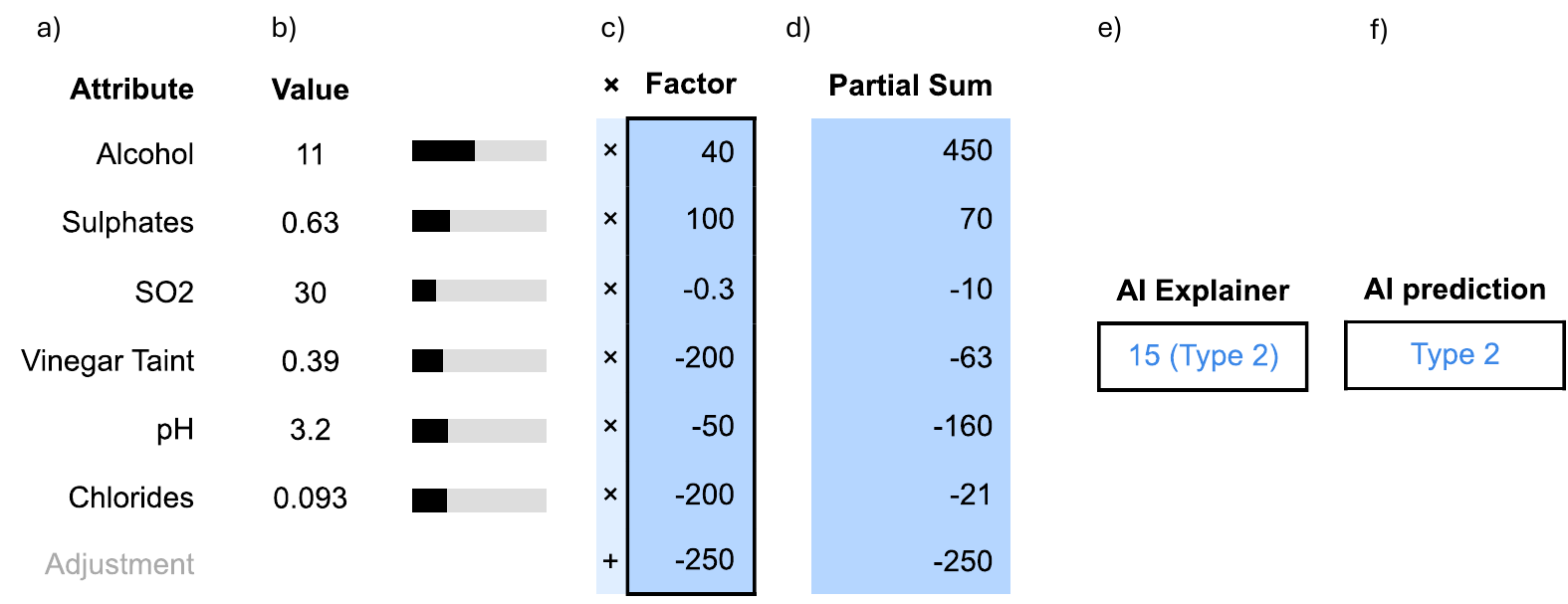}
    \caption{
    Weight-based XAI with Tabular UI components for:
    a) Attribute names of the instance.
    b) Value of each attribute with bar to indicate how high/low.
    c) Factors to multiply with each attribute value.
    d) Partial contributions of each attribute based on value $x$ factor.
    e) Prediction of Weights explanation for current instance.
    f) Prediction of AI.
    a--b are always shown; c is only shown in trials with Weights explanation; d, e, f is not shown for decision testing.
    }
    \Description{This figure shows the user interface for the Weights XAI condition in the tabular instance task. The interface presents six attributes with their corresponding values and visual bars indicating magnitude. For each attribute, the model displays its linear coefficient (Factor) and the resulting contribution (Partial Sum), computed as value × factor. The overall Factors prediction is displayed on the right alongside the AI’s predicted class label. Panels (a)–(b) are always visible, while panels (c)–(f) appear only when Weights explanation is provided.}
    \label{fig:ui-lr}
 \end{figure*}
\begin{figure*}[t]
    \centering
    \includegraphics[width=\textwidth]{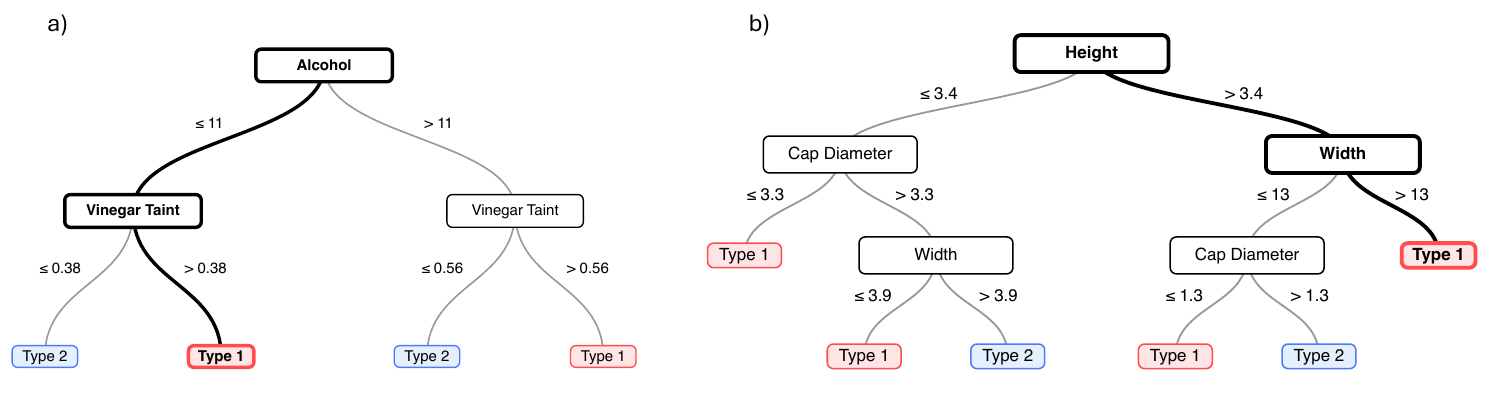}
    \caption{
    Rule-based XAI with Decision Tree UI for a) low (wine quality) and b) high complexity (mushrooms).  The black outline highlights the path that corresponds to the current instance values.
    }
    \Description{This figure shows the Rules explanation user interface used in the experiment for two example datasets. Panel (a) illustrates a low-complexity tree from the Wine Quality task, while panel (b) depicts a higher-complexity tree from the Mushrooms task. Each node represents an attribute and its decision threshold, with terminal leaves showing the predicted class labels. The black-highlighted path indicates the traversal corresponding to the current instance’s feature values, helping users visualize how the AI reaches its prediction.}
    \label{fig:ui-dt}
\end{figure*}

To convey the instance attributes and decision task, we employed a tabular user interface (UI), similar to Bo et al.~\cite{bo2024incremental}.
Fig.~\ref{fig:ui-lr} (a--b, e--f) shows the baseline components of the attribute values and prediction labels from the AI and XAI models.
Fig.~\ref{fig:ui-lr} (c--d) shows the Weights explanation from the Linear Regression XAI model.
Fig.~\ref{fig:ui-dt} (a--b) shows the Rules explanation from the Decision Tree XAI model.

    \vspace{0.5cm}

\subsection{Forward and Counterfactual Simulation Decision Tasks}

We focus on decision tasks that are commonly supported by XAI. Doshi-Velez and Kim~\cite{doshi2017towards} defined \textit{human simulation} tasks to assess users’ understanding of an AI model’s behavior. 
In this work, we focus on \textit{forward simulation}, where the user estimates (simulates) the AI’s prediction from the instance’s attribute values and any accompanying XAI explanation; and \textit{counterfactual simulation}, where the user changes an attribute to yield a different AI prediction. These tasks evaluate whether users have formed an accurate mental model of the AI’s decision logic, and their understanding of how specific attributes contribute to changes in the outcome.
Formally, given an input $\mathbf{x} = (x_1, \dots, x_n)^\top$, the AI predicts a binary classification
\begin{equation}
    \hat{y} = f_{\text{AI}}(\mathbf{x}),
\end{equation}
where $\hat{y} \in \{-1, 1\}$.
This can be explained by an XAI model with human-interpretable prediction:
\begin{equation}
    \tilde{y} = g_{\text{XAI}}(\mathbf{x}, \hat{y}, f_{\text{AI}}).
\end{equation}
In the forward simulation task, the user estimates the AI’s prediction based on the instance’s attributes and/or the explanation:
\begin{equation}
    \breve{y} = f_{\text{AI}}^{\text{user}}(\mathbf{x}, g_{\text{XAI}}),
\end{equation}
where ideally $\breve{y} \approx \hat{y}$ or $\breve{y} \approx \tilde{y}$, indicating that the user’s simulated prediction aligns with the AI or XAI model respectively.

For the counterfactual simulation task, the user modifies the value of an attribute $r$ in $\mathbf{x}$ by $\Delta x_r$ to obtain a new instance $\mathbf{x}'$, and the AI produces a new prediction:
\begin{align}
    \breve{\mathbf{x}}' &= h_{\text{CF}}^{\text{user}}(\mathbf{x}, \hat{y}, g_{\text{XAI}}), \\
    \hat{y}' &= f_{\text{AI}}(\breve{\mathbf{x}}').
\end{align}
A successful counterfactual $\breve{\mathbf{x}}'$ should lead to the target AI outcome, i.e., $\hat{y}' \neq \hat{y}$.
For this work, for user simplicity, we limit our scope to only changing one attribute in $\mathbf{x}$.

\subsection{Evaluation Context: Application Scenarios, Datasets, and AI Model}

For accessibility to lay users, we chose two publicly available datasets: the UCI Mushroom dataset~\cite{secondary_mushroom_848} (N = 61,068) featuring physical attributes for poisonous/edible classification and the Wine Quality dataset~\cite{wine_quality_186} (N = 4,898), binarized for a low/high quality classification task based on chemical compositions. We limited both datasets to only 6 attributes. 

For both datasets, we trained a simple Multi-Layer Perceptron (MLP) model to predict the respective classification targets: wine quality and mushroom edibility. The models were trained on 80\% of the data and tested on 20\%, resulting in 85.0\% test accuracy for the Wine Quality dataset and 78.2\% for the Mushrooms dataset.

\section{Formative Study of User Reasoning on XAI} \label{sec:formative-user-study}
We conducted a formative user study to elicit user reasoning strategies and mental models when using or recalling different XAI Schemas (Weights or Rules) for the forward and counterfactual simulation tasks (Task Type), under varying test conditions (with or without XAI).

\subsection{Experiment Method and Procedure}
\label{sec:formative-user-study-experimental procedure}

Each participant was randomly assigned to one Application Scenario (between-subject: Mushroom, or Wine Quality). To reduce bias and the impact of prior knowledge, we anonymized the prediction labels as Type 1 and Type 2, instead of their original connotations like "poisonous" or "poor quality".

After consenting to the study, the participant went through the following procedure.
For each of two XAI Schemas (randomly ordered, within-subject),
\begin{enumerate}
    \item Introduction to the Study (see Appendix Fig. \ref{fig:UI-start}).
    \item Consent to participate. This study was approved by the university institutional review board (IRB).
    \item Tutorial on user interface (Fig. \ref{fig:UI-LR} for Weights, Fig. \ref{fig:UI-DT} for Rules).
    \item Forward simulation session (x10 trials) 
    \begin{enumerate}
        \item Predict w/o XAI. The UI shows only values and a visualization slider (Fig. \ref{fig:ui-lr}b). Participants can hover over the slider bars to view attribute ranges in the dataset. They then make a binary choice (Type 1 vs. Type 2) for the AI System’s predicted output (Fig. \ref{fig:UI-FR-NO}). 
            \item Predict w/ XAI. Participants saw either factors for the Weights explanation (Fig. \ref{fig:UI-LR-FR}) or a decision tree for Rules explanation (Fig. \ref{fig:UI-DT-FR}, without the instance path highlighted). The XAI prediction itself was hidden to encourage reading the explanation. The thresholds and factors were shown to 3 significant figures on hover.
        \item Feedback of AI prediction. Participants reviewed their answers against the AI System’s prediction (Fig. \ref{fig:UI-LR-FR-FD} and \ref{fig:UI-DT-FR-FD}). Frequent review supported learning from mistakes and greater focus on the Weights or Rules explanations.
    \end{enumerate}
    \item Counterfactual simulation session (x10 trials). No AI feedback on the edited instances were shown during this session.
    \begin{enumerate}
        \item Change attribute value w/o XAI. Participants were instructed to "select one attribute and change it as little as possible" (via sliders; Fig. \ref{fig:UI-CF}) to alter the AI output.   
        \item Change attribute value w/ XAI. With either the Weights (Fig. \ref{fig:UI-LR-CF})  or  Rules (Fig. \ref{fig:UI-DT-CF}) explanation shown, participants reviewed the explanation, then chose and adjusted an attribute using the slider.
    \end{enumerate}
\end{enumerate}

We asked participants to think aloud and share their reasoning process as they used the XAI to perform forward and counterfactual simulations. This included identifying what parts of the UI they focused on, how they interpreted the explanations, their preferences of the XAI Schemas, and any points of confusion.

\subsection{Findings}
We recruited 24 participants from a local university, with an average age of 22, and 11 of whom were female (13 male). All were undergraduate students from various majors. With participant consent, we recorded both the interactions with the XAI UI and the conversation audio. Each participant was compensated with \$7.80 USD in local currency for 30 min of participation time. As participants were partly guided by the experimenters, their performance on the task is not reported.

Focusing on our objective of understanding how users interpret rule-based or weight-based XAIs, we center our qualitative analysis on their reasoning strategies across XAI Schemas (Weights, Rules) and under varying test conditions (with or without XAI).

We thematically coded the transcripts of the think-aloud study following the principles of grounded theory analysis~\cite{charmaz2014constructing, strauss1990basics, glaser2017discovery}. 
First, the first and second co-authors conducted open coding using in-vivo codes, with codes generated directly from participant utterances (e.g., \textit{``multiplying the value and factor''}, \textit{``changing the bottom attribute''}).
Next, these authors performed axial coding to identify relationships among the open codes and organized them into coherent reasoning strategies.
These two stages were conducted through multiple rounds of iteration.
Then, selective coding was used to merge similar strategies, including those observed across with- and without-XAI trials, but separated by XAI Schema. 
Finally, the identified reasoning strategies were reviewed by the last senior co-author and refined.
Throughout this process, we followed the grounded-theory heuristic of constant comparison to ensure that important distinctions between strategies were not overlooked.

\subsubsection{Forward Simulation Task}
For the forward simulation task, we identified two common strategies for the Weights explanation and one for the Rules explanation. 

\textbf{Approximate Calculation (Weights).}
Some participants attempted to follow the mathematical calculations of the XAI explicitly, by multiplying the factors with the attribute values, and summing up the partial contributions. They applied two mental shortcuts: 
1) focusing on attributes that had larger effects on the sum, and 
2) using approximations instead of exact calculations when determining the sign of the sum (for a binary decision).  
Regarding focused attributes, 
P5 dismissed small partial sums of pH and SO2 as \textit{``just fractions of the values''}, and 
P2 \textit{``mainly looked at the [mushroom] Height and Width, plus the adjustment [intercept]''}. 
Regarding approximate calculations,
P12 \textit{``didn’t do calculation to exact value, but I do [it] approximately. Like I just see if the number [partial sum] is big or small compared to each other, and do a very rough estimate''}.
However, approximations were insufficient for ambiguous cases, e.g.,
P3 found that \textit{``the exact math involved is too complicated for me, so when the partial sum [total sum] is near to the boundary of 0, it is hard to decide''}. 
When the XAI was not shown, participants recalled the factor values and carried out the same process. 
Confident in his memory, P11 felt that \textit{``there is no difference if I see the explanation because I can remember the whole thing [the factors] anyway''}.
But others were less certain: P3 \textit{``forgot the factors, I just remember the first one [for Bruises] is minus 500.''}

\textbf{Feature Attribution (Weights).}
Some participants avoided doing the multiplication math and instead conceived approximations of the partial sums (feature attribution = factor × value) based on the visual representation of the values (Fig. \ref{fig:ui-lr}). 
They neglected the factor numbers, except for noting the direction (sign) toward or against the decision. 
P16 mentioned \textit{``the [attribute] width value is very high [based on the visual representation], so maybe it is Type 1. I always remember this [Width factor] value is minus and this one [Height factor] is positive so I mainly do prediction by them''} .
P2 remarked that \textit{``if I can see that the Vinegar Taint and pH on the lower side and Alcohol on the higher side, then the [total sum] number will be positive, then [the AI prediction] will be Type 2 and vice versa''}. 
Participants may calibrate their weight of influence for attributes as they correct their thinking across trials, e.g.,
P13 realized that \textit{``Sulphates is high, maybe, so it contributed to Type 2 more than I thought''}.

\textbf{Attribute-Threshold-Attribute Traversal (Rules).} 
When reading the Rules explanation, participants traced the path from the root by navigating based on the thresholds at each node;
e.g., P7 \textit{``was able to follow the path, where the Alcohol [attribute value] now is 13, more than threshold of 11, and then Vinegar Taint is less than 0.56, it leads me to Sulphates in this case the value now is 0.77, more than 0.5, so it should be Type 2''}.
However, when deciding without XAI, 
participants recalled the root attribute and its threshold, applied the threshold, and recalled the subsequent attribute and its threshold, recursively. This allowed them to follow a trace along recalled branches.
For example, P7 \textit{``remember[ed] the Alcohol cutoff is at 11, so it’s the left branch, then I remember for Vinegar Taint it is less than 0.33, and then Alcohol is tested again and it is more than 9.8, so it is Type 2''}. 
However, participants may recall the wrong attribute or threshold: e.g., P14 said she \textit{``tried remembering the model [Decision Tree], and recalling it ..., [she] sometimes could not figure out the variables [attribute names]''}.
P10 \textit{``remembered the wrong [threshold] number, I compared it with 0.56, but I should compare with 0.72''};

\subsubsection{Counterfactual Simulation Task}
For the counterfactual task trials, we identified: 
two strategies for Weights explanation, 
two for Rules explanation exclusively, and
one common strategy for both XAI Schemas. 


\textbf{Inverse Calculation (Weights).}
When XAI was provided, participants would decide how to change one attribute with the following steps:
1) determine how much to change the predicted score to cross the decision threshold ($y = 0$),
2) select the most sensitive attribute (with highest factor magnitude),
3) derive how much to change by dividing the score change needed by the factor number.
For example, P11 thought \textit{``maybe I reduce Height by 4, because I need to make it [the total sum] a negative number, ..., so it should be 290 divided by 90 [Height factor] which is 3 point something''}. 
Some participants kept the new attribute values within realistic bounds, but sometimes found that this was not sufficient to achieve the target prediction, e.g., P3 \textit{``was trying to make the partial sum negative to make it to Type 1, and previously I changed Alcohol values to the minimum value, but realized the factor size is too low and it won’t contribute to the negative partial sum''}.
However, when no XAI was shown, participants were unable to sensibly determine the counterfactual change.
P11 recalled \textit{``first trying to think what is like the magnitude [of the total sum] like, ... and I am not sure, am I supposed to calculate the sum, because that is quite challenging. I will probably not calculate everything''}.


\textbf{Inverse Feature Attribution (Weights).}
Rather than explicitly perform Inverse Calculation, participants approximated partial sums using a Feature Attribution heuristic. They then intuitively estimated how much to adjust an attribute to bring the perceived total contribution in their mental model back to zero.
Attribute selection was again biased toward attributes with larger weights in the participant’s internal model. A particularly pronounced case occurred when participants focused solely on the most influential factor where P18 reasoned that the \textit{``Width has always been a big positive factor, so I’ll just change it until [I think] the prediction flips''}.
Crucially, this strategy remained effective without XAI, since participants relied on internalized models of Feature Attribution rather than the displayed total sum.


\textbf{Root Threshold Crossing (Rules).}
Some participants changed the value of the attribute corresponding to the root node, by moving it across the corresponding threshold value. This was a straightforward and cognitively simple strategy with P7 identifying that the same attribute can be changed consistently \textit{``rather than changing by following the branches [of the Decision Tree]''}. This strategy is robust, even without XAI, where P7 \textit{``remembered the threshold value and just hovered around [crossed] it''}.

\textbf{Leaf Threshold Crossing (Rules)}
Participants often modified the values of the attributes at the leaf nodes. This was an intuitive strategy, as in most cases, the neighboring node had the opposite label (though occasionally the neighbor was an internal node; see Fig.~\ref{fig:ui-dt}b, right-most node). 
For example, P5 \textit{``followed the paths, ... changed the Sulphate [leaf node] to less than 0.5, it will change to Type 1.''}
Without XAI, participants attempted to simulate traversal from memory, where P14 \textit{``recalled the graph [Decision Tree] mentally and went through it to decide''}. However, incomplete recall of deeper nodes sometimes caused participants to stop prematurely, making the strategy less consistent.

\textbf{Availability Heuristic.}
Participants reused changes they had applied in earlier trials. For example, P5 mentioned they could just use \textit{``the previous strategy of increasing the alcohol value''} while P11 \textit{``did the same change of Sulphates as last time''}.
Though cognitively cheap, this strategy was flawed because the previous change might not be applicable to the current instance.  

\section{CoXAM: Cognitive XAI-Adaptive Model}

We aim to understand, via modeling, how people can learn about, interpret, and decide on various XAI Schemas.
We propose CoXAM, a unified cognitive model that can adapt to interpreting different XAI Schemas---Rules and Weights---across decision tasks---forward simulation and counterfactual simulation.
Not only does this enable a fair comparison of these different XAI Schemas and decision tasks on the same platform,
it facilitates the modeling of a single user learning from and interpreting multiple XAI Schemas together.

We combine 
i) a memory model that determines what can be recalled, 
ii) a drift-diffusion process that converts an evidence number into a choice and decision time, and 
iii) a computational-rational controller that allocates effort and selects strategies to trade accuracy for time. We instantiate three forward simulation strategies and four counterfactual strategies within this framework.

\subsection{Memory Model}

Our memory model follows the account of declarative memory in ACT-R~\cite{anderson2004integrated} that defines memory \emph{chunks} and how they are retrieved.
Chunks are modeled as structured records of slot-value pairs (e.g., \{\textsc{Attribute}: Alcohol, \textsc{Value}: 10\}). 
Given a \emph{retrieval cue} $\mathbf{q}$ specified as a target slot value (e.g., \{\textsc{Attribute}: Alcohol\}),
chunks that were used more often, more recently, and that match the cue are more likely to be recalled. 
Concretely, for a retrieval request at time, $t$, each chunk, $i$, is given an activation value, $A_i(\mathbf{q}, t)$, and the chunk with the highest activation is recalled if it exceeds a retrieval threshold, $\kappa$; otherwise, retrieval fails.
For each chunk, these criteria are computed with an \textit{activation} at retrieval time $t$:
\begin{equation}
A_i(\mathbf{q}, t)
= \ln\!\Bigg(\sum_{j=1}^{n_i} (\Delta t_{ij})^{-0.5}\Bigg) - m_i(\mathbf{q}) + \varepsilon,
\label{eq:activation-simplified}
\end{equation}
where $n_i$ is the number of past uses of the chunk, $\Delta t_{ij}$ is the time since the $j$-th use of the chunk, $m_i(\mathbf{q})$ is the count of slot mismatches with the cue, $\varepsilon \sim \mathrm{Logistic}(0,\zeta)$ is an error noise following a Logistic distribution, and $\zeta>0$ is the activation noise scale.
The chunk will be retrieved if the activation exceeds a retrieval threshold, i.e., $A_i(\mathbf{q}, t) > \kappa$.

\subsubsection{Memory Chunks for Weights explanation}
Each factor is stored as its own chunk and retrieved using the attribute as the cue. \{\textsc{Attribute}: \emph{name}, \textsc{Factor}: \emph{value}\} describes a template chunk, which would be recalled with a cue: \{\textsc{Attribute}: \emph{name}\}. Each time the factor is read or recalled, the chunk’s usage count is incremented, which increases its activation and subsequent recall likelihood. Strategy-specific variants may add slots to the chunk and are noted later.

\subsubsection{Memory Chunks for Rules explanation}
We represent the Rules explanation using four chunk types, with each node keyed by a unique \textsc{Node id}. Repeated usage increases activation similar to the Factor chunks.
\begin{enumerate}
    \item \textit{Attribute}: stores the node’s attribute, retrieved by the cue of the node id. Example: \{\textsc{Node id}: \emph{i}, \textsc{Attribute}: \emph{name}\}.
    \item \textit{Threshold}: stores the node’s threshold, with retrieval cue of node id and attribute. Using both as cues can increase mix-ups when multiple nodes share an attribute. Example: \{\textsc{Node id}: \emph{i}, \textsc{Attribute}: \emph{name}, \textsc{Threshold}: \emph{$\tau$}\}.
    \item \textit{Child}: stores each outgoing branch, with retrieval cue of node id and branch. Example: \{\textsc{Node id}: \emph{i}, \textsc{Branch}: Left/Right, \textsc{Child node id}: \emph{j}\}.
    \item \textit{Leaf Label}: stores the class at a leaf (no threshold/children). Example: \{\textsc{Node id}: \emph{i}, \textsc{Label}: +1/-1\}.
\end{enumerate}

\subsection{Forward Simulation Strategies}
\label{sec:forward-strategies-technical}

We formalize the three reasoning strategies elicited in our formative study: Approximate Calculation and Feature Attribution on Weights explanation, and Attribute-Threshold-Attribute Traversal for Rules explanation.

We model how each reasoning strategy supports or refutes a decision.
Given a binary decision $y \in \mathcal{Y} = \{-1, +1\}$, each reasoning strategy (described in Sections~\ref{sec:forward-strategies-technical} and \ref{sec:counterfactual-strategies-technical}) will compute an evidence $e$ for ($+1$) or against ($-1$) the decision.
We use a Drift-Diffusion Model (DDM)~\cite{ratcliff1978memory, ratcliff2004comparison, bogacz2006physics, wiecki2013computational} to estimate the \textit{probability} of deciding $y = 1$ with the reasoning strategy that produces evidence $e$, and the expected \textit{execution time} $T$.
Specifically, we use the standard logistic approximation~\cite{bogacz2006physics, navarro2009fast}:
\begin{align}
    P(y=+1|e) &\approx \sigma\!\left(\frac{2 a\, e}{\kappa^{2}}\right), \\
    \mathbb{E}[T(e)] &\approx \dfrac{a}{|e|}\,\tanh\!\Big(\dfrac{a|e|}{\kappa^{2}}\Big), 
\end{align}
where $\sigma$ is the sigmoid function, 
$a > 0$ is an effort parameter that controls the time spent and consequent probability, 
and $\zeta > 0$ is a diffusion noise parameter that slows the speed of reasoning. $a$ is set dynamically using the controller described in section \ref{sec:computational-rationality}, while $\zeta$ is tuned as a cognitive parameter of each participant.

\subsubsection{Approximate Calculation (Weights)}
For each trial, the strategy chooses to look at a subset of the attributes, $\mathcal{X}$, which is again chosen by the controller in section \ref{sec:computational-rationality}. 
It multiplies each of the selected attribute values $x_r$ with the corresponding factor $w_r$ to get a feature attribution "partial sum" $a_r = w_r x_r$, sums the total $\breve{y} = \sum_{r \in \mathcal{X}} a_r$, determines whether the total sum $\breve{y}$ is positive or negative, and makes the binary decision $\text{sgn}(\breve{y})$. 
We define the DDM evidence $e$ of the strategy based on 
whether the total sum $|\breve{y}|$ is distinct (far from 0) or ambiguous (close to 0), and
whether the feature attributions $|a_r|$ are concordant (mostly in the same direction) or ambivalent (many on either direction). Formally,
\begin{equation}
    \label{evidence-approximate-calculation}
    e = \frac{|\sum_{r \in \mathcal{X}} w_r x_r|}{\sum_{r \in \mathcal{X}} |w_r x_r|}
\end{equation}

\subsubsection{Feature Attribution (Weights)} 
This strategy shares some similarities with the Approximate Calculation strategy, but employs mental simplifications for an intuitive approach rather than the rational one of the latter~\cite{kahneman2011thinking}.
For the $r$th attribute, instead of reading or recalling the factor $w_r$ this strategy uses a nebulous mental representation $\omega_r$ instead.
At retrieval, following ~\cite{karelaia2008determinants}, we define this as a random sample from a Normal distribution:
\begin{equation}
    \label{factor-mental-representation}
    \omega_r \sim \mathcal{N}(\mu_r,\sigma_r^2)
\end{equation}
where $\mu_r$ and $\sigma_r$ are the mean and standard deviation parameters, respectively, of the mental model's guess of the factor. The $\mu_r$ and $\sigma_r$ are stored as individual chunks in memory replacing the explicit factor values for the Approximate Calculation strategy.
For the first trial, we initialize each $\mu$ to either $+1$ or $-1$ depending on the sign of the provided factor, and each $\sigma$ to 1.
These parameters are updated as more trials are encountered using the provided AI prediction $\hat{y}$ as feedback, with a Laplace update for a linear regression model under a gaussian prior~\cite{bishop2006pattern, murphy2012machine, mackay1992practical} (refer to Appendix \ref{sec:feature-attribution-update}).

The DDM evidence is estimated similarly to Eq. \ref{evidence-approximate-calculation} but with the latent mental factors:
\begin{equation}
    \label{evidence-feature-attribution}
    e = \frac{|\sum_{r \in \mathcal{X}} \omega_r x_r|}{\sum_{r \in \mathcal{X}} |\omega_r x_r|}
\end{equation}
Note that the calculations are only performed in the cognitive model as an implicit, intuitive mechanism and does not represent an explicit rational calculation by the user.

\subsubsection{Attribute-Threshold-Attribute Traversal (Rules)}
When the Rules explanation is provided, starting from the root node, 
this reasoning strategy reads the node $\eta$,
evaluates the attribute value $x_\eta$ against the threshold value $\tau$, and decides whether to follow the left (lesser $x_\eta < \tau_\eta$) or right (greater $x_\eta \geq \tau_\eta$) branch. 
It then recursively applies this comparison at each node until it reaches a leaf.

Without XAI, the strategy recalls the Rules explanation by recalling the root node, evaluates its threshold to select the subsequent branch, then recalls the next node and recurses.
We define the DDM evidence $e$ of the strategy based on 
whether each attribute was distinctly clear from the threshold for all traced nodes, i.e.,
\begin{equation}
    \label{evidence-traversal}
    e = \sum_{\eta} | x_\eta - \tau_\eta |
\end{equation}


\subsection{Counterfactual Simulation Strategies}
\label{sec:counterfactual-strategies-technical}

We formalize the four reasoning strategies elicited in our formative study: Inverse Calculation and Inverse Feature Attribution for Weights explanation, and a merged Node Threshold Crossing and Availability Heuristic for Rules explanation.
Unlike the forward simulation task, counterfactual simulation is not a binary decision task, so we cannot use DDM to model probabilities of the choice. We do not model the execution timing of the different strategies, though we include reading and calculation timing measures based on Section \ref{sec:reasoning-time-technical-approach}.



\subsubsection{Inverse Calculation (Weights)}
With XAI provided, this strategy selects one attribute $\rho$ based on a probability distribution proportional to the factor magnitudes $|w_r|$ across attributes, i.e., attributes with larger factors are more likely to be selected, i.e.,
\begin{equation}
    P(\rho | \tilde{y}) = \frac{|w_\rho|}{\sum_r |w_r|}
\end{equation}
With the $\rho$th attribute selected, the strategy adjusts its value in the opposite direction of the current XAI Total Sum $\tilde{y} = \sum_r w_r x_r$, just enough to cross 0 to the other prediction label, normalized by the attribute's factor $w_\rho$ i.e.,
\begin{equation}
    \label{evidence-inverse-calculation}
    \Delta x_\rho = -\Big( \frac{\sum_r w_r x_r}{w_\rho} + \varepsilon \Big)
\end{equation}
where $\varepsilon > 0$ is a margin parameter to cross the $y = 0$ decision boundary.
Without XAI, this strategy was not used by participants in the formative study, and would not work.



\subsubsection{Inverse Feature Attribution (Weights)}
This strategy is the same as the Inverse Calculation, but uses internal weights $w_{r}^{\mathrm{(int)}}$ and the calculated internal sum instead.
Like the Feature Attribution strategy for forward simulating on Weights explanation, 
this strategy is the mentally intuitive simplification of Inverse Calculation, where explicit factors $w_r$ are replaced by mental representation $\omega_r$, i.e.,
\begin{equation}
    \label{evidence-inverse-feature-attribution}
    \Delta x_\rho = -\Big( \frac{\sum_r \omega_r x_r}{\omega_\rho} + \varepsilon \Big)
\end{equation}
The mental factors are the same as those in Eq. \ref{factor-mental-representation}, since the same cognitive model can handle both forward and counterfactual simulation tasks.


\subsubsection{Node Threshold Crossing (Rules)}
With XAI, along the trace $\mathcal{T}$ for an instance's attributes, 
this strategy chooses the tree node at depth $d$ to change the attribute value $x_d$ toward the node's threshold $\tau_d$.
In our formative study, we observed that participants chose to change the attribute at the root ($d = d_{\min} =0$) or leaf ($d = d_{|\mathcal{T}|}$).
We generalize this by allowing any other depth $d \in [0, d_{|\mathcal{T}|}]$ to be predetermined as a cognitive parameter.
This strategy assumes that the choice of node $\eta$ is probabilistic following the truncated Normal distribution with normal mean $\mu = d$ and standard deviation $\sigma$, i.e., 
\begin{equation}
    P(\eta | d) = \frac{1}{\sigma} 
    \frac{\mathcal{N}(\frac{\rho - d}{\sigma})}{\varPhi(\frac{d_{\mathcal{T}} - d}{\sigma}) - \varPhi(\frac{d_{\min} - d}{\sigma})}
\end{equation}
where $\mathcal{N}$ is the probability density function of the standard normal distribution, and $\varPhi$ is its cumulative distribution function.

Given a selected node $\eta$, this strategy shifts the attribute value $x_\eta$ in the direction toward the threshold $\tau_\eta$ and exceeding by a margin $\varepsilon$, i.e., the counterfactual change in attribute value is
\begin{align}
    \Delta x_\eta &= -(x_\eta - \tau_\eta) \\
    \Delta x_\rho &= -(\Delta x_\eta + \text{sgn}(\Delta x_\eta) \varepsilon)
\end{align}
where $\text{sgn}(*) = +1$ if $* \geq 0$ and $-1$ otherwise is the sign function, 
$\Delta x_\eta$ is the measured difference between attribute value and threshold for the considered node $\eta$, 
$\Delta x_\rho$ is the amount to change the selected attribute $\rho$.

Without XAI, the strategy forward simulates the corresponding path based on its memory but might be forced to stop an earlier depth if it is unable to recall the full path.

\subsubsection{Availability Heuristic}
Since performing the inverse reasoning is mentally demanding, participants may take a mental shortcut of reusing their recent decisions.
This strategy models this by storing each recently used strategy as a memory chunk, recording the change made. 
It then recalls a chunk probabilistically and repeats it, using the retrieval cue of \{\textsc{Target Class}: \{-1, +1\}\}.

\subsection{Reasoning Strategy Selection}
We introduce a method to select between the 3 strategies for forward simulation and the 4 strategies for counterfactual simulation, unifying the model into a single combined framework. The strategy selection is based on the concept of computational rationality, where the reasoning time is compromised with the utility of each strategy. We first describe how we modify the probability distribution from the above strategies to account for random errors, then describe how total reasoning time is computed for the strategies and subsequently describe the computational rationality formulation and our specific implementation of it using a Reinforcement Learning (RL) model.

\subsubsection{Choice Lapsing}
To account for occasional random choices, which could occur due to mistakes we do not model such as motor or visual lapses, we 
adjust the probability $P$ of each reasoning strategy deciding $y = +1$ as a combination of the strategy-based probability, and random chance for choosing $y = +1$, i.e., 

\begin{equation}
    P(\tilde{y} = +1) \leftarrow (1-\lambda) P(y=+1) + \lambda/|\mathcal{Y}|
\end{equation}
where $0 \leq \lambda < 1$ is the lapse rate to choose randomly~\cite{wichmann2001psychometric, prins2012psychometric, frund2011inference}, 
normalized by number of possible decisions $|\mathcal{Y}| = 2$ for binary decisions. We fix $\lambda$ to be $.05$, based on prior works~\cite{prins2012psychometric,witton2017psychophysical,ratcliff2008diffusion}.

\subsubsection{Reasoning Time}
\label{sec:reasoning-time-technical-approach}
In addition to the execution time $T(e)$, we add coarse times for reading and explicit mental calculation. We use a simplistic model where each read takes an expected time of 1s, to account for visual search over $\sim$15--20 items ($\approx$0.4--0.8s), and $\sim$200ms to internalize the number or text~\cite{wolfe2021guided,rayner1998eye}. Each mental calculation takes an expected time of 2s, consistent with timing for two-significant-figure arithmetic~\cite{ashcraft1992cognitive}. The expected total time for the strategy, is,
\begin{equation}
    \mathbb{E}[T(s)] = \mathbb{E}[T(e)] + \mathbb{E}[T_r(s)] + \mathbb{E}[T_c(s)].
\end{equation}
where $\mathbb{E}[T_r(s)]$ and $\mathbb{E}[T_c(s)]$ are the total times taken for reading and explicit calculations respectively. Explicit calculations are only used in the Approximate Calculation and Inverse Calculation strategies.
Our objective is not to match response timings precisely, but rather coarsely account for the variation for time between strategies, which is an important feature of the strategy selection.

\subsubsection{Computational Rationality for Strategy Selection}
\label{sec:computational-rationality}
Following Lieder and Griffiths~\cite{lieder2017strategy},
we model the selection of a reasoning strategy $s$ as a rational choice.
For each strategy, this determines the Value $V$ of the strategy based on the Utility $U$ of the decision and the Cost of the expected reasoning time $T$ to arrive at the decision.
For forward simulation trials, the utility is based on whether the decision $\breve{x}$ from the strategy correctly matches the AI label $\hat{y}$, i.e.,
\begin{equation}
    U(\breve{y} = \hat{y} | \hat{y}) = 
    \begin{cases}
      P(\breve{y} = +1) & \text{, if $\hat{y} = +1$} \\
      1 - P(\breve{y} = +1) & \text{, if $\hat{y} = -1$}
    \end{cases}
\end{equation}

The utility and time varies based on the reasoning strategy $s$ chosen and its cognitive parameters $\mathbf{\theta}_s$ (e.g., effort $a$, tree trace depth $d$).
Hence, the Value of the reasoning strategy is:
\begin{equation}
    V(s) = U(\breve{y} = \hat{y},s, \mathbf{\theta}_s) - \gamma \mathbb{E}[T(s, \mathbf{\theta}_s)],
\end{equation}
where $\gamma$ is the opportunity-cost parameter to determine whether the cognitive model prioritizes utility or speed.

For counterfactual trials, the utility term is instead instantiated with the probability of changing the AI prediction under the proposed edit distribution. 
\begin{equation}
    U(\hat{y}' \neq \hat{y}, s, \theta_s)
= \Pr_{\rho \sim P}\!\left[\, f_{\text{AI}}\!\big(x+\Delta x_\rho\big) \neq \hat{y} \,\right].
\end{equation}

where $P$ is the strategy output distribution for selecting the attributes and $f_{AI}(x + \Delta x_\rho)$ is the AI prediction for edited counterfactual of the instance.

\subsubsection{Controller and Learning a Strategy Policy}
\label{sec:controller-rl}

We use a multi-layer perceptron (MLP) trained using a reinforcement learning (RL) framework, to model how a participant might adaptively choose reasoning strategies over time. We call this model a controller.

In reinforcement learning, an agent learns through trial and error: it observes a \emph{state} of the environment, takes an \emph{action}, and receives a numerical \emph{reward} indicating how good that choice was.

At each trial $t$, the controller observes a continuous state vector which concatenates three components: 1) cognitive parameters of retrieval threshold, $\alpha$, diffusion noise, $\nu$ and opportunity cost, $\gamma$; 2) strategy history which includes summary statistics of each strategy’s past use for the current episode (mean time, success rate, and frequency); and 3) mean partial sum values of each feature.
This state captures what the controller currently “knows” about their own performance and the task context.

The controller then chooses an action
\[
a_t = (s, a, \mathcal{X}, d) \in \mathcal{A},
\]
where $s$ is a discrete choice of reasoning strategy, $a$ is a the effort level, $\mathcal{X}$ is an optional subset of attended attributes for Approximate Calculation and Feature Attribution strategies and $d$ is an optional trace depth for the Node Threshold Crossing strategy.

After taking an action, the controller receives a Reward $R = V(s)$
where $V(s)$ is the Value of the selected strategy. 
The controller learns a stochastic policy $\pi_\phi(a_t \mid s_t)$ that increases the likelihood of actions leading to higher long-term reward.
We optimize this policy using proximal policy optimization (PPO)~\cite{schulman2017proximal}, a standard RL algorithm that balances improvement and stability.

Because the controller cannot know in advance how accurate or time-consuming each strategy will be on a new dataset, training it only on the target task would make it overfit and start with poor expectations.  
We therefore first \emph{pre-train} the controller on several auxiliary tabular datasets to learn general priors about strategy success rates and timing.  
During inference on the target domain, these priors allow the controller to make adaptive choices under uncertainty, approximating how a participant might draw on past experience when facing a new task.

\subsection{Cognitive Parameters}

We therefore tune four core cognitive parameters to match each individual participant:
\begin{enumerate}
\item Diffusion noise $\nu>0$ (only for forward simulation): Higher $\nu$ shrinks the effective evidence, yielding slower and less accurate choices for the same boundary $a$.
\item Retrieval threshold $\kappa$: Higher $\kappa$ increases retrieval failures from memory.
\item Opportunity cost $\gamma>0$: Larger $\gamma$ favors faster strategies (lower $a$, smaller $|\mathcal{X}|$) at the expense of accuracy.
\item Margin $\varepsilon > 0$ (only for counterfactual simulation): Larger $m$ pushes edits further beyond the decision boundary.
\end{enumerate}

\section{Summative Study}

We conducted a summative user study to investigate the effect of the three different XAI Schemas on user understanding. These collected responses also provide a validation basis for CoXAM.


\subsection{Experimental Design}

We manipulated three independent variables:
i) XAI Schemas (Rules, Weights, and Hybrid; between subjects), ii) XAI complexity (Low and High; between subjects), and iii) Tested with XAI (with or without XAI; within subjects).

Each participant was assigned to one of three \textit{XAI Schemas}: Rules, Weighs or Hybrid. In the Hybrid condition, the XAI Schema was randomly chosen on each trial between Rules and Weights, thereby exposing participants to both forms of XAI. This condition provides a stronger test of our model’s XAI adaptability, as participants could flexibly employ strategies suited to either explanation type. We also varied \textit{XAI complexity} at two levels (between subjects): Low, featuring a Decision Tree of depth 2 or a sparsified Linear Regression with 3 non-zero factors, and High, featuring a Decision Tree of depth 3 or a full Linear Regression with 6 non-zero factors. These manipulations allowed us to better evaluate the model’s generalizability.

For both the forward simulation and counterfactual phases, we controlled the instance selection to satisfy (i) prediction labels of the AI are balanced to be 50\% for each label type and (ii) each XAI prediction matched the AI prediction for 90\% of the instances. For each trial with XAI shown, participants using the Hybrid XAI are randomly shown one of Weights or Rules explanations.

We measured dependent variables:
\begin{enumerate}
\item Forward Simulation Accuracy: proportion of matches between user labels and AI predictions.
\begin{equation}
\text{Forward Accuracy } = \frac{1}{N} \sum_{k=1}^{N} { \mathbbm{1} (\breve{y}_k = \hat{y}_k)}
\end{equation}
 where $\mathbbm{1}()$ is the indicator function, $\breve{y}_k$ is the user response and $\hat{y}_k$ is the indicator function.

Since the fidelity of each XAI was controlled to be 90\%, we do not report the redundant comparison between user labels and XAI predictions. 

\item Counterfactual Simulation Accuracy: proportion of trials in which participants successfully changed the AI’s prediction by modifying an attribute, creating new instance $\mathbf{\breve{x}'}$. 
\begin{equation}
\text{Counterfactual Accuracy} = \frac{1}{N} \sum_{k=1}^{N} {\mathbbm{1}(f_{AI}(\breve{x}') \neq \hat{y}_k)}
\end{equation}
where $f_{AI}(\breve{x}')$ is the AI prediction for the new instance. 

\end{enumerate}

\subsubsection{Experiment Procedure}
Each participant performed the following procedure: Trials in the forward and counterfactual sessions are similar to the formative user study (refer to Section \ref{sec:formative-user-study}), 

\begin{enumerate}
    \item Introduction to the study.
    \item Consent to participate.
    \item Tutorial on the user interface.
    \item{} Tutorial on the XAI UI. Following which, we asked three screening questions about the UI, and the two XAI explanations (see in Appendix, Figs. \ref{fig:UI-data-scr}, \ref{fig:UI-LR}, \ref{fig:UI-DT}).
    \item Session 1: Forward Simulation ($\times40$ trials)
    \begin{enumerate}
        \item View the interface with attribute values (with or without XAI) and estimate the AI prediction (see Fig. \ref{fig:UI-FR1} for without XAI; Fig. \ref{fig:UI-LR-FR}–\ref{fig:UI-DT-FR} for with XAI).
        \item Receive feedback on the AI prediction [with XAI (Figs. \ref{fig:UI-LR-FR-FD}–\ref{fig:UI-DT-FR-FD}) or without XAI (Fig. \ref{fig:UI-FR-NO})].
    \end{enumerate}
    \item Session 2: Counterfactual Simulation ($\times40$ trials): Select an attribute and change its value [with or without XAI] (Fig. \ref{fig:UI-CF}-\ref{fig:UI-LR-CF})
    \item Answer demographic questions.
\end{enumerate}

We recruited 340 participants (202 female, median age 36) via Prolific with the minimum requirement of 20 participants per condition. An additional 300 participants were excluded based on a straightforward comprehension screening (Figs. \ref{fig:UI-LR}, \ref{fig:UI-DT}). Each was compensated \pounds{7.00} for an average completion time of 51 mins. The variation in complexity and datasets, is intended to test the generalization of CoXAM and whether the previous elicited strategies persist in a broader online population.

\subsection{Results}
\begin{figure*}[t]
    \centering
    \includegraphics[width=12cm]{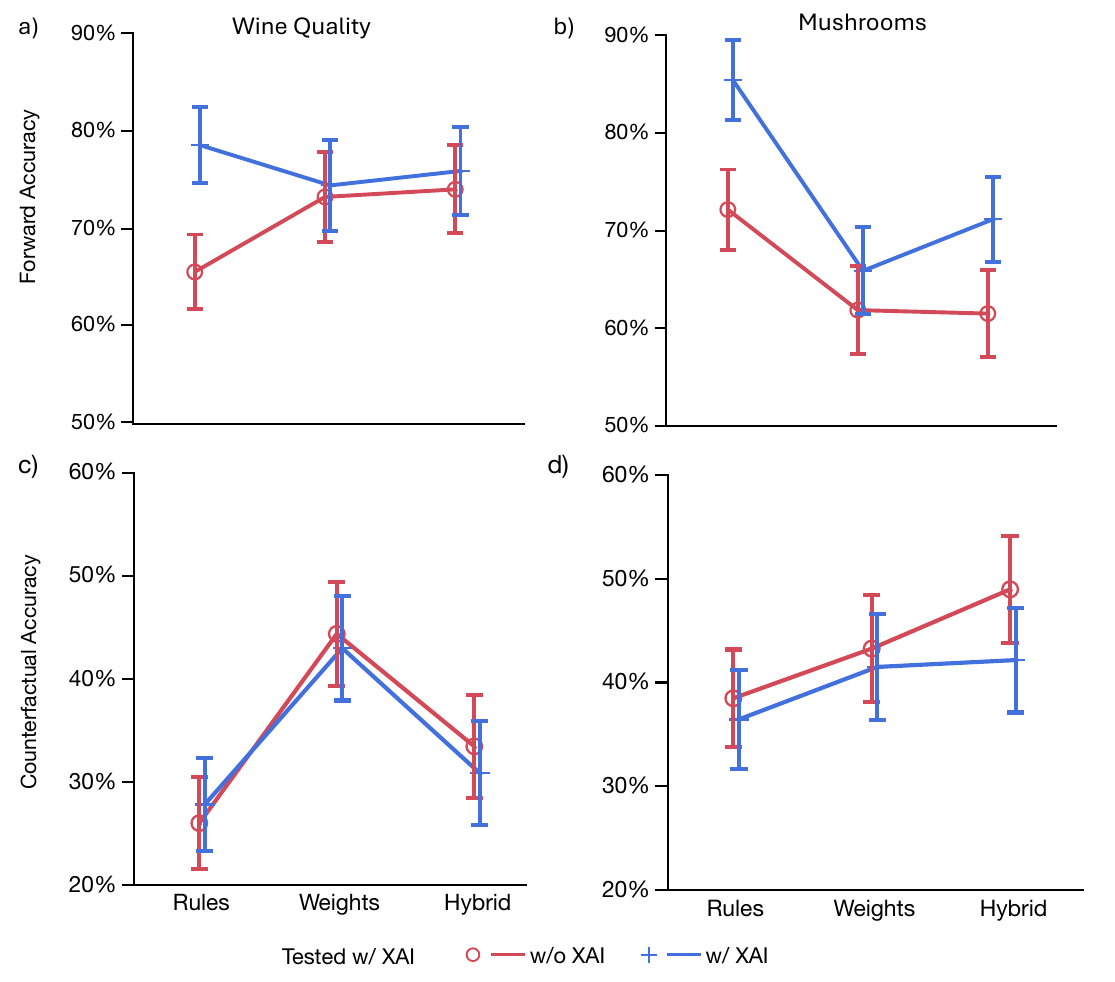}
    \caption{
        User performance of forward simulation task in a), and b) and counterfactual simulation task in c), and d), across datasets: Wine Quality (left), and Mushrooms (right) across XAI Schemas, and testing conditions (w/o XAI, w/ XAI). Error bars are 95\% CIs. Random performance for the forward simulation task would be 50\%.
    }
    \Description{Panels (a) and (b) show forward simulation accuracy for the Wine Quality and Mushrooms datasets, respectively, while panels (c) and (d) present counterfactual simulation accuracy for the same datasets. Results are reported for three XAI Schemas (Rules, Weights, and Hybrid) under two testing conditions: without explanations (red, w/o XAI) and with explanations (blue, w/ XAI). Error bars represent 95\% confidence intervals. Forward simulation accuracy is consistently above the random baseline of 50\%, though the relative benefits of explanations differ across datasets and model types. Counterfactual simulation accuracy is lower overall, with modest improvements depending on model type and explanation condition.}
    \label{fig:res-forward-counterfactual}
\end{figure*}

We performed a linear mixed effects model fit on each dependent variable as the response, Complexity nested within XAI Schemas, Trial Index and Tested with XAI (with or without XAI trial) as fixed effects, an interaction effect between the XAI Schemas and Tested with XAI, and Participant as a random effect.
Complexity was found to have insignificant fixed and interaction effects, and hence we omitted it from our analysis for clarity. We briefly discuss significant results as these findings are not the main contribution of our work. We only report pairwise comparisons using contrast t-tests that are highly significant, p $<$ .005.

\subsection{Forward Simulation}

Participants in the Rules condition performed significantly better in With-XAI trials compared to Without-XAI trials across both datasets (Fig.~\ref{fig:res-forward-counterfactual}a--b), Mushrooms (78.7\% vs.\ 63.9\%, contrast test, $\text{p} < .001$) and Wine Quality (74.8\% vs.\ 63.9\%, $\text{p} < .001$). In contrast, participants in the Weights and Hybrid conditions showed no significant difference between with- and without-XAI trials. This suggests that Rules explanations are harder to recall than to read, whereas Weights explanations can be more easily internalized.

Across XAI Schemas, the Wine Quality dataset appeared more amenable to Weights explanation than to Rules explanation with a significant improvement in accuracy in Without-XAI trials (72.3\% vs. 63.9\%, contrast test: $\text{p} < .005$). Conversely, for the Mushrooms dataset, Weights had lower accuracy in with-XAI trials relative to Rules (65.6\% vs. 78.7\%, contrast test: $\text{p} < .0001$). Two factors likely contribute to this increased performance of Rules: (1) the Decision Tree for the Mushrooms data is slightly smaller, containing two fewer leaf nodes (see Fig.~\ref{fig:ui-dt}b), and (2) the Mushrooms dataset exhibits more nonlinear relationships than Wine Quality.

\subsection{Counterfactual Simulation}

Counterfactual performance showed a similar pattern as shown in Fig.~\ref{fig:res-forward-counterfactual}c--d, with Weights outperforming Rules (43.6\% vs. 26.8\%, contrast test: $\text{p} < .001$) on the Wine Quality dataset. For Mushrooms, no significant differences were found across XAI Schemas, possibly because the dataset’s structure aligns well with the Decision Tree representation, mitigating its typical disadvantages for counterfactual reasoning. Participants using the Rules explanation changed values more conservatively (in Fig. \ref{fig:absolute-change}), likely because they relied on recalling explicit threshold values when identifying attribute changes.

\section{Modeling Analysis}

To understand why certain XAI and reasoning strategies are more effective compared to others, we use CoXAM to simulate participant behavior and explain performance and trends using underlying cognitive causes. 

\subsection{Fitting CoXAM for Forward Simulation Trials}

\begin{table*}[t]
\centering
\small
\caption{Comparison of Negative Log-Likelihood (NLL) and Bayesian Information Criterion (BIC) for CoXAM against baseline proxies for forward simulation task for the Wine Quality and Mushrooms datasets. Lower values are better. Bold is best fit.}
\Description{This table compares model fits for the forward simulation task using Negative Log-Likelihood (NLL) and Bayesian Information Criterion (BIC) across two datasets (Wine Quality and Mushrooms) and three explanation conditions (Rules, Weights, and Hybrid). Lower values indicate better model fit. Overall, CoXAM achieves the best fit across most conditions, showing lower NLL and BIC values than baseline models (Decision Tree, Linear Regression, and KNN without XAI). An exception occurs in the Mushrooms dataset under the Rules condition, where the Decision Tree baseline attains a slightly lower BIC, suggesting close performance for that specific case.}
\begin{tabular*}{\textwidth}{@{\extracolsep{\fill}}lcccccccccccc}
\toprule
& \multicolumn{6}{c}{Wine Quality} & \multicolumn{6}{c}{Mushrooms} \\
\cmidrule(lr){2-7} \cmidrule(lr){8-13}
& \multicolumn{2}{c}{Rules} & \multicolumn{2}{c}{Weights} & \multicolumn{2}{c}{Hybrid}
& \multicolumn{2}{c}{Rules} & \multicolumn{2}{c}{Weights} & \multicolumn{2}{c}{Hybrid} \\
\cmidrule(lr){2-3} \cmidrule(lr){4-5} \cmidrule(lr){6-7}
\cmidrule(lr){8-9} \cmidrule(lr){10-11} \cmidrule(lr){12-13}
Models & NLL & BIC & NLL & BIC & NLL & BIC & NLL & BIC & NLL & BIC & NLL & BIC \\
\midrule
Decision Tree & 26.8 & 57.3 & --- & --- & 26.2 & 56.1 & 21.7 & \textbf{47.1} & --- & --- & 29.7 & 63.1 \\
Linear Regression & --- & --- & 28.0 & 59.7 & 26.8 & 57.3 & --- & --- & 37.0 & 77.7 & 35.5 & 74.7 \\
KNN w/o XAI   & 29.7 & 66.8 & 30.1 & 67.6 & 29.9 & 67.2 & 24.0 & 55.4 & 28.1 & 63.6 & 27.2 & 61.8 \\
\midrule
CoXAM  & \textbf{18.9} & \textbf{48.8} & \textbf{19.9} & \textbf{50.9} & \textbf{20.2} & \textbf{51.5} & \textbf{20.7} & 52.5 & \textbf{21.5} & \textbf{54.1} & \textbf{20.8} & \textbf{52.7} \\
\bottomrule
\end{tabular*}
\label{tab:coxam-vs-baselines-forward}
\end{table*}

To validate our modeling approach, we fit CoXAM to the 40 forward simulation trials for each participant individually with three tunable parameters: retrieval threshold, $\kappa$, opportunity cost, $\gamma$, and diffusion noise, $\nu$. The fitting was done using a Gaussian process-based optimization~\cite{snoek2012practical}. Due to the difficulty of integrating a static, global XAI explanation directly into a naïve machine-learning model as suggested by Chen et al.~\cite{chen2022use}, we implemented three alternative baseline proxies to approximate human predictions:
\begin{enumerate}
\item K-Nearest Neighbors (no XAI): Each participant is represented by a KNN model that incrementally updates its training set with the instances encountered so far, without incorporating any XAI information.
\item Decision Tree Proxy: For participants in the Rules and Hybrid conditions, we used the corresponding Decision Tree XAI model to generate predictions. To better match participants’ stochastic errors, we applied a single \emph{smoothing parameter} that blends the model’s predicted probability with a uniform distribution over possible outcomes. This accounted for occasional lapses or uncertainty in human responses.
\item Linear Regression Proxy: Analogous to the Decision Tree proxy, we used the Linear-Regression XAI to approximate participant behavior in the Weights and Hybrid conditions.\end{enumerate}
Table~\ref{tab:coxam-vs-baselines-forward} shows that CoXAM provides a more faithful account of participant responses than all three proxy models. The only exception occurs for the Decision Tree proxy on the Mushrooms dataset, which attains a slightly lower BIC despite a higher NLL, due to having less free parameters.

\begin{figure*}[t]
    \centering
    \includegraphics[width=14.0cm]{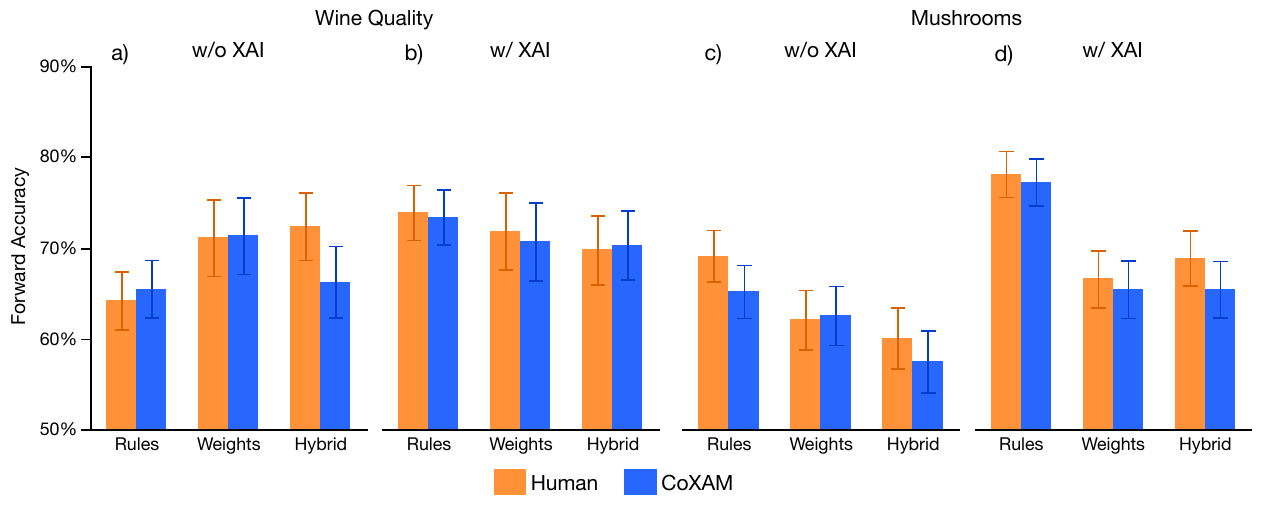}
    \caption{
        Results of summative user study (orange) compared to CoXAM (blue) of accuracy in the forward simulation task. Error bars are 95\% CI.
    }
    \Description{Forward simulation accuracy for human participants (orange) and the CoXAM model (blue) across explanation types and datasets, evaluated with and without XAI support. Results are reported for the Wine Quality dataset (a–b) and the Mushrooms dataset (c–d), under testing conditions without XAI (a, c) and with XAI (b, d). Explanation types include Rules, Weights, and Hybrid representations. Accuracy is computed with respect to the AI system’s ground-truth predictions, and error bars denote 95\% confidence intervals. Overall, CoXAM closely mirrors human performance patterns across explanation types and conditions, capturing both relative performance differences among explanation formats and the impact of XAI availability.}
    \label{fig:human-vs-coxam-forward}
\end{figure*}

\begin{figure*}[t]
    \centering
    \includegraphics[width=14.0cm]{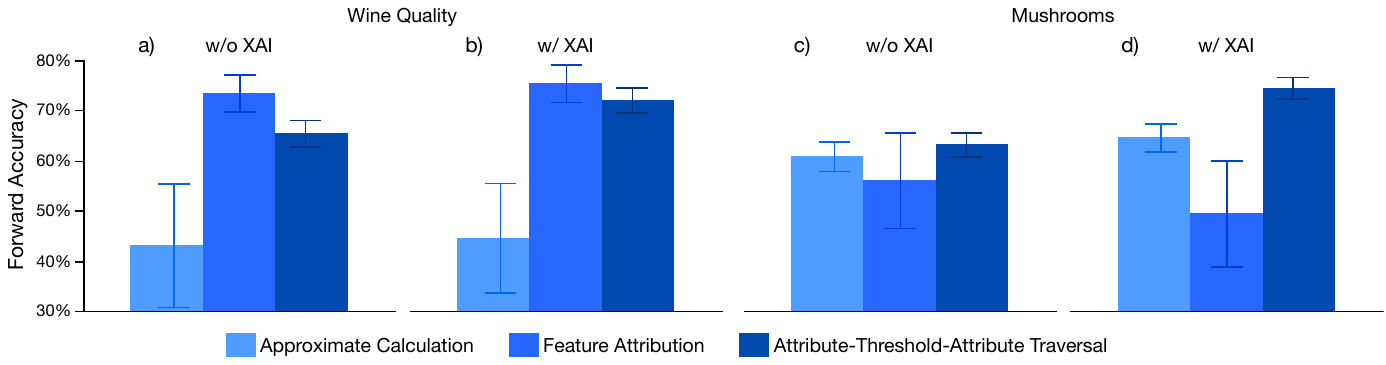}
    \caption{
        Forward simulation accuracy of two Weights and one Rules strategies for with and without XAI trials across datasets.
    }
    \Description{Forward simulation accuracy across explanation strategies, datasets, and XAI conditions.
The figure reports mean forward simulation accuracy (y-axis, \%) with error bars indicating variability across participants. Results are shown for two datasets—Wine Quality (a–b) and Mushrooms (c–d)—under conditions without XAI (a, c) and with XAI (b, d). Three explanation strategies are compared: Approximate Calculation, Feature Attribution, and Attribute–Threshold–Attribute (ATA) Traversal. For the Wine Quality dataset, Feature Attribution and ATA Traversal outperform Approximate Calculation in both conditions, with a consistent improvement when XAI is available. For the Mushrooms dataset, ATA Traversal achieves the highest accuracy with XAI, while Feature Attribution shows reduced performance relative to the other strategies. Overall, the results indicate that explanation strategy and the presence of XAI substantially influence users’ ability to accurately simulate model behavior, with rule-based ATA Traversal benefiting most from XAI support.}
    \label{fig:forward-strategy-acc}
\end{figure*}

\begin{figure*}[t]
    \centering
    \includegraphics[width=13cm]{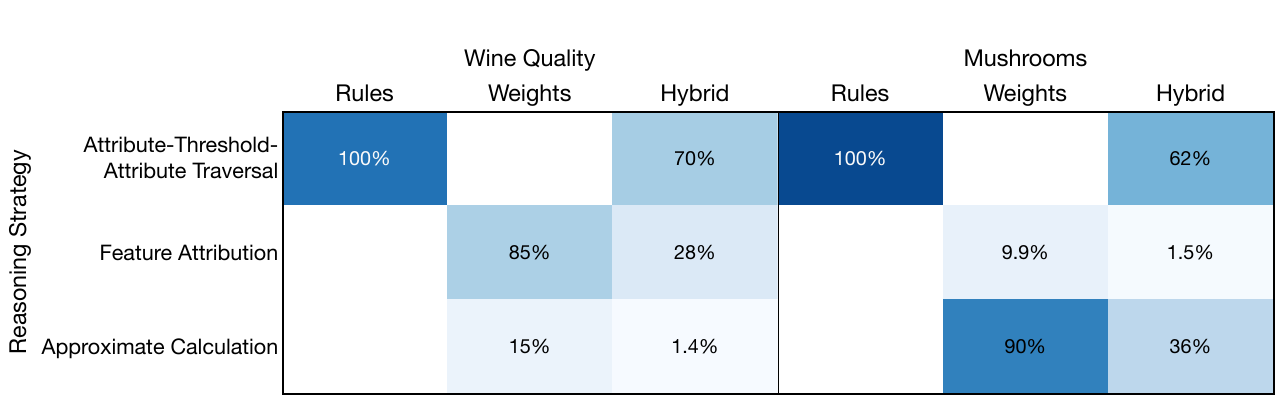}
    \caption{
        Strategy prevalence by XAI Schema for forward simulation.
    }
    \Description{This figure shows the prevalence of different reasoning strategies used by participants in the forward simulation task, grouped by XAI Schema and dataset. Each cell represents the proportion of trials where a given strategy was applied. For the Wine Quality dataset, participants predominantly used the Attribute Threshold/Traversal strategy with Rules and the Feature Attribution strategy with Weights XAI. In contrast, for the Mushrooms dataset, the Approximate Calculation strategy was dominant for Weights XAI, while Attribute Threshold/Traversal remained consistent for Rules. Hybrid XAI conditions show mixed usage patterns, indicating flexible strategy selection depending on task complexity.}
    \label{fig:forward-strategy-prevalence}
\end{figure*}

\begin{table*}[t]
\centering
\small
\caption{Comparison of Negative Log-Likelihood (NLL) and Bayesian Information Criterion (BIC) for CoXAM against baseline proxies for participant attribute-selection in the counterfactual task for the Wine Quality and Mushrooms datasets.Lower is better. Bold is best fit.
}
\Description{This table compares the Negative Log-Likelihood (NLL) and Bayesian Information Criterion (BIC) of CoXAM against two baseline models—Random and Global SHAP—across the Wine Quality and Mushrooms datasets, for three explanation types (Rules, Weights, and Hybrid). Lower values indicate better model fit. CoXAM consistently achieves the lowest NLL and BIC across all conditions, showing strong agreement with user behavior. While Global SHAP improves upon the Random baseline, it remains less predictive of participant decisions compared to CoXAM.}
\begin{tabular*}{\textwidth}{@{\extracolsep{\fill}}lcccccccccccc@{}}
\toprule
& \multicolumn{6}{c}{Wine Quality} & \multicolumn{6}{c}{Mushrooms} \\
\cmidrule(lr){2-7} \cmidrule(lr){8-13}
& \multicolumn{2}{c}{Rules} & \multicolumn{2}{c}{Weights} & \multicolumn{2}{c}{Hybrid}
& \multicolumn{2}{c}{Rules} & \multicolumn{2}{c}{Weights} & \multicolumn{2}{c}{Hybrid} \\
\cmidrule(lr){2-3} \cmidrule(lr){4-5} \cmidrule(lr){6-7}
\cmidrule(lr){8-9} \cmidrule(lr){10-11} \cmidrule(lr){12-13}
Model & NLL & BIC & NLL & BIC & NLL & BIC & NLL & BIC & NLL & BIC & NLL & BIC \\
\midrule
Random & 71.67 & 143.34 & 71.67 & 143.34 & 71.67 & 143.34 &
        71.67 & 143.34 & 71.67 & 143.34 & 71.67 & 143.34 \\
Global SHAP & 42.1 & 84.2 & 56.3 & 112.6 & 54.7 & 106.4 &
             50.2 & 100.4 & 55.1 & 110.2 & 51.0 & 102.0 \\
\midrule
CoXAM  & \textbf{35.6} & \textbf{71.2} & \textbf{51.8} & \textbf{103.6} & \textbf{45.0} & \textbf{90.0} &
         \textbf{45.8} & \textbf{91.6} & \textbf{50.6} & \textbf{101.2} & \textbf{46.6} & \textbf{93.2} \\
\bottomrule
\end{tabular*}
\label{tab:coxam-vs-baselines}
\end{table*}

\begin{figure*}[t]
    \centering\includegraphics[width=14cm]{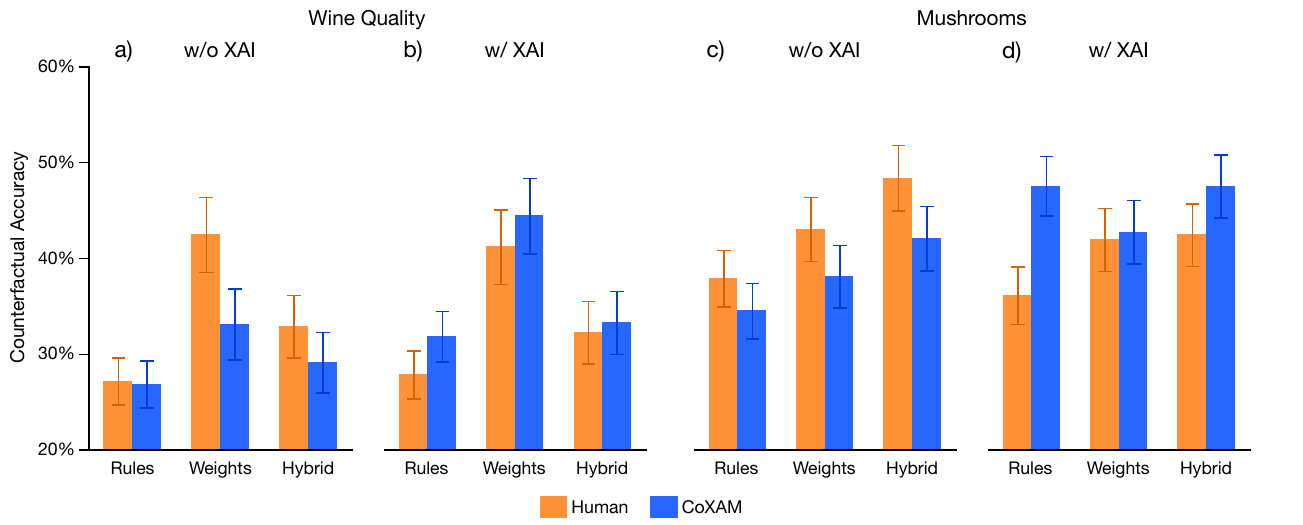}
    \caption{
        Results of summative user study (orange) compared to CoXAM (blue) of accuracy in counterfactual simulation task. Error bars are 95\% CI.
    }
    \Description{Counterfactual simulation accuracy for human participants (orange) and the CoXAM model (blue) across explanation types and datasets, evaluated with and without XAI support. Results are shown for the Wine Quality dataset (a–b) and the Mushrooms dataset (c–d), under testing conditions without XAI (a, c) and with XAI (b, d). Explanation types include Rules, Weights, and Hybrid representations. Accuracy is computed with respect to the AI system’s prediction change, and error bars denote 95\% confidence intervals. Overall, CoXAM closely tracks human performance across datasets, explanation types, and XAI conditions, reproducing relative performance differences and condition-dependent trends observed in human counterfactual reasoning.
    }
    \label{fig:human-vs-coxam-counterfactual}
\end{figure*}

\begin{figure*}[t]
    \centering\includegraphics[width=14cm]{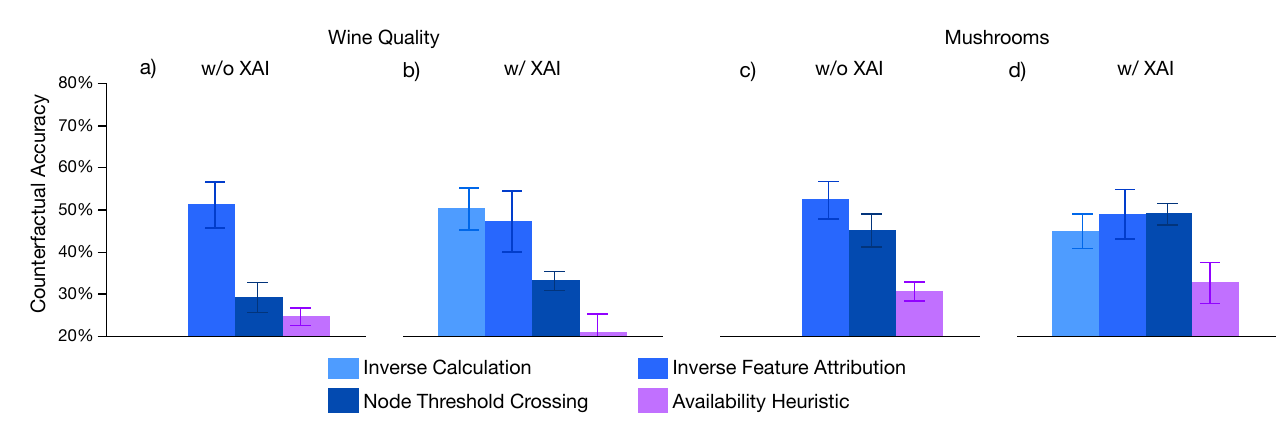}
    \caption{
        Results of CoXAM simulated user accuracy in the counterfactual simulation task, for different strategies. Error bars are 95\% CI.
    }
    \Description{Figure 9. Counterfactual simulation accuracy of CoXAM-simulated users across two datasets (Wine Quality and Mushrooms), with and without XAI support. Results are shown for four reasoning strategies: Inverse Calculation, Node Threshold Crossing, Inverse Feature Attribution, and Availability Heuristic. Error bars represent 95\% confidence intervals. Accuracy is highest for Inverse Calculation and Node Threshold Crossing, while Availability Heuristic consistently yields lower performance.
    }
    \label{fig:coxam-counterfactual-acc-strategy}
\end{figure*}

\begin{figure*}[t]
    \centering\includegraphics[width=13cm]{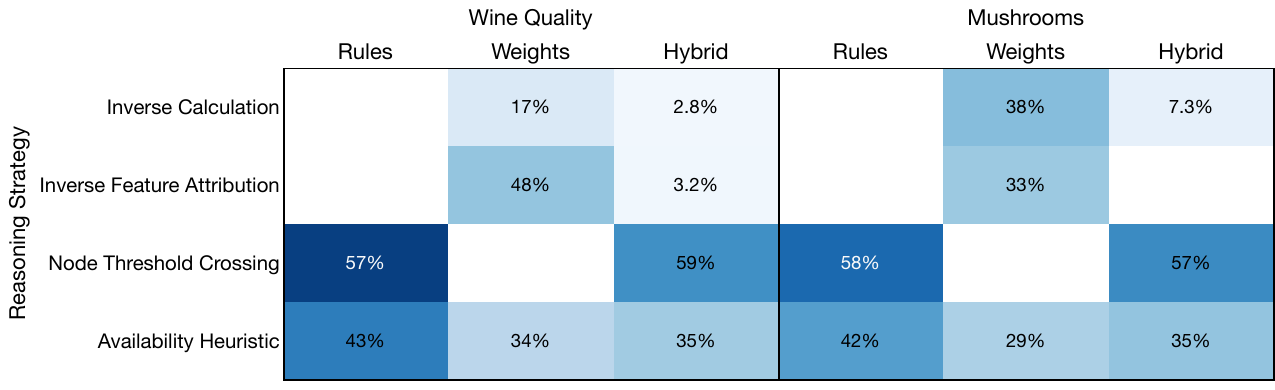}
    \caption{
        Strategy prevalence by XAI Schema for counterfactual simulation.
    }
    \Description{This figure shows how often different reasoning strategies were used in the counterfactual simulation task, grouped by XAI Schema and dataset. The rows list the four strategies: Inverse Calculation, Inverse Feature Attribution, Node Threshold Crossing, and Availability Heuristic. The columns represent the Wine Quality and Mushrooms datasets under Rules, Weights, and Hybrid explanation conditions. Each cell shows the percentage of trials in which that strategy was used. For Wine Quality, participants most often used Node Threshold Crossing with Rules explanations and Inverse Feature Attribution with Weights explanation. For Mushrooms, Availability Heuristic and Inverse Calculation were most common under Weights explanations, while Node Threshold Crossing remained frequent under Rules and Hybrid conditions.}
    \label{fig:counterfactual-strategy-prevalence}
\end{figure*}

We sampled responses from \textsc{CoXAM} using parameter distributions fitted per-participant to reproduce human forward-simulation performance across 12 experimental conditions (2 datasets × 2 conditions [with vs.without XAI] × 3 XAI Schemas). The resulting model accuracies (Fig.~\ref{fig:human-vs-coxam-forward}) were highly correlated with human performance (Pearson $r = .91$, $p < .0001$) and closely matched in magnitude (RMSE = 1.8\%), indicating that \textsc{CoXAM} effectively captured both the relative trends and absolute levels of participant accuracy.

\subsection{Analysis of Reasoning Strategies for Forward Simulation}

We analyzed the strategy selection distribution of CoXAM in the forward simulation task (Fig.~\ref{fig:forward-strategy-prevalence}). Under the Weights explanation condition, CoXAM primarily used the Feature Attribution strategy for the Wine Quality dataset, while favoring the Approximate Calculation strategy for the Mushrooms dataset. When analyzing strategy performance separately (Fig.~\ref{fig:forward-strategy-acc}), the model achieved higher accuracy with Feature Attribution (73.3\%) than with Approximate Calculation (48.1\%) on Wine Quality (contrast t-test, $p<.001$), prompting the controller to prefer Feature Attribution. In contrast, for Mushrooms, Approximate Calculation outperformed Feature Attribution (62.7\% vs.\ 52.8\%; $p<.001$), reflecting adaptation to dataset properties.

These differences align with task structure. Factor magnitudes in Mushrooms are smaller, so approximate mental arithmetic is easier, which benefits Approximate Calculation and supports the observed preference. By comparison, Wine Quality involves larger-valued attributes that make mental sums and differences harder, which reduces the effectiveness of approximate arithmetic and favors Feature Attribution under Weights explanations.

For the Hybrid XAI Schema, CoXAM favored the Attribute-Threshold-Attribute Traversal method across datasets. Although Feature Attribution slightly outperformed Attribute–Threshold
–Attribute Traversal strategy for the Wine Quality dataset, CoXAM preferred the latter due to its lower predicted execution time (9.1s versus 11.1s for Feature Attribution), indicating that the controller optimizes accuracy and reasoning time rather than accuracy alone. Detailed accuracy and timing per strategy are reported in Table~\ref{tab:forward-performance} in the Appendix.

We then examined sensitivity to cognitive parameters. Increasing the opportunity cost led the controller to select a smaller subset of attributes under Approximate Calculation (Fig.~\ref{fig:forward-against-cog-params}b), with a similar trend for Feature Attribution, indicating that higher perceived effort shifts behavior toward fewer, more diagnostic attributes. Accuracy declined with higher opportunity cost, diffusion noise, and retrieval threshold (Fig.~\ref{fig:forward-against-cog-params}a). Notably, retrieval threshold had the weakest effect on Feature Attribution, suggesting that this strategy is comparatively robust to memory limitations, while traversal and approximate arithmetic depend more on successful recall of intermediate quantities and path information.

\subsection{Fitting CoXAM for Counterfactual Simulation Trials}

We also fit CoXAM to the counterfactual simulation trials of participants with three parameters, analogous to the forward-simulation fitting, but replacing the diffusion noise parameter $\nu$ with the margin parameter $\varepsilon$. CoXAM was optimized to minimize the sum of (i) the negative log-likelihood of the user’s attribute-selection choices and (ii) the mean absolute error (MAE) between the participant’s change magnitude and CoXAM’s predicted change.

To evaluate how well CoXAM reproduces participants’ attribute-selection behavior, we compared it against two baseline models:
\begin{enumerate}
\item Random: A simple baseline assuming that each of the six attributes has an equal probability of being selected.
\item Global SHAP: A knowledge-based baseline assuming participants can identify and select attributes important to the underlying AI model. Each attribute’s selection probability is set proportional to its global importance, computed as the mean absolute SHAP value of the attribute across all instances in the training dataset.
\end{enumerate}

The comparison results demonstrate that CoXAM provides a closer approximation to human behavior than either baseline. The overall distribution of attribute selections is reported in the Appendix \ref{fig:attribute-select-counterfactual-distribution}.
Sampling from the attribute distribution output by CoXAM with its parameters fit on the users, we compare CoXAM's overall accuracies against human results (Fig. \ref{fig:human-vs-coxam-counterfactual}). Again, CoXAM aligns well with user performance across the 12 conditions (Pearson correlation r=72.1\%, RMSE=5.49\%).
\subsection{Analysis of Reasoning Strategies for Counterfactual Simulation}

Analyzing strategy selection revealed patterns consistent with the forward-simulation modeling while also highlighting counterfactual-specific effects. Under the  Weights explanation, CoXAM favored Inverse Feature Attribution more strongly for the Wine Quality dataset, whereas strategy use was more balanced between Inverse Feature Attribution and Inverse Calculation for Mushrooms.
Despite this difference in selection, both strategies achieved similar accuracies (Wine Quality: 50.3\% vs.\ 47.3\%; Mushrooms: 44.9\% vs.\ 48.9\%), indicating no clear performance advantage. A substantial proportion of trials relied on the Availability Heuristic, which consistently yielded the lowest accuracy, but was selected due to its speed of use.

In the Hybrid XAI condition, CoXAM strongly preferred the Node Threshold Crossing strategy, although the Availability Heuristic still appeared in a notable fraction of trials. For the Rules exlanation, CoXAM predicts that about 43\% of trials relied on the Availability Heuristic across both datasets, underscoring its frequent use despite poor accuracy.
Additionally, the Node Threshold Crossing method was very ineffective for the Wine Quality dataset, signalling that the decision threshold was not properly represented by the Rules explanation.

Sensitivity analysis of the cognitive parameters (Fig.~\ref{fig:counterfactual-accuracy-vs-parameters}) showed weaker and less consistent effects than in the forward-simulation task. Even for Node Threshold Crossing, which requires a forward traversal of the decision tree before counterfactual editing, accuracy was only modestly influenced by the retrieval threshold, suggesting that participants had largely internalized the XAI through prior forward-simulation exposure. A clearer effect emerged for the margin parameter $\varepsilon$, where counterfactual accuracy increased with larger margins across strategies. This trend indicates that the explainer boundaries differ substantially from those of the underlying AI, such that optimal counterfactual edits often require discounting the explainer’s boundary.

\section{Discussion}
We discuss our key findings about Rules and Weights explanation understanding, and the generalization and limitations of CoXAM.

\subsection{Rules vs. Weights}

Our experimental findings show that 
trends are dataset and strategy dependent even if the fidelity and cognitive load of the XAI is kept consistent (i.e., similar depth trees and values rounded to similar significant numbers in the interface).
We saw that participants are able to flexibly identify the more successful strategy and use it, with the Approximate Calculation and Feature Attribution strategies being preferred differently across the two datasets.
On the other hand, we found a consistent choice of favoring Rule-based strategies (Attribute-Threshold-Attribute Traversal and Node Threshold Crossing) in the Hybrid XAI condition for both forward and counterfactual tasks. This was despite the fact that the Feature Attribution method slightly outperformed the Attribute-Threshold-Attribute strategy for the Wine Quality dataset.
The reason for this choice was due to a shorter reasoning time with the latter method, emphasizing the importance of considering cognitive load while improving performance of metrics.

We also discovered that different strategies have different advantages. While Attribute-Threshold-Attribute is fast and effective when the XAI is shown, it is susceptible to poor memory recall. On the other hand, Feature Attribution is robust to memory recall, though by its incremental learning nature, has to take a few instances to be effective. Comparing the strategies and XAI Schemas without knowledge of the task requirements is not fruitful, and CoXAM provides an evaluation basis for future developers to identify the best reasoning strategies based on their requirements.

\subsection{Implications of Explainable User Interpretation with CoXAM}

Using the CoXAM cognitive model to simulate user interpretation of XAI, AI and HCI researchers, designers and developers can investigate the effectiveness of user reasoning or debug the causes of user misinterpretations.
Rather than focusing solely on "explainable AI," CoXAM prioritizes the "explainable human" within the XAI interaction loop.
This approach is not intended to replace human-subjects in a validation or evaluation study, but to facilitate the XAI method iteration.
Specifically, CoXAM assists in hypothesis formation to anticipate experimental results, allows designers to pre-empt interpretation failures to avoid wasteful user studies, and provides a framework to analyze why an XAI method might fail. Ultimately, subsequent user studies remain essential to validate the insights generated by the CoXAM model.

Suppose that CoXAM modeling finds that an XAI method would lead to poor understanding.
There are three approaches that could be used by various stakeholders to improve the XAI: tutorial, UI design, optimization.
The HCI researcher can prepare a \textit{tutorial} for the user study to prime the participants by training them employ the identified effective reasoning strategies instead of fallacious ones; e.g., using the Feature Attribution reasoning strategy rather than Approximate Calculation as we identified for the Wine Quality task.
The XAI UI designer can use CoXAM to systematically \textit{filter} the attributes included in an explanation by analyzing the trade-off between information gain and cognitive load; by simulating how user performance marginally improves or diminishes as the number of attributes increases, designers can identify the optimal attribute count and calibrate specific cognitive parameters to maximize interpretability without exceeding the target user's capacity.
The XAI developer can \textit{optimize} the XAI method to maximize user decision performance by formally regularizing for desiderata (e.g., cognitive load~\cite{abdul2020cogam}), or training an XAI model via backpropagation from a pre-trained neural network proxy based on CoXAM-labeled decisions like in \cite{li2024utilizing}.

\subsection{Scope and Generalization of CoXAM}

In this work, we had focused on binary classification over tabular datasets with six attributes, using forward and counterfactual simulation tasks to examine the effective use of XAI. 
While this design provided controlled insights, future work can examine implementing other types of explanation to scale our approach for XAI cognitive modeling. 
We examine generalizing CoXAM to different global XAI Schemas, explanation modalities, and decision tasks.

\newcommand{\R}[1]{\parbox[t]{\linewidth}{\raggedright\hspace{0pt}#1}}
\begin{table*}[ht]
\centering
\small
\caption{Scope of CoXAM for user interpretation of global XAI Schemas on interpretable features.}
\label{tab:coxam-scope}

\setlength{\tabcolsep}{4pt} 
\begin{tabular*}{\textwidth}{@{\extracolsep{\fill}}%
  p{0.18\textwidth}
  p{0.28\textwidth}
  p{0.54\textwidth}}
\toprule
\R{\textbf{XAI Schema}} &
\R{\textbf{Representative XAI Methods}} &
\R{\textbf{CoXAM Coverage}} \\
\midrule
\R{Rule-based} &
\R{Decision Tree~\cite{breiman2017classification}, Rule List~\cite{angelino2018learning}, Rule Set~\cite{lakkaraju2016interpretable}} &
\R{Covered: Topology-agnostic by converting to equivalent DNF. \newline
Future work: Account for differences in parsing topology.} \\
\addlinespace
\R{\makecell[tl]{Weight-based\\(Linear Additive)}} &
\R{Linear Regression~\cite{bo2024incremental}, Logistic Regression ~\cite{poursabzi2021manipulating}} &
\R{Covered.} \\
\addlinespace
\R{\makecell[tl]{Additive\\(Non-Linear)}} &
\R{Partial Dependence Plot ~\cite{krause2016interacting}, Generalized Additive Model (GAM)~\cite{hastie1990generalized}, Piecewise Linear~\cite{friedman1991multivariate,bo2024incremental}} &
\R{Future work: encode interpretation of nonlinear trends as local linear based on prototype examples~\cite{kalish2004population}.} \\
\addlinespace
\R{\makecell[tl]{Symbolic Regression\\(Non-Linear)}} &
\R{Domain-specific equations ~\cite{alaa2019demystifying}} &
\R{Future work: encode interpretation with rational model of function learning~\cite{lucas2015rational}.} \\
\addlinespace
\R{Example-based} &
\R{Prototypes ~\cite{narayanan2024prototype}, Criticisms~\cite{kim2016examples}} &
\R{Future work: encode examples in memory, retrievable based on computational rationality, to infer labels based on similarity~\cite{nosofsky1986attention}.} \\
\bottomrule
\end{tabular*}
\end{table*}

\subsubsection{Global XAI Schemas on Interpretable Features}

Focusing on global explanations with interpretable features, Table~\ref{tab:coxam-scope} shows which XAI Schemas are currently supported in CoXAM and how it could be extended to other schemas.
Although we had evaluated on decision trees for \textit{rule-based} XAI, other methods like Rule Lists~\cite{angelino2018learning} or Decision Sets~\cite{lakkaraju2016ids} with different topologies can also be converted to a unified representation in disjunctive normal form (DNF) to be ingested into CoXAM. 
However, this neglects the relative difficulties or cognitive short-cuts in interpreting these structures, which future work needs to elicit and model.

While CoXAM implements linear \textit{weight-based} explanations, this is a simplification of explainable \textit{additive} models, where each attribute contributes independently but nonlinearly to the AI decision. Such models include explanations with partial dependence plots~\cite{hastie1990generalized}, generalized additive models (GAM)~\cite{hastie1990generalized}, and piecewise linear models~\cite{friedman1991multivariate,bo2024incremental}.
Future work should explore how users encode nonlinear information, perhaps, piecewise with local linear segments around prototype examples~\cite{kalish2004population}.
For domains with specific \textit{symbolic} formulas or equations, or users with knowledge of basis functions (e.g., exponential, logarithm, sinusoidal), future work can model how users learn and interpret explicit functions~\cite{lucas2015rational}.

Finally, using analogous reasoning, people often learn key examples and make inferences based on them. To model user interpretation of example-based explanations, such as prototypes ~\cite{narayanan2024prototype} and criticisms~\cite{kim2016examples}, future work can model example retrieval based on computational rationality and infer labels based on similarity with the Generalized Context Model~\cite{nosofsky1986attention}.

\subsubsection{Multimodal Explanations}

Although predicting and explaining on structured data with symbolic reasoning is still paramount
~\cite{rudin2019stop, shwartz2022tabular, fang2024large}, 
many explanations are presented as visualizations, and 
recent machine learning models, such as deep learning and large language models (LLM), are facilitating new modalities in AI and multimodal explanations.
We discuss how CoXAM and cognitive modeling, can account for visual explanations, explanations of unstructured image data and LLM-generated text explanations.

Many XAI methods present their explanations as visualizations. For example, linear LIME as tornado plot~\cite{ribeiro2016should}, SHAP attributions as scatter plot~\cite{lundberg2017unified}, linear weights as lollipop charts~\cite{abdul2020cogam}, or in tables~\cite{bo2024incremental, poursabzi2021manipulating}.
In this work, we had focused on user reasoning of the semantic information in the explanations, rules and weights, which we had parsed before ingesting into CoXAM as memory chunks. 
However, the visual representations can affect the intuitiveness and readability of the explanations.
Following past works to model human visual attention and encoding~\cite{anderson1997actr_visual}, and computational rationality for chart reading~\cite{shi2025chartist}, future work can model how users read popular XAI visualizations to more precisely account for varying visual representations.

Other than tabular data, deep learning has been successful in reasoning on image data, and this necessitates image-based XAI.
Popular explanation techniques include pixel-based saliency maps
~\cite{selvaraju2017grad} and image segmentation~\cite{kirillov2023segany}. 
These explanations require users to take additional steps to interpret them~\cite{lim2025diagrammatization}, such as identifying relevant objects or concepts from the image and matching them to salient regions.
Indeed, concept-based explanations~\cite{koh2020concept, kim2018interpretability} support these steps by extracting concepts and communicating their prevalence or importance to the user.
Nonetheless, this clarifies that image-based explanations are two-staged: i) inferring concepts from perception, ii) symbolic reasoning on the interpretable concepts.
CoXAM already handles the latter, albeit without accounting for uncertain concepts.
Future work could be extended with a concept inference module, though this also requires encoding or making assumptions about the user's prior visual knowledge.

Recent developments in large vision-language models (VLMs) have led to highly usable, chat-based interactions for decision systems.
For example, instead of inputting a feature vector to a model, a user could supply an image and prompt \textit{``Here's a photo of a mushroom, GPT, tell me whether it is edible.''}, and receive a natural language explanation.
To interpret this, users construct mental models by extracting symbolic relations from the language~\cite{johnson1983mental}.
For future work, CoXAM could simulate this by integrating ACT-R-style production parsing~\cite{lewis2005activation}, or leveraging LLMs as symbolic parsers to map unstructured text onto structured attributes and rules~\cite{tian2024large}.
However, because LLM-based decision systems still have critical limitations of hallucination~\cite{ji2023survey}, sycophancy~\cite{fanous2025syceval}, and human-like biases~\cite{deng2025deconstructing}, they are not yet reliable for high-stakes settings. 
Nevertheless, given their rapid development and increasing adoption, this is a crucial extension for CoXAM.

\subsubsection{Decision Tasks}
We evaluated CoXAM user understanding based on forward and counterfactual simulation tasks, where the user estimates what the AI would predict, and how to change its prediction, respectively.
Another way to demonstrate understanding is to correctly \textit{attribute} the important factors for an AI decision.
This is attribution simulation~\cite{Waa2021EvaluatingXAI}.
Since the schema of XAI can take various forms, users can further demonstrate their understanding by reconstructing or recalling the explanation for decisions, such as in the form of decision trees~\cite{lim2009and}, non-linear line charts~\cite{abdul2020cogam}, saliency maps~\cite{boggust2022shared}, etc.

Future work could also examine human-AI collaboration tasks, where performance is defined by joint decision accuracy~\cite{doshi2017towards}. 
This would require modeling users with two mental models, one of the AI’s decision process and one of the ground-truth labels, as well as incorporating additional cognitive variables such as trust in the AI~\cite{wang2022will}. 
Finally, self-reported measures such as perceived usefulness, cognitive load, and ease-of-use are common in XAI user studies~\cite{rong2023towards}. Prior work has shown that such measures are often uncorrelated with objective assessments of understanding, and we therefore leave their modeling to future work.

\subsection{LLMs as Cognitive Models}

With the advent of Large Language Models (LLMs) and other foundation models, there has been a line of work attempting to use them to model human behavior~\cite{park2023generative, hwang2025human, park2024generative}.
While this field looks extremely promising, we argue that on its own, an LLM is still not at a phase to replace more traditional modeling approaches.
LLMs can replicate external human behavior, but they do not explicitly model components of the human brain and are not explicitly controllable, in contrast to cognitive models where behavior is defined through explicit functions.
Fitting an LLM to a human through persona-based prompting~\cite{sun2025persona, yeo2025phantom} with chain-of-thought reasoning~\cite{wei2022chain} may appear to rationalize~\cite{ehsan2019automated} like humans, but these are proxy explanations and not mechanistically traceable~\cite{rudin2019stop}.

We instead suggest, as in Chartist~\cite{shi2025chartist}, that LLMs could be used in conjunction with cognitive modeling 
to supplement the controller module in our framework with prior knowledge and flexible reasoning strategies while not operating independently as a black box model.

\subsection{Risk of Overreliance on XAI Cognitive Modeling}

While CoXAM facilitates systematic XAI evaluation, over-relying on it without considering its underlying assumptions poses several risks.
We discuss implications due to neglecting: real user validation, target user representation, other explanation goals, and human-AI collaborative decisions.
First, using CoXAM is ultimately for hypothesis formation of how users may (mis)interpret XAI methods; to verify these results, evaluations with real users must be conducted.
Next, the reasoning strategies implemented in the cognitive models should be informed from formative user studies of representative target users, otherwise, the decision outcomes may be spurious and deviate from how the relevant users would perform.
Furthermore, evaluating XAI benefits solely with the two decision tasks we have implemented (forward and counterfactual simulation) omits other goals and desiderata for explanation, such as trust, satisfaction and robustness~\cite{wang2019designing, langer2021we, thomas2022reliance};
neglecting these risk unintended consequences, i.e., Goodhart's law~\cite{strathern1997improving}.

Finally, users can also use AI collaboratively rather than only for automated decisions. 
Although human-aligned XAI could lead to convergent thinking where both AI and human share the same decision flaws, our work on CoXAM works toward human-aligned cognitive modeling for human-augmenting XAI.
Instead of increasing the illusion of explanation for increased subjective satisfaction or perceived understanding~\cite{skirzynskiinterpretability, vasconcelos2023explanations}, iterating XAI methods with CoXAM aims to improve the objective performance of understanding correctness.
The goal is to improve actual understanding rather than make understanding easier.
Rather than aligning the XAI which could be induced with human biases~\cite{li2024utilizing}, CoXAM helps to account for fallacious reasoning such that XAI can be designed to mitigate them.
Thus, CoXAM could improve human-AI collaborative decisions as long as users verify XAI~\cite{buccinca2021trust}, though this may be cognitively demanding since they need to maintain contrasting mental models of themselves and the AI.

\subsection{Limitations of Ecological Validity}

The CoXAM controller is pre-trained on four datasets, which we assume provide inductive biases about strategy selection that participants may have acquired from real-world experience. However, this constitutes a limited corpus and may embed priors specific to those datasets. A more comprehensive training regime using a broader and more diverse dataset collection would yield a more faithful representation of human experience.
Furthermore, our experimental design fixed the counterfactual simulation session after the forward simulation session. This ordering ensured that participants had an adequate forward-model understanding before attempting counterfactual reasoning, but it may have reduced ecological validity by introducing potential order effects. Notably, we observed that memory limitations did not influence counterfactual performance under this setup.

Finally, since we focused on semantic understanding, we chose assessable explanation schemas (decision trees and linear regression), and low-dimensional domains with understandable contexts (e.g, mushrooms).
This facilitated evaluations with lay participants, and is in line with user studies of XAI (e.g., \cite{poursabzi2021manipulating, abdul2020cogam, lai2022human, lim2009and, buccinca2021trust, bo2024incremental, ma2023should, lakkaraju2016interpretable, rong2023towards}).
Future work can investigate modeling user understanding for more specialized, complex, high-dimensional domains (e.g., healthcare~\cite{lim2025diagrammatization, panigutti2023co, van2025human}, and high-dimensional visual-analytic explanations~\cite{lyu2024if, kahng2017cti, kahng2017cti, boggust2022shared}).

\section{Conclusion}
We have presented CoXAM, a cognitive XAI-Adaptive Model, as a framework for debugging XAI interpretability. We conducted a formative elicitation user study which revealed insights into 7 user reasoning strategies for interpreting Rules, Weights and Hybrid explanations, which we then evaluated in a summative validation study. We implemented the XAI-adaptive CoXAM cognitive model based on the computational rationality to choose among reasoning strategies based on the trade-off between utility and reasoning time. CoXAM demonstrated its superior alignment with human decision-making compared to baseline machine learning proxy models. CoXAM successfully replicated and provided cognitive explanations for key findings: that counterfactual tasks are inherently harder, that decision tree rules are harder to recall than weight-based linear factors, and that the optimal XAI explanation type depends heavily on the application data context. Ultimately, CoXAM provides an interpretable cognitive basis for accelerating the debugging and benchmarking of disparate XAI techniques.

\section*{GenAI Usage Disclosure}   
The authors used generative AI tools (ChatGPT, Gemini) to assist with language refinement and clarity.
The authors reviewed and edited all AI-generated content and take full responsibility for the final version of the manuscript.

\begin{acks}
We thank Emily Chen, Li Zhaobin and Kevin Chang for their insightful inputs during the exploratory stage of this work.
This research is supported by the National Research Foundation, Singapore and Infocomm Media Development Authority under its Trust Tech Funding Initiative (Award No. DTC-RGC-09).
Any opinions, findings and conclusions or recommendations expressed in this material are those of the author(s) and do not reflect the views of National Research Foundation, Singapore and Infocomm Media Development Authority.
\end{acks}

\bibliographystyle{ACM-Reference-Format}
\bibliography{manuscript}


\appendix
\onecolumn
\section{Appendix}
\setcounter{figure}{0}
\renewcommand{\thefigure}{A\arabic{figure}}
\subsection{Technical modeling details}
In this section we detail the strategy update method for the feature attribution strategy as well as the hyperparameters of Reinforcement Learning controller in CoXAM.

\subsubsection{Update method for the Feature Attribution strategy}
\label{sec:feature-attribution-update}

After each trial, the Feature Attribution strategy updates its internal
Gaussian beliefs over each coefficient using the observed AI prediction
$\hat{y} \in \{0,1\}$ and the model’s predicted probability $p = P(y=\hat{y} | e)$. Each feature
parameter is represented as a Gaussian distribution
$\omega_{\rho} \sim \mathcal{N}(\mu_{\rho}, \sigma_{\rho}^2)$, where $\rho$
indexes the feature dimension.

We perform an incremental Bayesian update following a Laplace approximation
for logistic regression with a Gaussian prior
\cite{bishop2006pattern,murphy2012machine,mackay1992practical}. For each
feature $\rho$ with input value $x_{\rho}$, we compute the curvature term
\[
\chi = p(1 - p),
\]
and update the posterior precision, variance, and mean as:
\begin{align}
\lambda_{\rho}'   &= \frac{1}{\sigma_{\rho}^2} + \chi x_{\rho}^2, \\
\sigma_{\rho}'^{2}  &= \frac{1}{\lambda_{\rho}'}, \\
\mu_{\rho}'       &= \mu_{\rho} + \sigma_{\rho}'^{2} x_{\rho} (\hat{y} - p).
\end{align}
where $\lambda_{\rho}'$ is the updated precision, $\sigma_{\rho}'^{2}$ is the updated variance and $\mu_{\rho}'$ is the updated mean.

These updates refine each feature’s mean and variance based on the discrepancy
between the predicted probability $p$ and the observed AI output $\hat{y}$,
while weighting the change by both attribute magnitude $x_{\rho}$ and current
uncertainty $\sigma_{\rho}^2$.

\subsubsection{Reinforcement Learning Training}
\label{sec:ppo-params}
We detail the hyperparameters for the PPO algorithm that was trained using stable baselines 3:
\begin{table}[!htbp]
\centering
\caption{Hyperparameters used for counterfactual RL training (PPO).}
\label{tab:cf-rl-hparams}
\begin{tabular}{ll}
\toprule
\textbf{Setting} & \textbf{Value} \\
\midrule
Algorithm & PPO \\
Policy & MlpPolicy \\
Policy activation & Tanh \\
Policy network (actor) & [64, 64] \\
Value network (critic) & [64, 64] \\
Learning rate & $3\times 10^{-4}$ \\
Entropy coefficient (ent\_coef) & 0.01 \\
Discount factor ($\gamma$) & 0.8 \\
Rollout steps per env (n\_steps) & 1024 \\
Device & cpu \\
Verbose & 0 \\
\midrule
Total timesteps & 400{,}000 \\
Number of vector envs ($n\_envs$) & 8 (SubprocVecEnv) \\
Eval frequency & 1250 steps \\
\midrule
Env: instances per episode & 40 \\
Env: max attributes & 6 \\
\bottomrule
\end{tabular}
\end{table}

\onecolumn
\subsection{Full summative study results}

\begin{figure}[h]
    \centering
    \includegraphics[width=12cm]{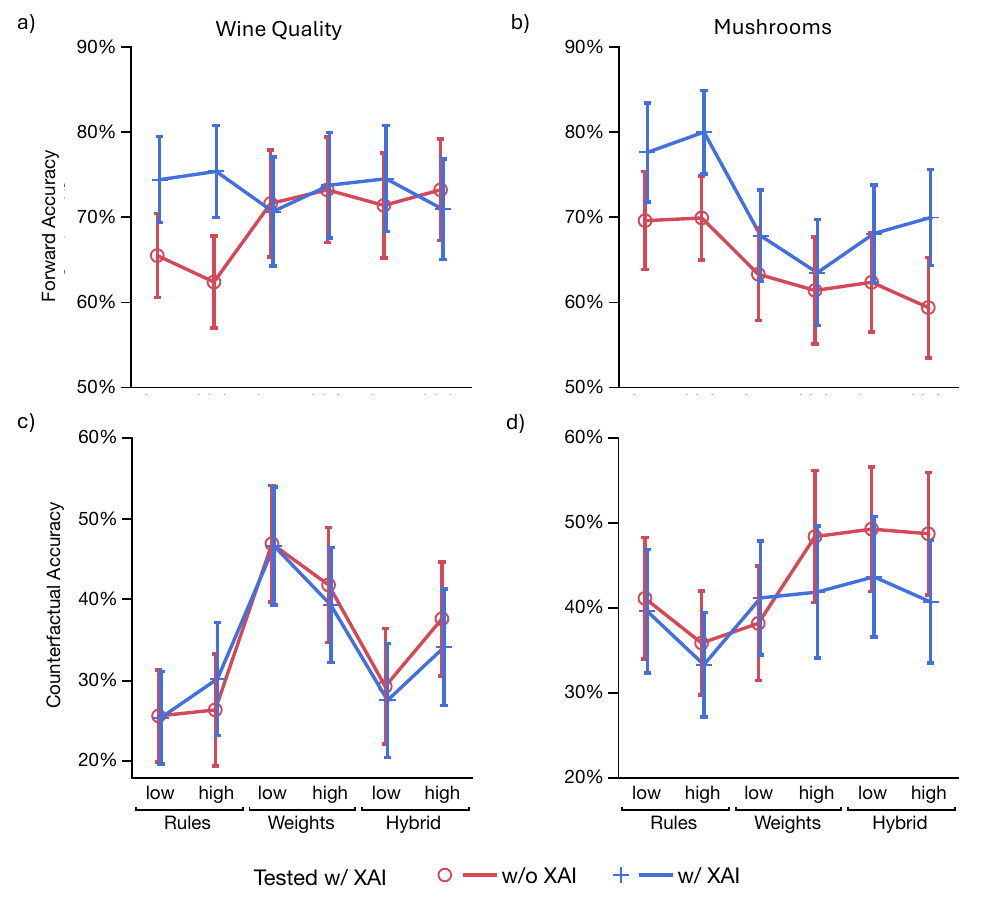}
    \caption{
        Forward and Counterfactual Simulation Accuracy, measured with respect to AI outputs across all conditions, and dataset application scenarios.
    }
    \Description{Shows four line plots comparing forward and counterfactual simulation accuracy for the Wine Quality and Mushrooms datasets. The top row shows forward accuracy and the bottom row shows counterfactual accuracy. The left column corresponds to Wine Quality and the right column to Mushrooms. Results are shown for three explanation types (Rules, Weights, and Hybrid), each under low and high conditions. Red circle markers indicate performance without XAI and blue plus markers indicate performance with XAI. Error bars show variability. Overall, forward accuracy is higher than counterfactual accuracy, and the effect of XAI varies across explanation types, conditions, and datasets.}
    \label{fig:forward-counterfactual-acc}
\end{figure}
\subsection{Modeling Results}

\begin{figure}[h]
    \centering
    \includegraphics[width=12cm]{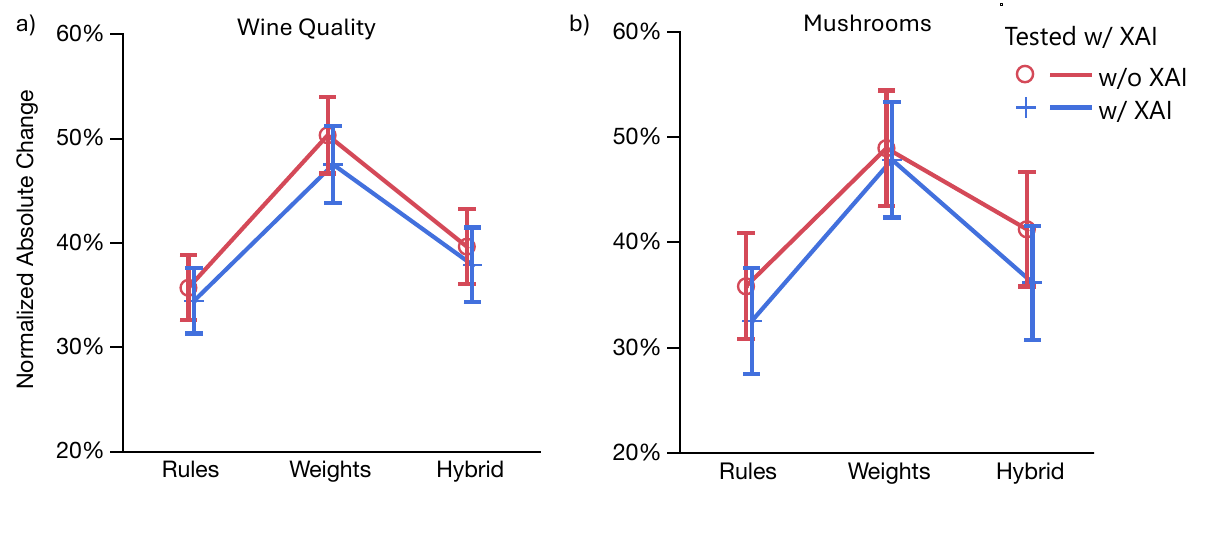}
    \caption{
        Normalized absolute change in attribute values across different XAI Schemas, datasets and with or without XAI trials.
    }
    \Description{Figure A2 shows two line plots of normalized absolute change in attribute values for the Wine Quality dataset (left) and the Mushrooms dataset (right). The x-axis lists explanation types (Rules, Weights, Hybrid) and the y-axis shows normalized absolute change in percent. Red circle markers indicate trials without XAI and blue plus markers indicate trials with XAI, with error bars showing variability. Across both datasets, the Weights explanation shows the largest changes, followed by Hybrid, with Rules showing the smallest changes.}
    \label{fig:absolute-change}
\end{figure}

\subsubsection{Individual accuracies and timings of each strategy}
Table. \ref{tab:forward-performance} and Table. \ref{tab:counterfactual-performance} show the accuracies and timings of the individual strategies for forward and counterfactual tasks respectively.

\begin{table}[!htbp]
\centering
\caption{Forward simulation performance (mean $\pm$ SE) across reasoning strategies, XAI conditions, and datasets. Accuracy is reported in \%, and time in seconds.}
\label{tab:forward-performance}
\begin{tabular}{llccc}
\toprule
\textbf{Dataset} & \textbf{Tested w/ XAI} & \textbf{Strategy} & \textbf{Accuracy (\%)} & \textbf{Time (s)} \\
\midrule
\multirow{6}{*}{\textbf{Mushrooms}} 
& w/ XAI  & Attribute-Threshold-Attribute Traversal  & 74.6 ± 1.1 & 9.2 ± 0.1 \\
&          & Approximate Calculation       & 64.6 ± 1.4 & 11.0 ± 0.2 \\
&          & Feature Attribution         & 49.4 ± 5.3 & 11.2 ± 0.6 \\
\cmidrule(l){2-5}
& w/o XAI & Attribute-Threshold-Attribute Traversal   & 63.2 ± 1.2 & 8.9 ± 0.1 \\
&          & Approximate Calculation        & 60.9 ± 1.5 & 10.7 ± 0.2 \\
&          & Feature Attribution          & 56.1 ± 4.8 & 12.3 ± 0.6 \\
\midrule
\multirow{6}{*}{\textbf{Wine Quality}} 
& w/ XAI  & Attribute-Threshold-Attribute Traversal   & 72.1 ± 1.3 & 8.7 ± 0.1 \\
&          & Approximate Calculation        & 48.6 ± 5.5 & 11.4 ± 1.2 \\
&          & Feature Attribution          & 75.5 ± 1.9 & 11.2 ± 0.2 \\
\cmidrule(l){2-5}
& w/o XAI & Attribute-Threshold-Attribute Traversal   & 65.5 ± 1.3 & 8.3 ± 0.1 \\
&          & Approximate Calculation        & 49.1 ± 6.2 & 9.4 ± 1.2 \\
&          & Feature Attribution          & 73.5 ± 1.9 & 11.0 ± 0.2 \\
\bottomrule
\end{tabular}
\end{table}

\begin{table}[!htbp]
\centering
\caption{Counterfactual accuracy (mean $\pm$ SE, \%) across reasoning strategies, whether XAI is shown, and datasets. Accuracy is reported against both the AI and XAI.}
\label{tab:counterfactual-performance}
\begin{tabular}{llccc}
\toprule
\textbf{Dataset} & \textbf{Tested w/ XAI} & \textbf{Strategy} & \textbf{Accuracy vs. AI (\%)} & \textbf{Accuracy vs. XAI (\%)} \\
\midrule
\multirow{7}{*}{\textbf{Mushrooms}} 
& w/ XAI  & Availability Heuristic      & 32.7 ± 2.5 & 43.0 ± 2.6 \\
&          & Inverse Calculation         & \textbf{48.9 ± 3.0} & 69.5 ± 2.7 \\
&          & Inverse Feature Attribution & 44.9 ± 2.1 & \textbf{71.4 ± 1.9} \\
&          & Node Threshold Crossing     & 49.0 ± 1.3 & 69.4 ± 1.2 \\
\cmidrule(l){2-5}
& w/o XAI & Availability Heuristic       & 30.7 ± 1.2 & 43.4 ± 1.2 \\
&          & Inverse Feature Attribution & \textbf{45.8 ± 1.7} & \textbf{70.4 ± 1.6} \\
&          & Node Threshold Crossing     & 45.1 ± 2.0 & 68.3 ± 1.9 \\
\midrule
\multirow{7}{*}{\textbf{Wine Quality}} 
& w/ XAI  & Availability Heuristic       & 20.9 ± 2.3 & 31.4 ± 2.6 \\
&          & Inverse Calculation         & 47.3 ± 3.7 & 57.0 ± 3.6 \\
&          & Inverse Feature Attribution & \textbf{50.3 ± 2.5} & 55.8 ± 2.5 \\
&          & Node Threshold Crossing     & 33.2 ± 1.1 & \textbf{61.3 ± 1.2} \\
\cmidrule(l){2-5}
& w/o XAI & Availability Heuristic       & 24.6 ± 1.0 & 34.7 ± 1.2 \\
&          & Inverse Feature Attribution & \textbf{50.1 ± 2.3} & \textbf{55.3 ± 1.5} \\
&          & Node Threshold Crossing     & 29.3 ± 1.8 & 53.8 ± 2.0 \\
\bottomrule
\end{tabular}
\end{table}

\onecolumn
\subsection{Analysis of cognitive parameters}
\begin{figure}[h]
    \centering\includegraphics[width=15cm]{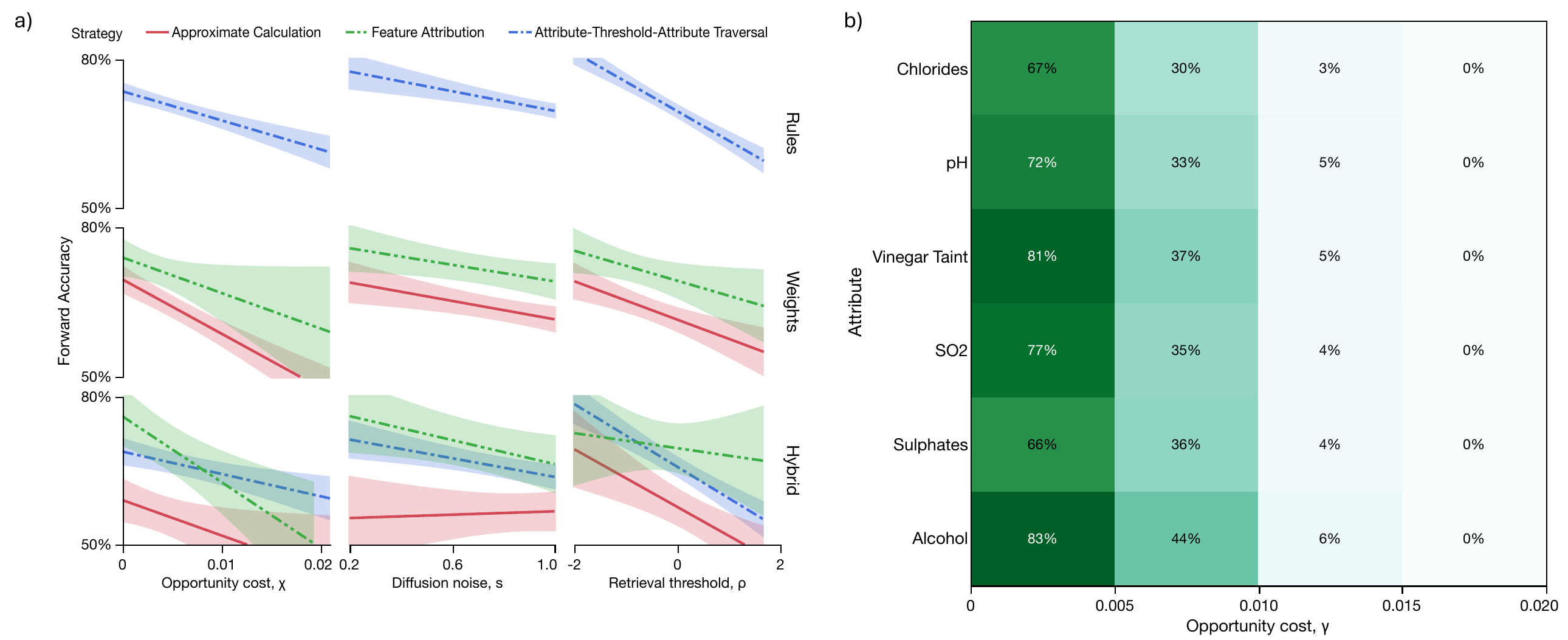}
    \caption{
        Variations in CoXAM performance against cognitive parameters. a) Strategy accuracy in the forward simulation task against cognitive parameters and specific trial strategies. b) Attribute selection, $\mathcal{X}$ of the six attributes in Wine Quality dataset against opportunity cost, $\gamma$.
    }
    \Description{This figure illustrates how forward simulation performance varies with cognitive parameters and task strategies.
    Panel (a) plots human and CoXAM forward accuracy as a function of three fitted cognitive parameters—opportunity cost ($\gamma$), diffusion noise ($\nu$), and retrieval threshold ($\kappa$)—for each explanation type (Rules, Weights, Hybrid). Shaded regions indicate 95\% confidence intervals. Accuracy declines as cognitive costs increase, particularly under higher γ and ν, showing greater sensitivity for strategies requiring more computation.
    Panel (b) shows the distribution of attribute selections (\mathcal{X}) across the six Wine Quality attributes as a function of opportunity cost. Participants increasingly focused on fewer, more diagnostic attributes (e.g., Alcohol, Vinegar Taint) as $\gamma$ rose, reflecting selective attention under higher cognitive effort.
    }
    \label{fig:forward-against-cog-params}
\end{figure}

\begin{figure}[!htbp]
    \centering\includegraphics[width=9cm]{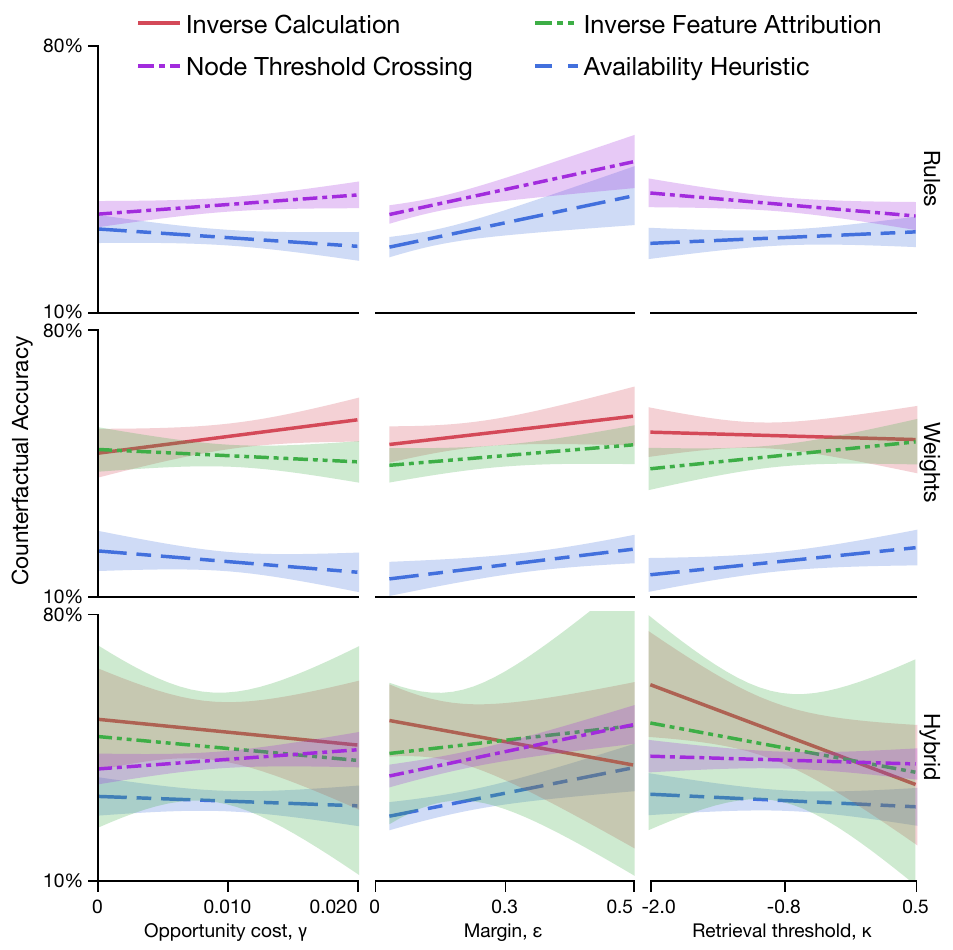}
    \caption{
       Variation in strategy accuracy in counterfactual task with varying cognitive parameters. Error areas are 95\% CI. 
    }
    \Description{This figure shows how counterfactual accuracy changes across reasoning strategies as cognitive parameters vary. Each row corresponds to one XAI Schema: Rules, Weights, and Hybrid. The three columns represent variations in opportunity cost gamma, margin epsilon, and retrieval threshold kappa. The four curves represent the strategies Inverse Calculation, Inverse Feature Attribution, Node Threshold Crossing, and Availability Heuristic. Shaded areas show 95 percent confidence intervals. Accuracy trends show that Inverse Feature Attribution and Node Threshold Crossing maintain stable performance under higher cognitive cost, while the Availability Heuristic declines more sharply with increasing gamma or kappa.
    }
    \label{fig:counterfactual-accuracy-vs-parameters}
\end{figure}

\subsection{Counterfactual Selection Probabilities of CoXAM}
\begin{figure}[h]
    \centering
\includegraphics[width=15cm]{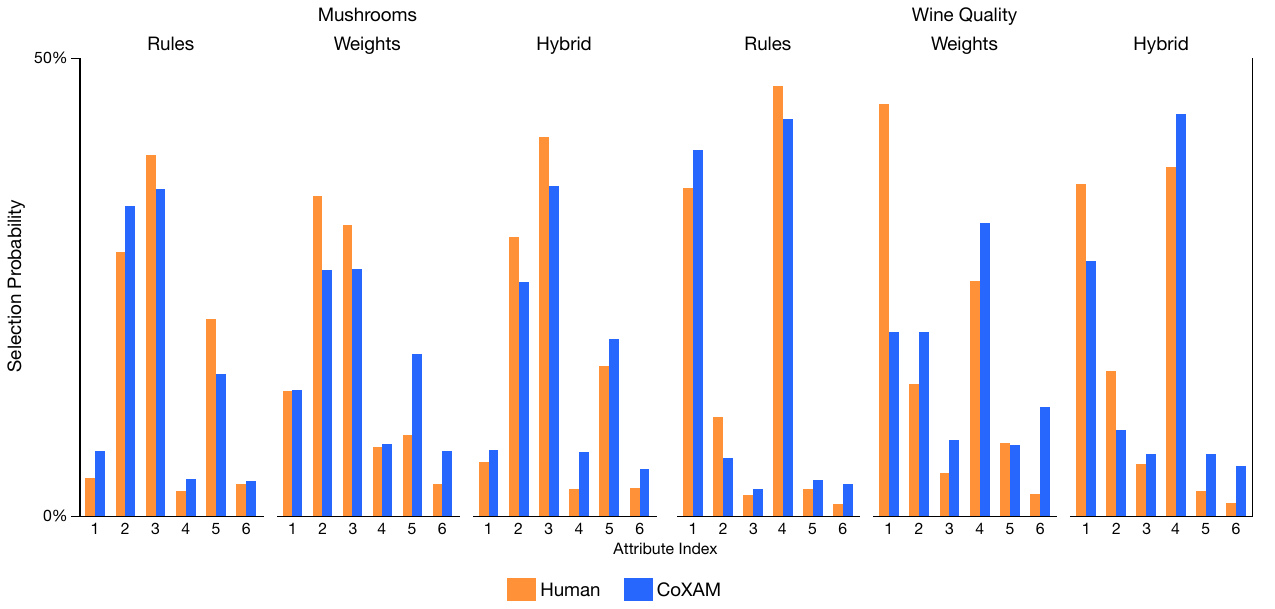}
    \caption{Attribute Selection by dataset and XAI Schemas (CoXAM Counterfactual Simulation Results).} 
    \Description{}
    \label{fig:attribute-select-counterfactual-distribution}
\end{figure}

\onecolumn
\subsection{Survey Flow: User Study and XAI UI}

\begin{figure}[h]
    \centering\includegraphics[width=12cm]{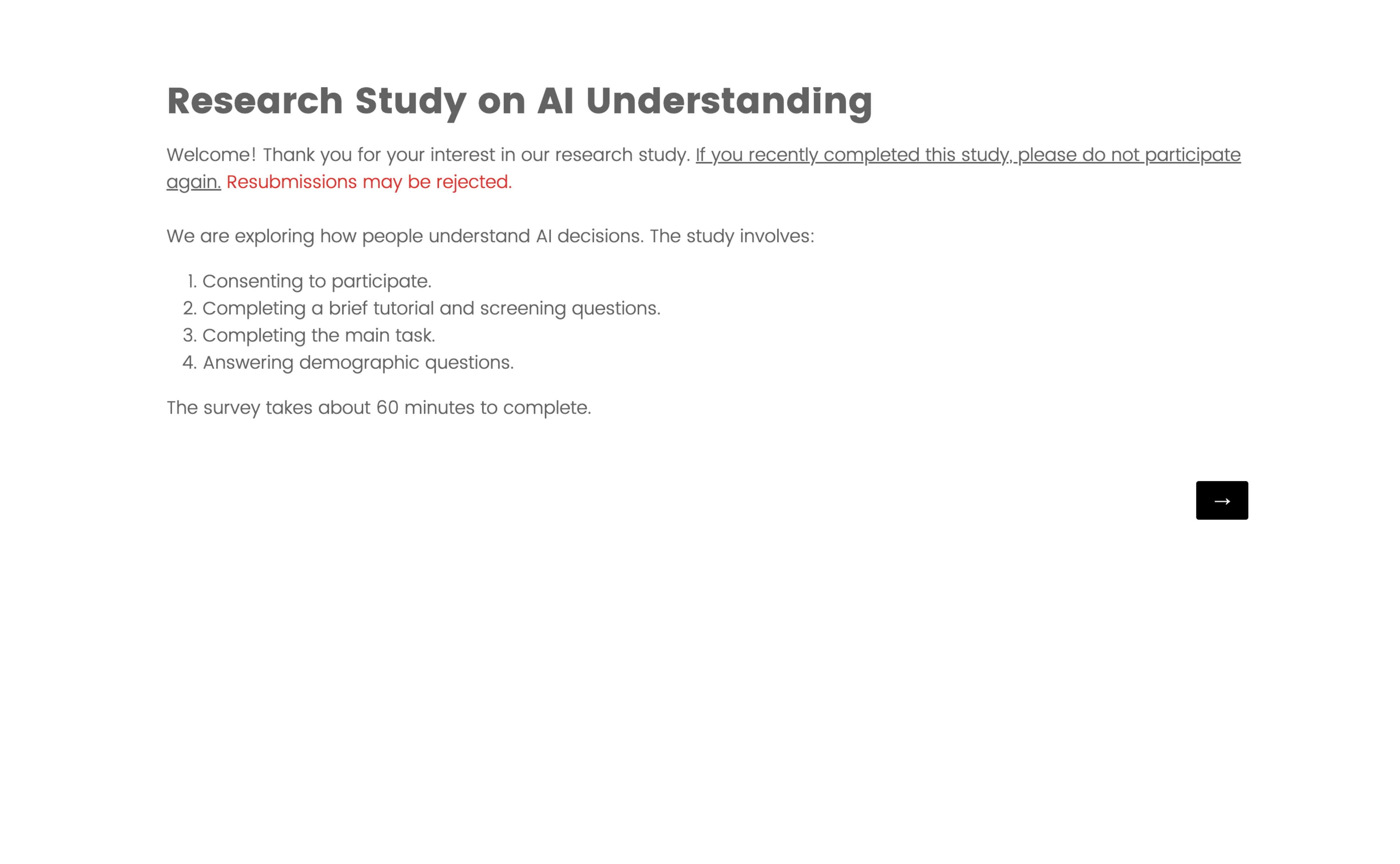}
    \caption{Introduction to User Study.}
    \Description{A study start page interface showing a title introducing a research study on AI understanding, a short description of the study purpose, a list of study steps, an estimated completion time, and a button to proceed.}
    \label{fig:UI-start}
\end{figure}

\begin{figure}[!htbp]
    \centering\includegraphics[width=12cm]{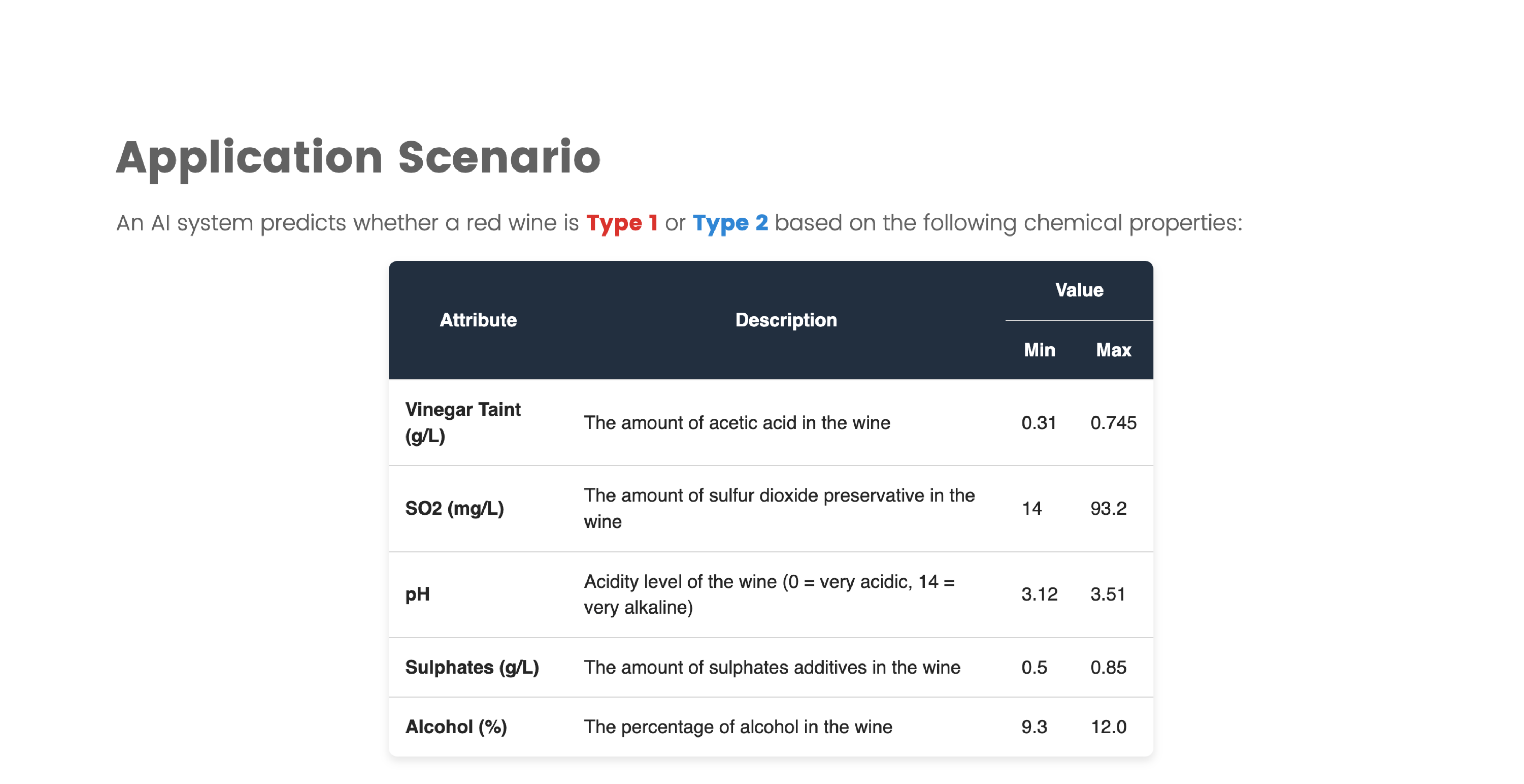}
    \caption{
        System UI of User Study and Tutorial.
    }
    \Description{An application scenario screen explaining that an AI predicts one of two wine types based on chemical attributes. A table lists attributes such as alcohol, sulphates, sulfur dioxide, vinegar taint, pH, and chlorides, along with their descriptions and minimum and maximum values.}
    \label{fig:UI-data}
\end{figure}

\begin{figure}[h]
    \centering\includegraphics[width=12cm]{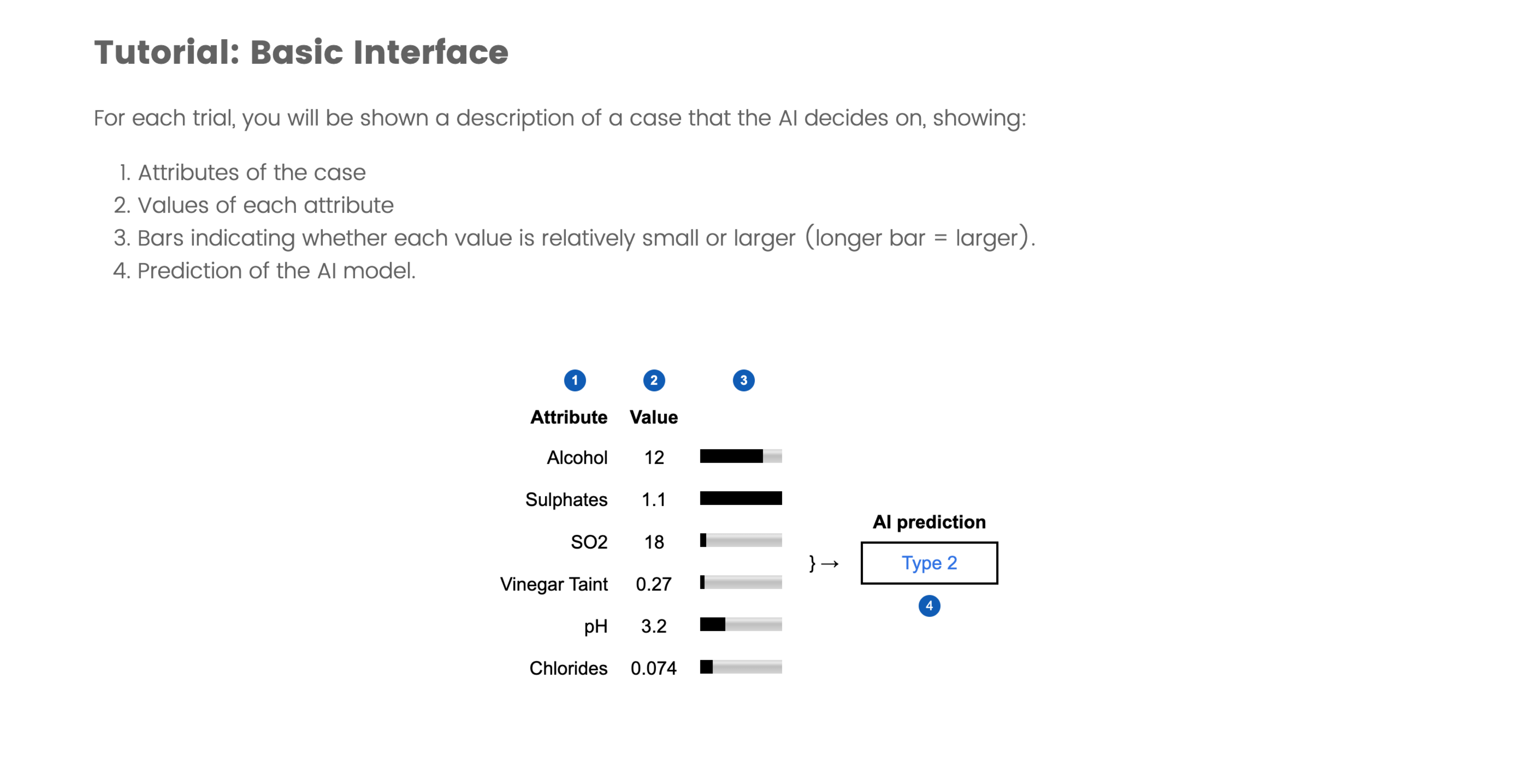}
    \caption{
        UI of Data Instance (Attribute Name, Value and Visual Slider).
    }
    \Description{A tutorial screen showing a single data instance. Each row lists an attribute name, its numeric value, and a horizontal bar indicating the relative magnitude of that value. An AI prediction label is shown to the right.}
    \label{fig:UI-basic}
\end{figure}

\begin{figure}[h]
    \centering\includegraphics[width=12cm]{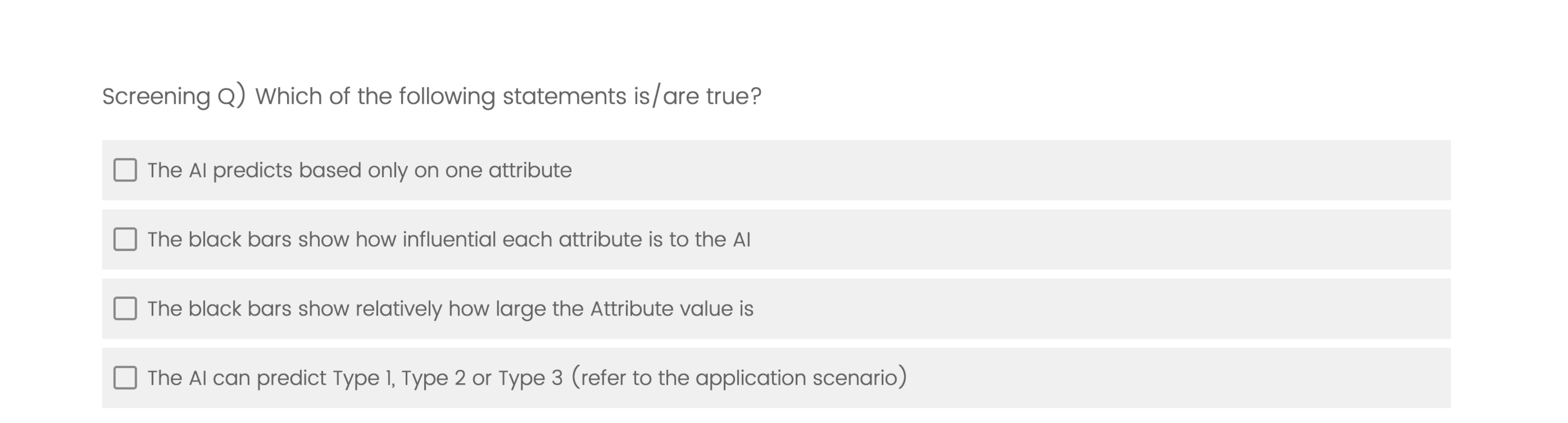}
    \caption{
        Screening Question (Data and Prediction Task).
    }
    \Description{A multiple-choice screening question asking participants to identify correct statements about how the data values, bars, and AI prediction should be interpreted.}
    \label{fig:UI-data-scr}
\end{figure}

\begin{figure}[h]
    \centering\includegraphics[width=12cm]{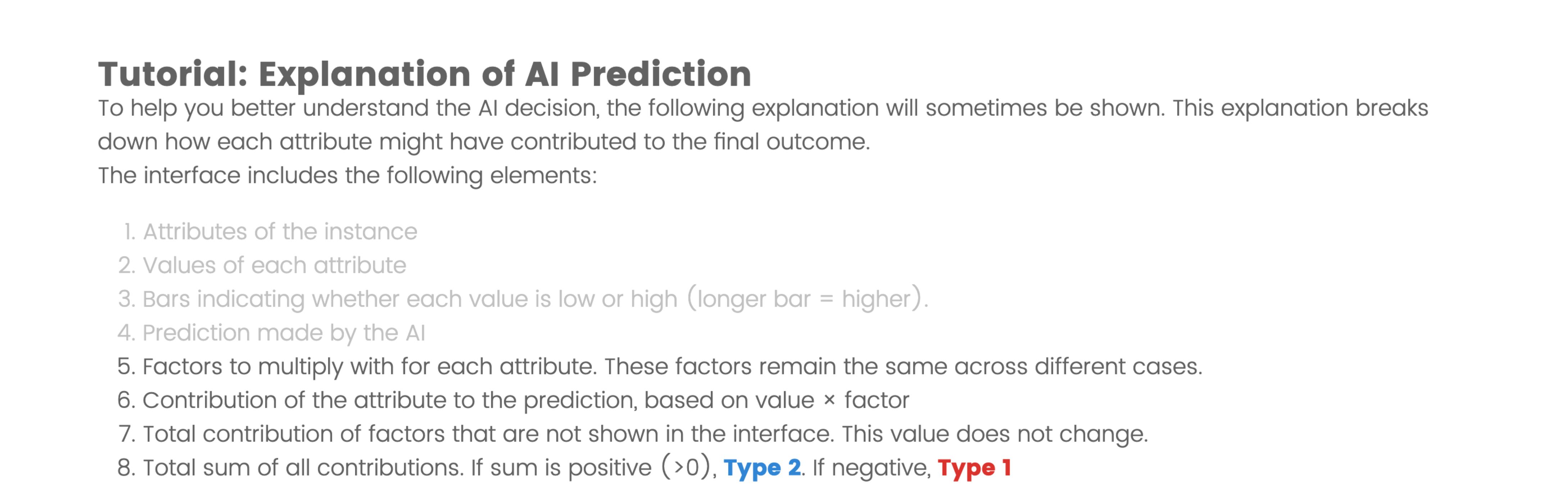}\\[1em]
    \includegraphics[width=12cm]{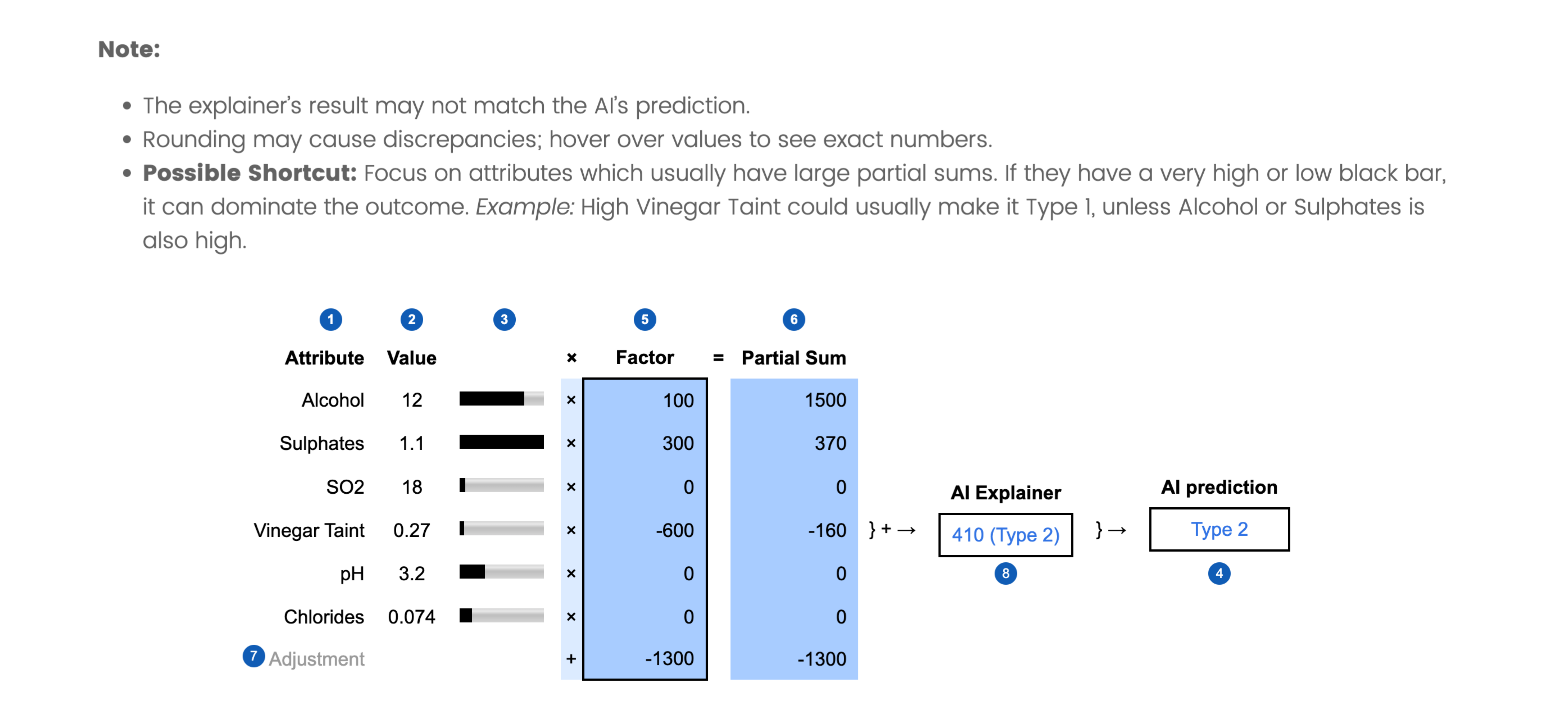}\\[1em]
    \includegraphics[width=12cm]{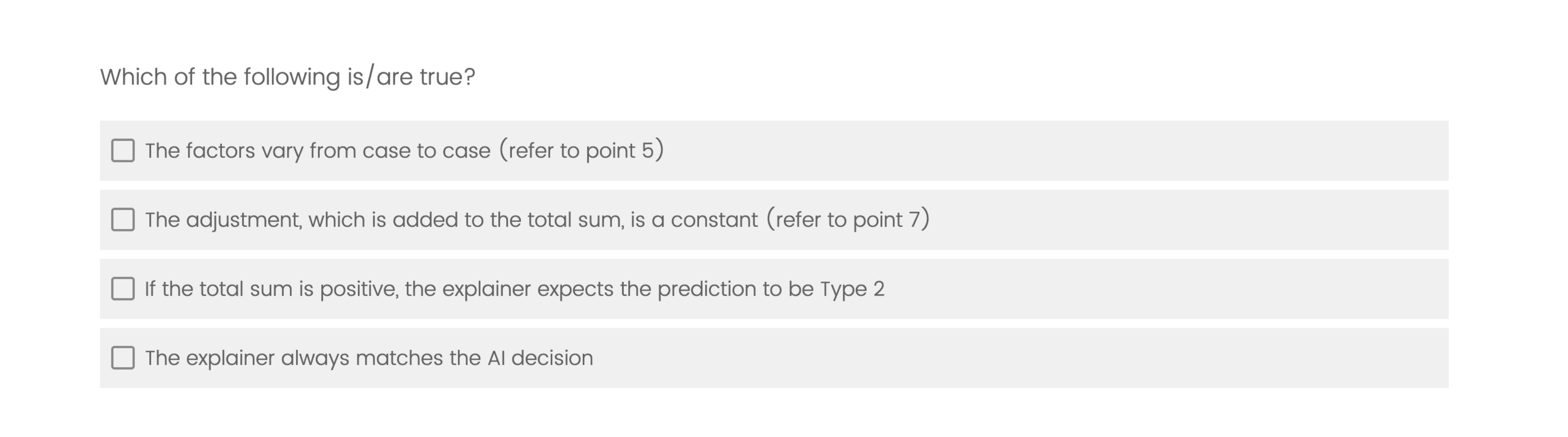}\\[1em]
    \caption{
        Tutorial for Weight-based XAI (top) and screening question (bottom).
    }
    \Description{A tutorial screen explaining a weight-based AI explanation. Attributes and values are shown alongside fixed numeric factors, partial sums for each attribute, and a total contribution. The summed value leads to an explained prediction, which may or may not match the AI’s prediction. Also, below, a multiple-choice screening question asking participants to identify correct statements about how the weight-based explanation works, including whether factors are constant and how the final prediction is determined.}
    \label{fig:UI-LR}
\end{figure}

\begin{figure}[h]
  \centering
  \includegraphics[width=12cm]{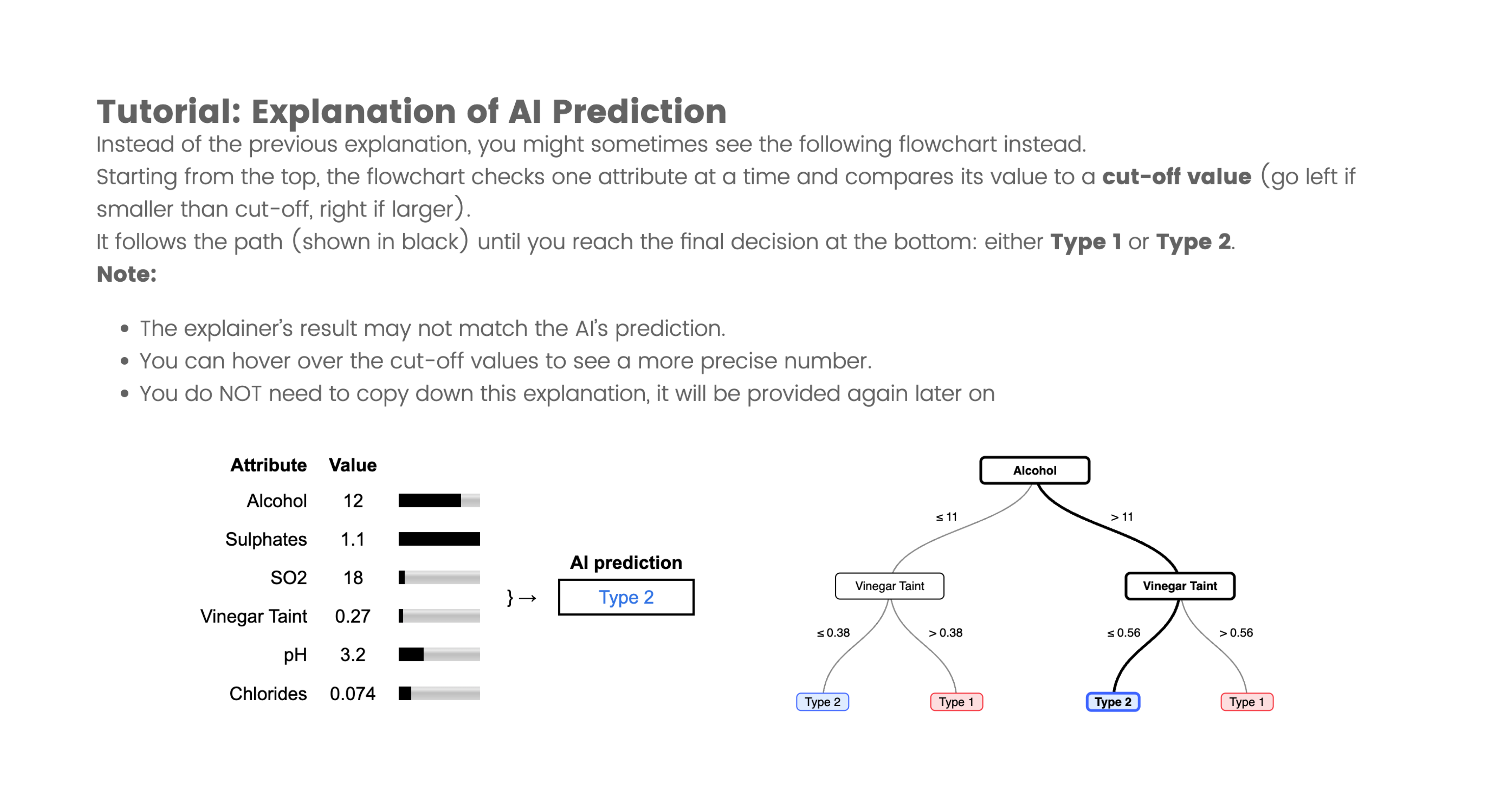}\\[1em]
   \includegraphics[width=12cm]{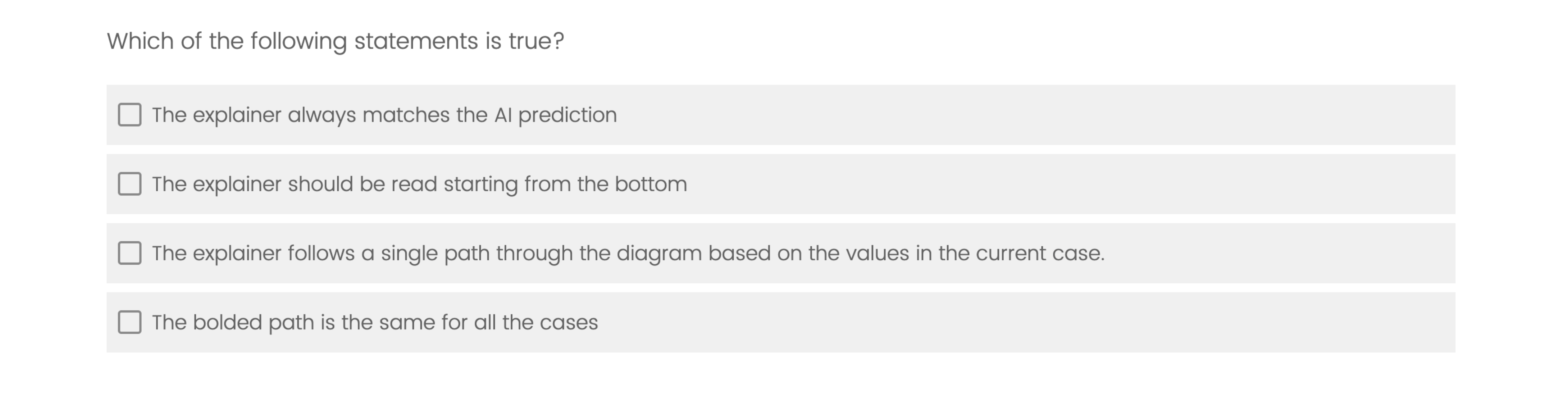}
  \caption{
    Tutorial for Rule-based XAI (top) and screening question (bottom).
  }
  \Description{A tutorial screen explaining a rule-based AI explanation using a decision tree. The tree checks attributes one at a time against thresholds, follows a single highlighted path, and ends at a predicted class. Also, below, a multiple-choice screening question asking participants to identify correct statements about how the rule-based explanation should be read and interpreted.}
  \label{fig:UI-DT}
\end{figure}

\begin{figure}[h]
    \centering
\includegraphics[width=12cm]{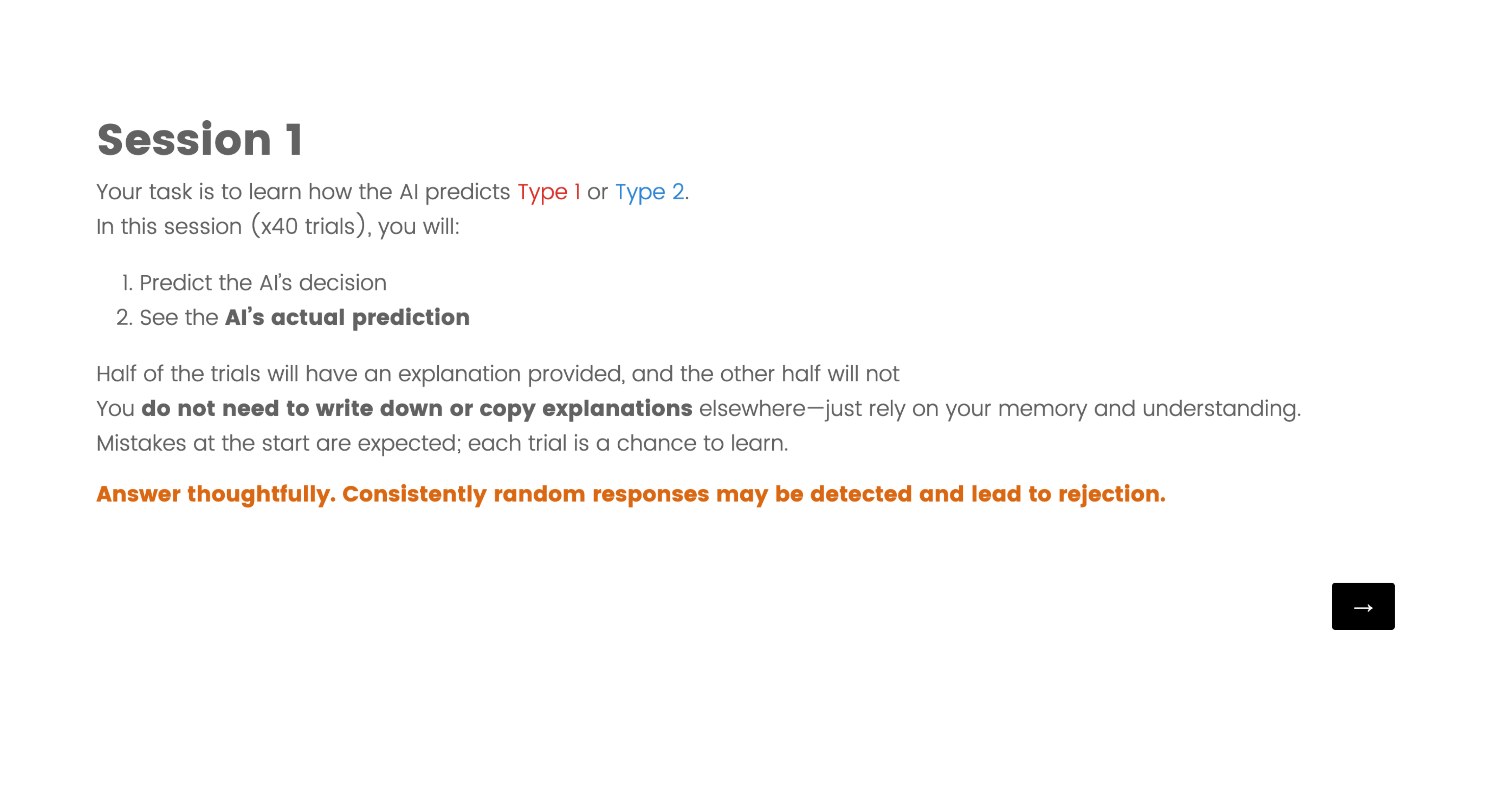}
    \caption{Start page of 40-trial forward simulation session.}
    \Description{A session introduction page explaining that participants will complete multiple trials where they learn how the AI predicts one of two classes. Instructions emphasize learning from feedback and answering thoughtfully.}
    \label{fig:UI-FR1}
\end{figure}

\begin{figure}[h]
    \centering
    \includegraphics[width=12cm]{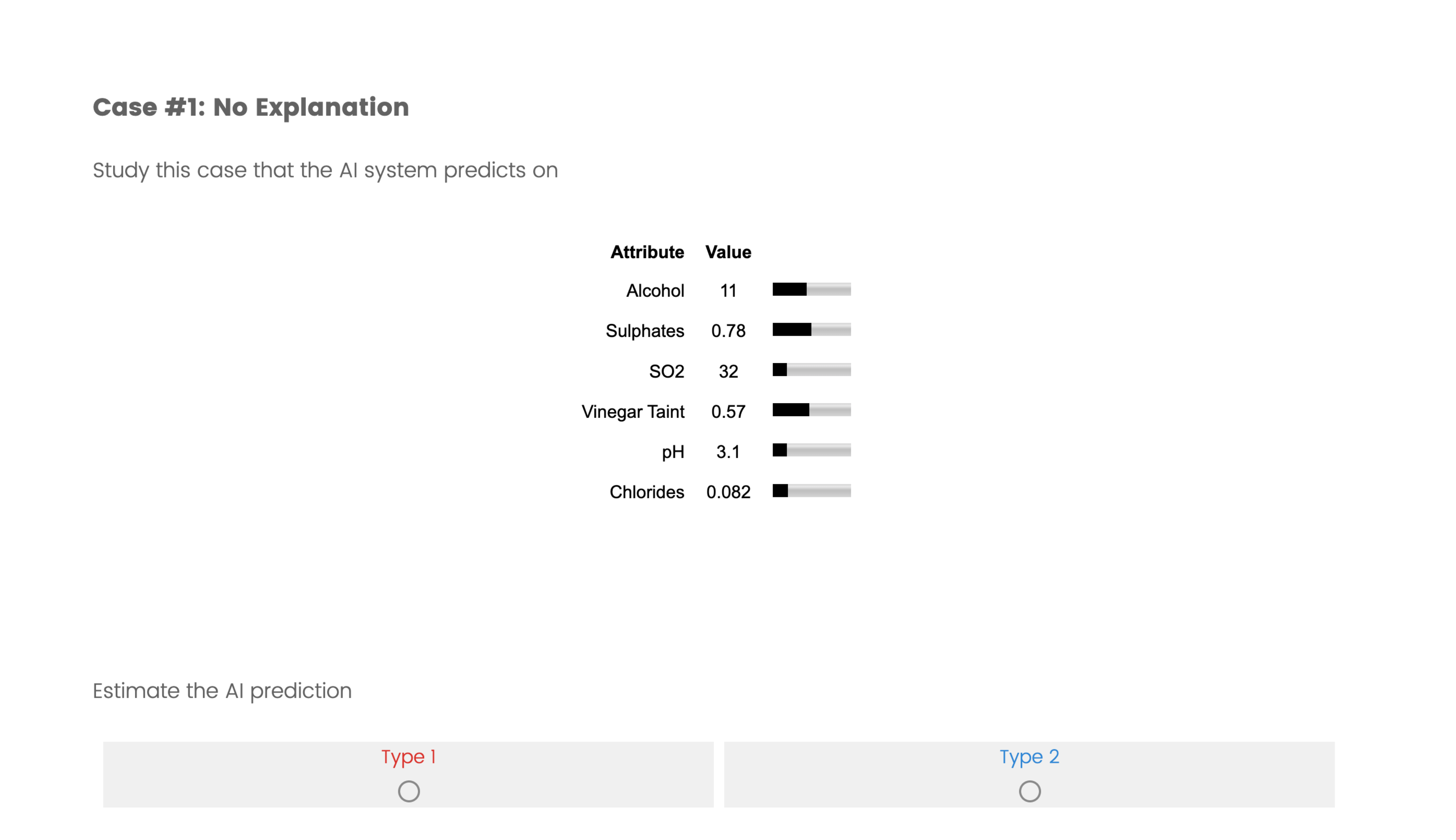}\\[1em]
    \caption{
        User predicts AI output without XAI.
    }
    \Description{A task screen without explanation where participants view attribute values with bars and must estimate the AI’s prediction by selecting one of two options.}
    \label{fig:UI-FR-NO}
\end{figure}

\begin{figure}[h]
    \centering
    \includegraphics[width=12cm]{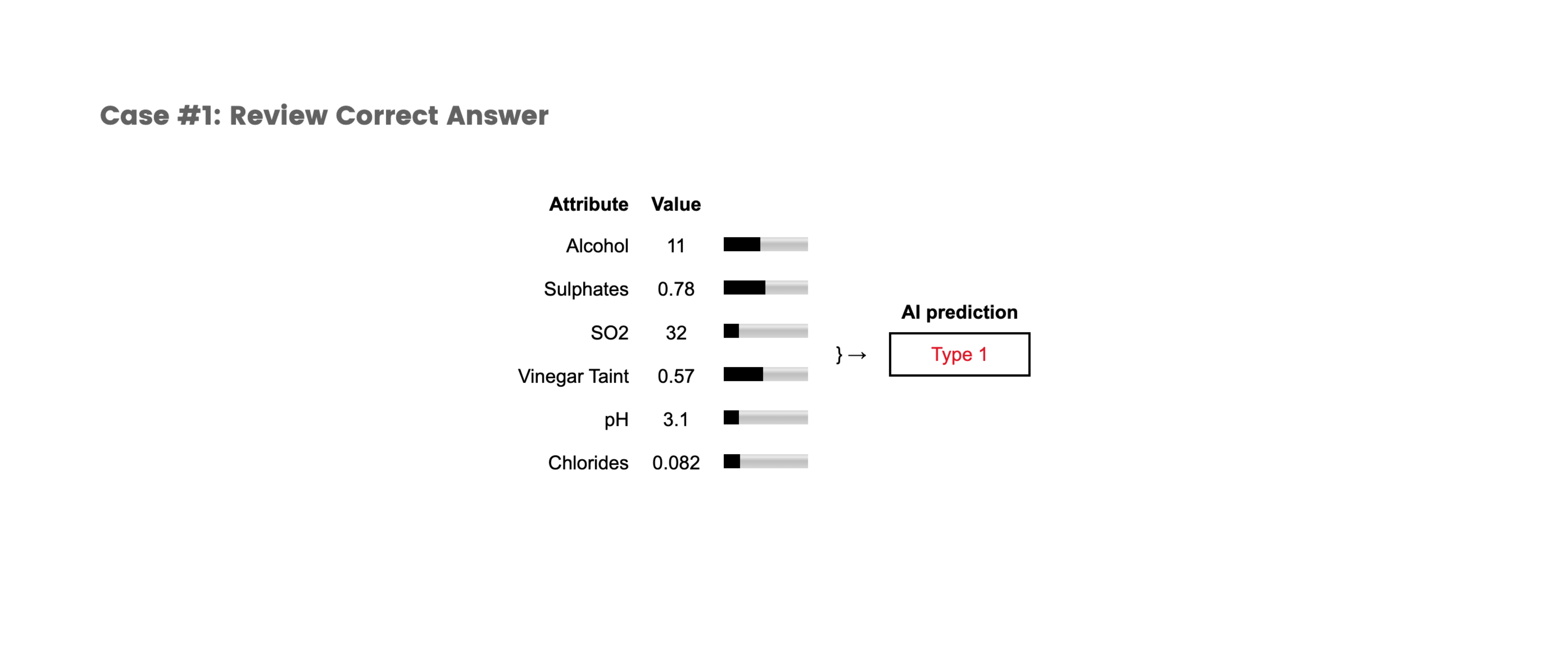}\\[1em]
    \caption{Feedback on AI output without XAI.}
    \Description{A feedback screen showing the correct AI prediction for the previous case, alongside the same attribute values and bars.}
    \label{fig:UI-FR-NO}
\end{figure}

\begin{figure}[h]
    \centering
    \includegraphics[width=12cm]{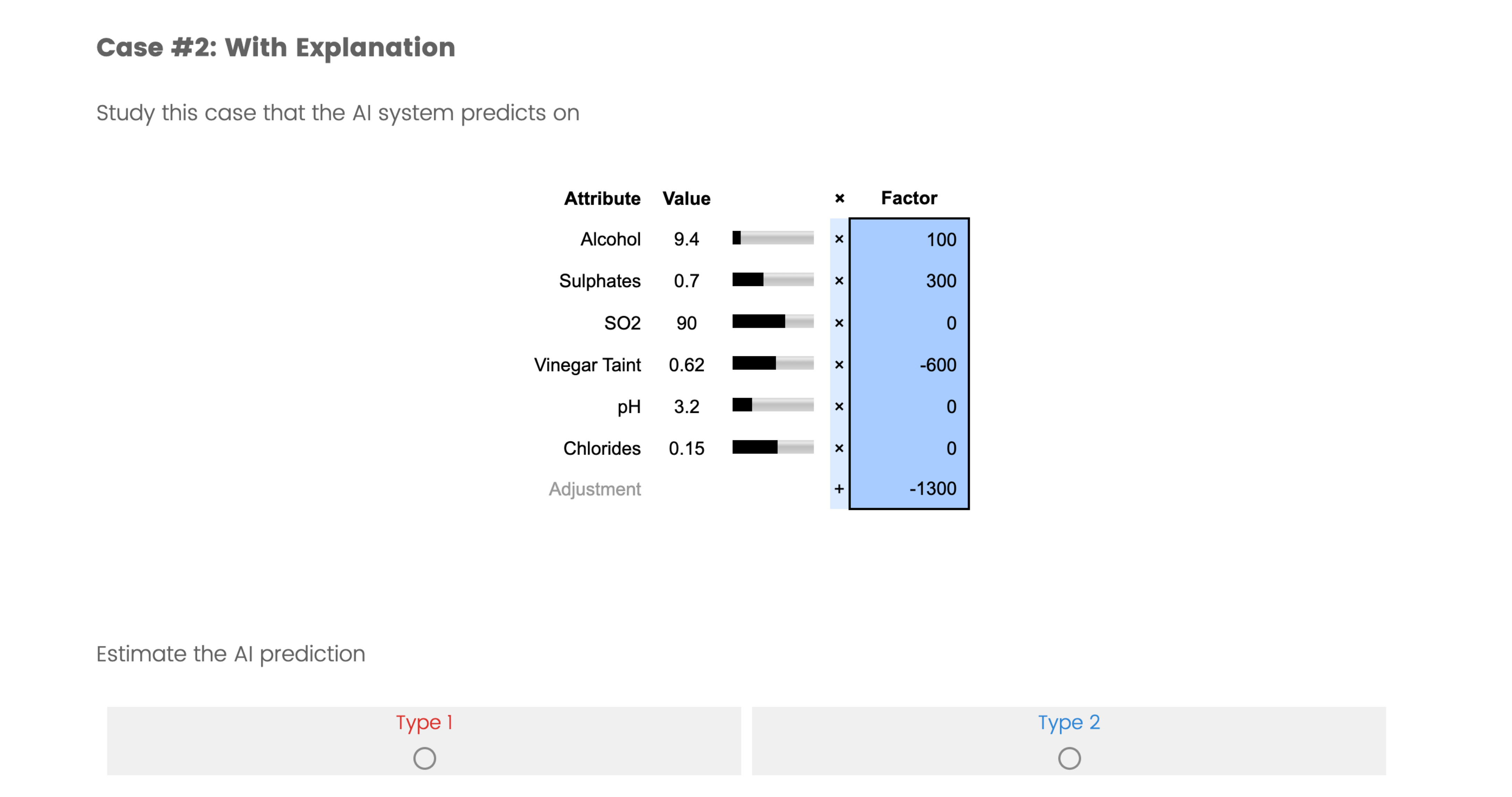}
    \caption{
        User predict AI output with the Weight-based XAI.
    }
    \Description{A task screen with a weight-based explanation where participants view attribute values, factors, and partial sums, and then estimate the AI’s prediction.}
    \label{fig:UI-LR-FR}
\end{figure}

\begin{figure}[h]
    \centering
    \includegraphics[width=12cm]{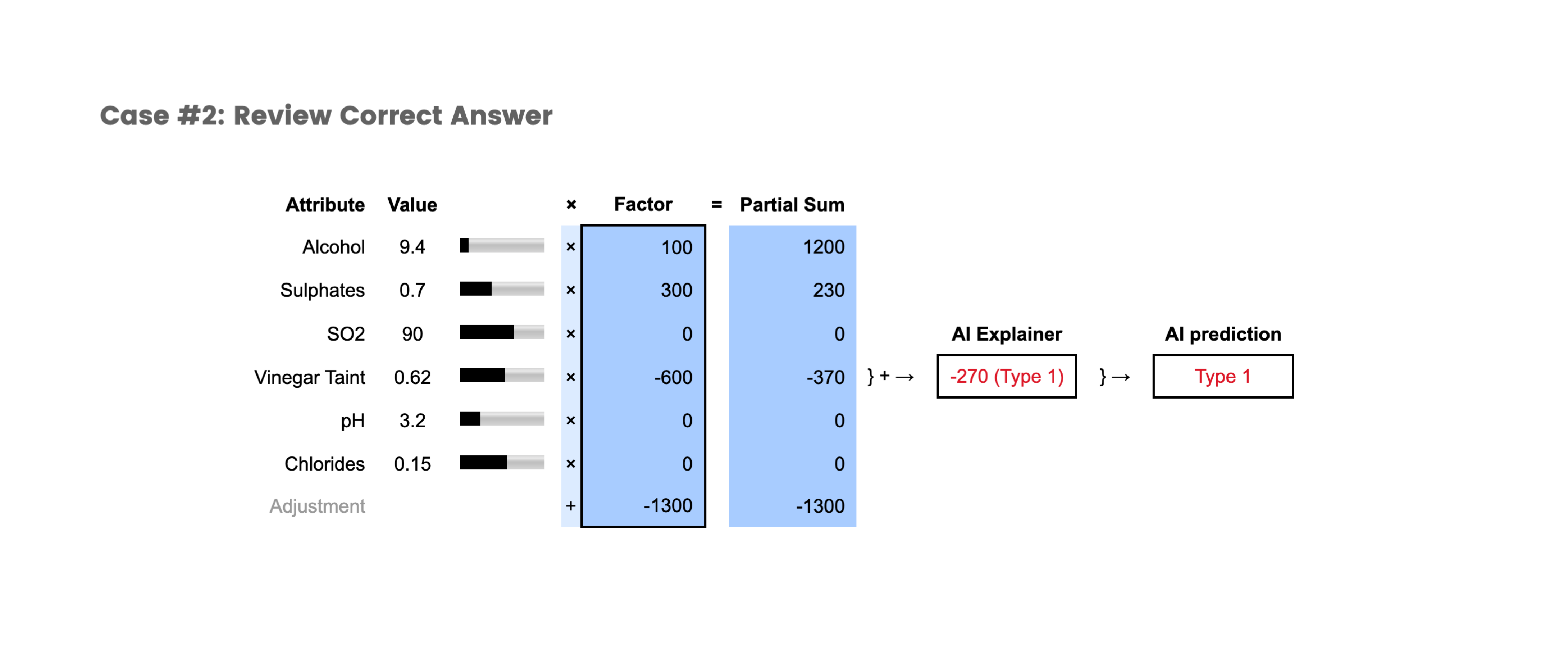}
    \caption{
        Feedback on AI output with the Weight-based XAI.
    }
    \Description{A feedback screen for the weight-based explanation showing the computed total contribution, the explained prediction, and the AI’s actual prediction.}
    \label{fig:UI-LR-FR-FD}
\end{figure}

\begin{figure}[h]
    \centering
    \includegraphics[width=12cm]{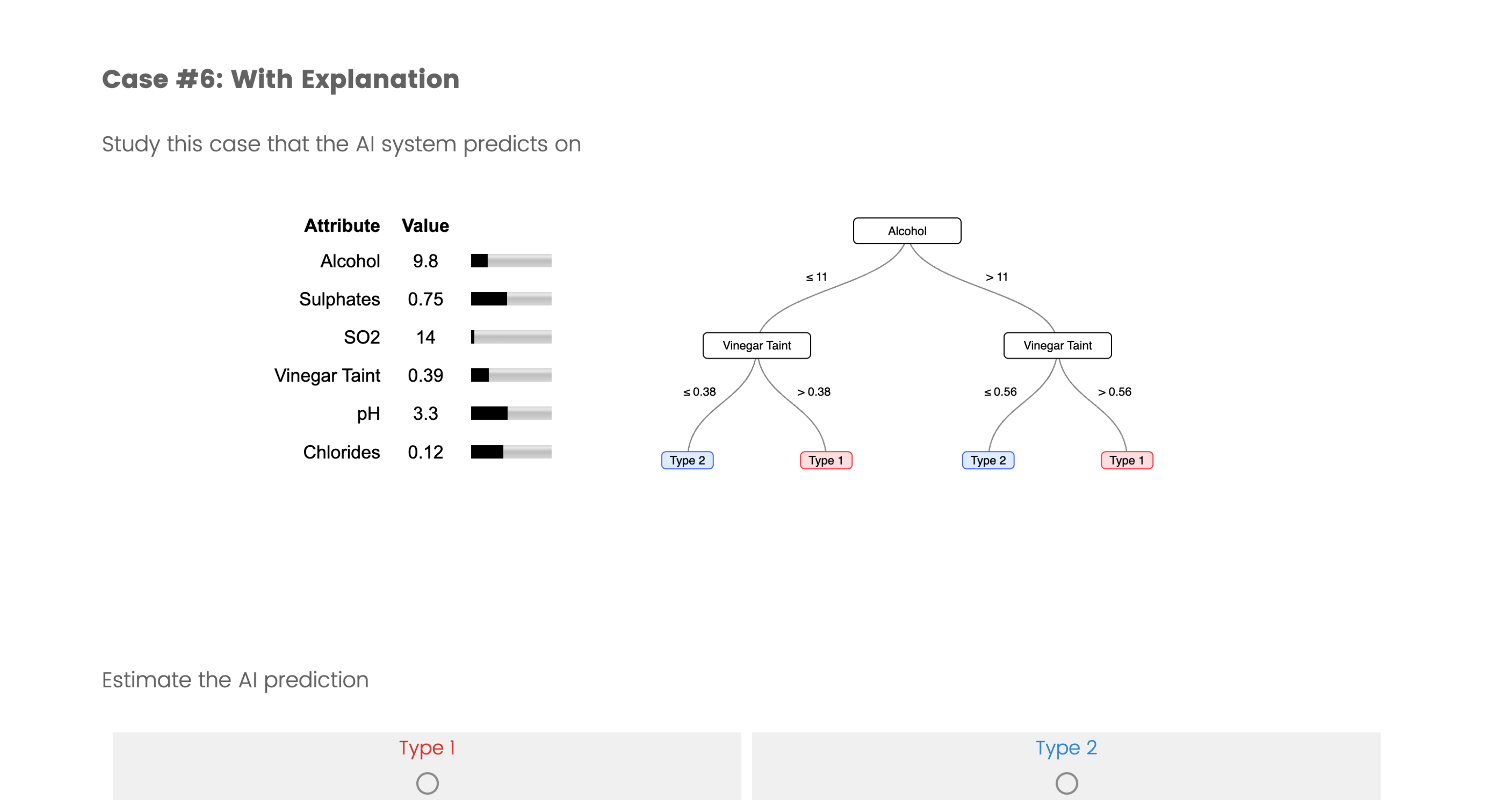}
    \caption{
        User predicts AI output with the Rule-based XAI.
    }
    \Description{A task screen with a rule-based explanation where participants see attribute values, a decision tree, and must estimate the AI’s prediction.}
    \label{fig:UI-DT-FR}
\end{figure}

\begin{figure}[h]
    \centering
    \includegraphics[width=12cm]{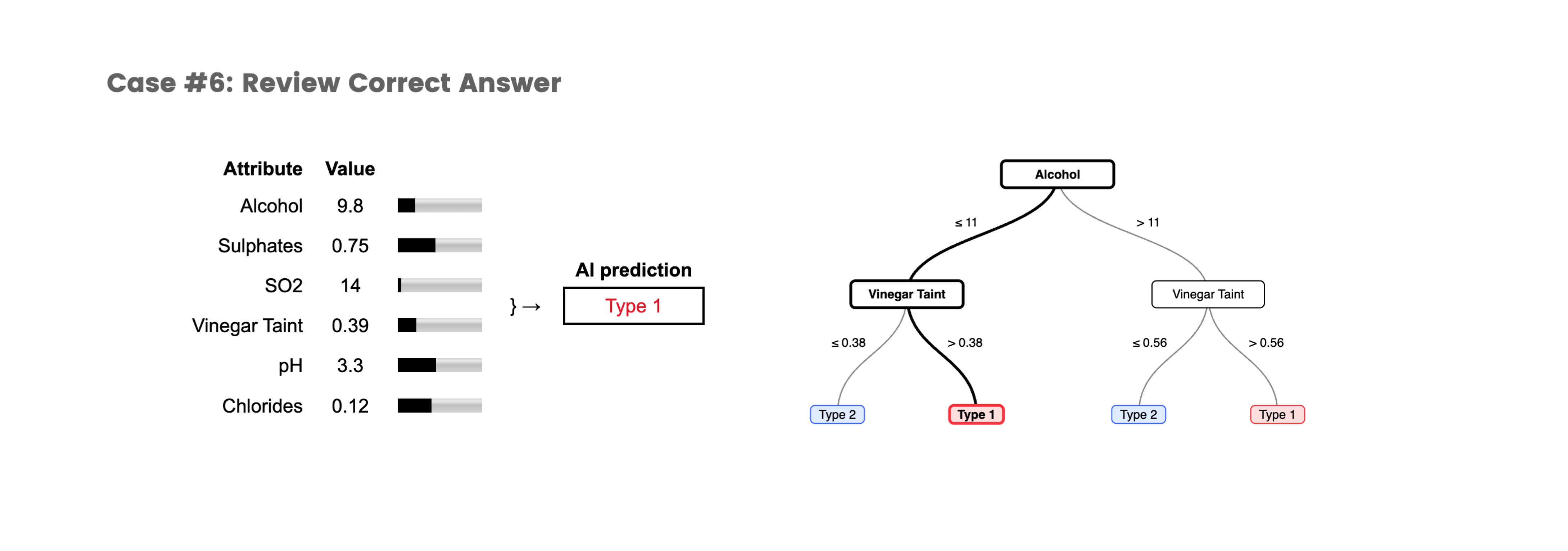}
    \caption{
        Feedback on AI output with the Rule-based XAI.
    }
    \Description{A feedback screen for the rule-based explanation showing the AI’s prediction and the path taken through the decision tree.}
    \label{fig:UI-DT-FR-FD}
\end{figure}

\newpage
\begin{figure}[h]
    \centering\includegraphics[width=12cm]{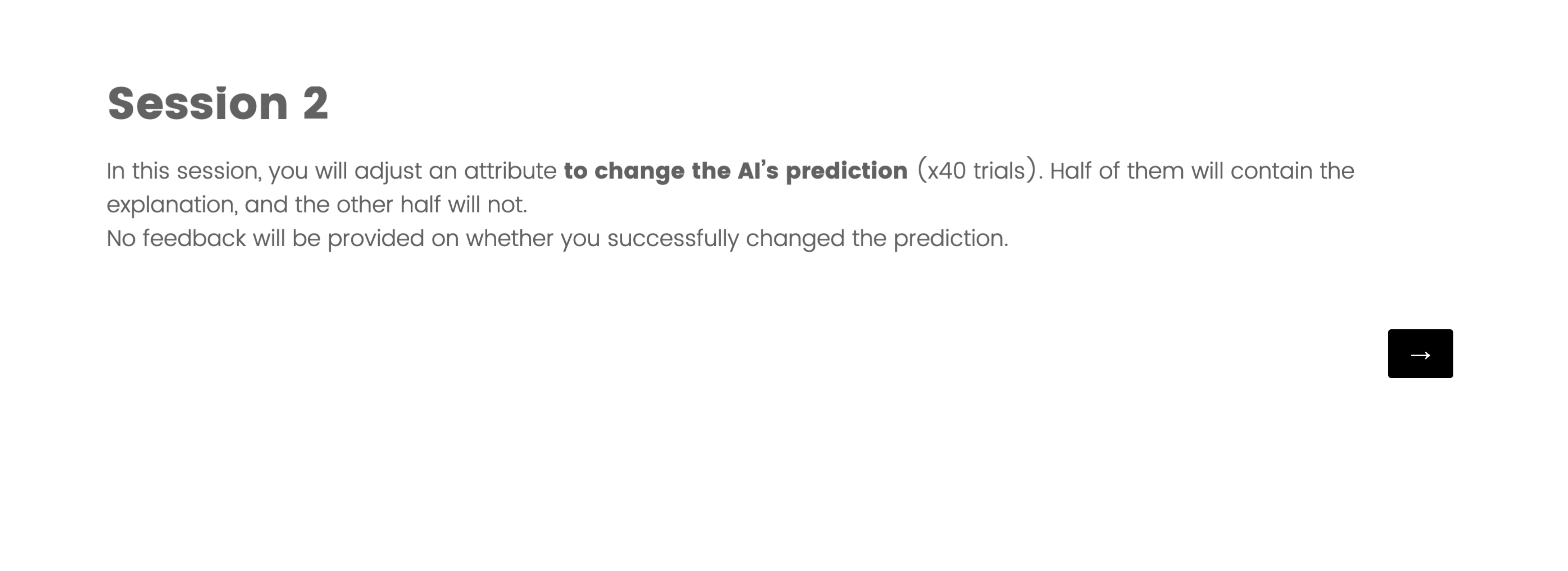}
    \caption{
        Start page of 40-trial counterfactual simulation session.
    }
    \Description{A session introduction page explaining a counterfactual task where participants must change exactly one attribute by as little as possible to flip the AI’s prediction, with no feedback provided on success.}
    \label{fig:UI-CF}
\end{figure}

\begin{figure}[h]
    \centering
    \includegraphics[width=12cm]{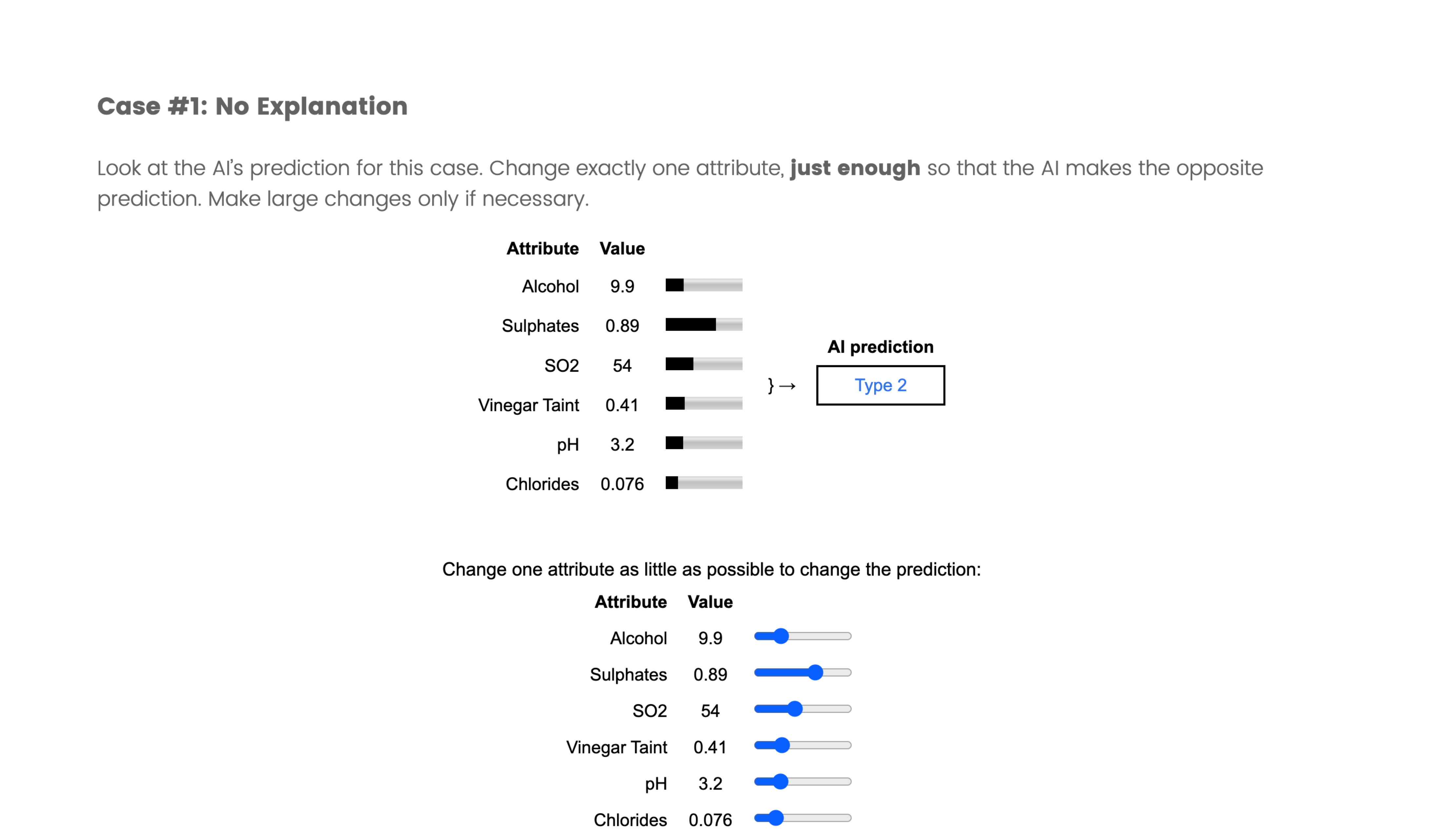}
    \caption{
        Change an attribute value without XAI.
    }
    \Description{A counterfactual task screen without explanation showing current attribute values, the AI’s prediction, and sliders allowing one attribute to be adjusted to change the prediction.}
    \label{fig:UI-CF}
\end{figure}

\begin{figure}[h]
    \centering
    \includegraphics[width=12cm]{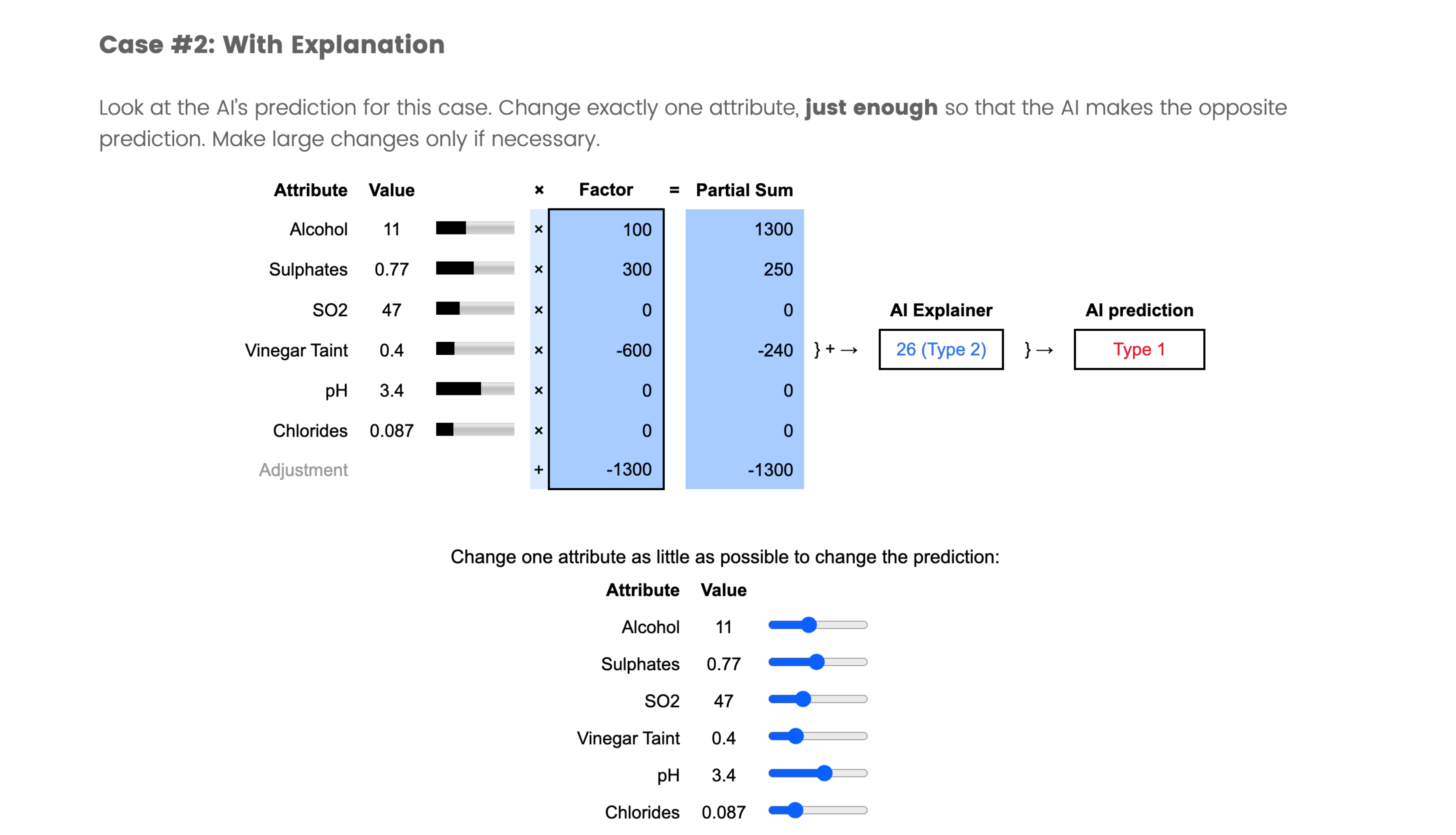}
    \caption{
        Change an attribute value with the Weight-based XAI (No AI feedback).
    }
    \Description{A counterfactual task screen with a weight-based explanation showing attribute values, factors, partial sums, and sliders for adjusting a single attribute, without feedback on whether the prediction changed.}
    \label{fig:UI-LR-CF}
\end{figure}

\begin{figure}[h]
    \centering
    \includegraphics[width=12cm]{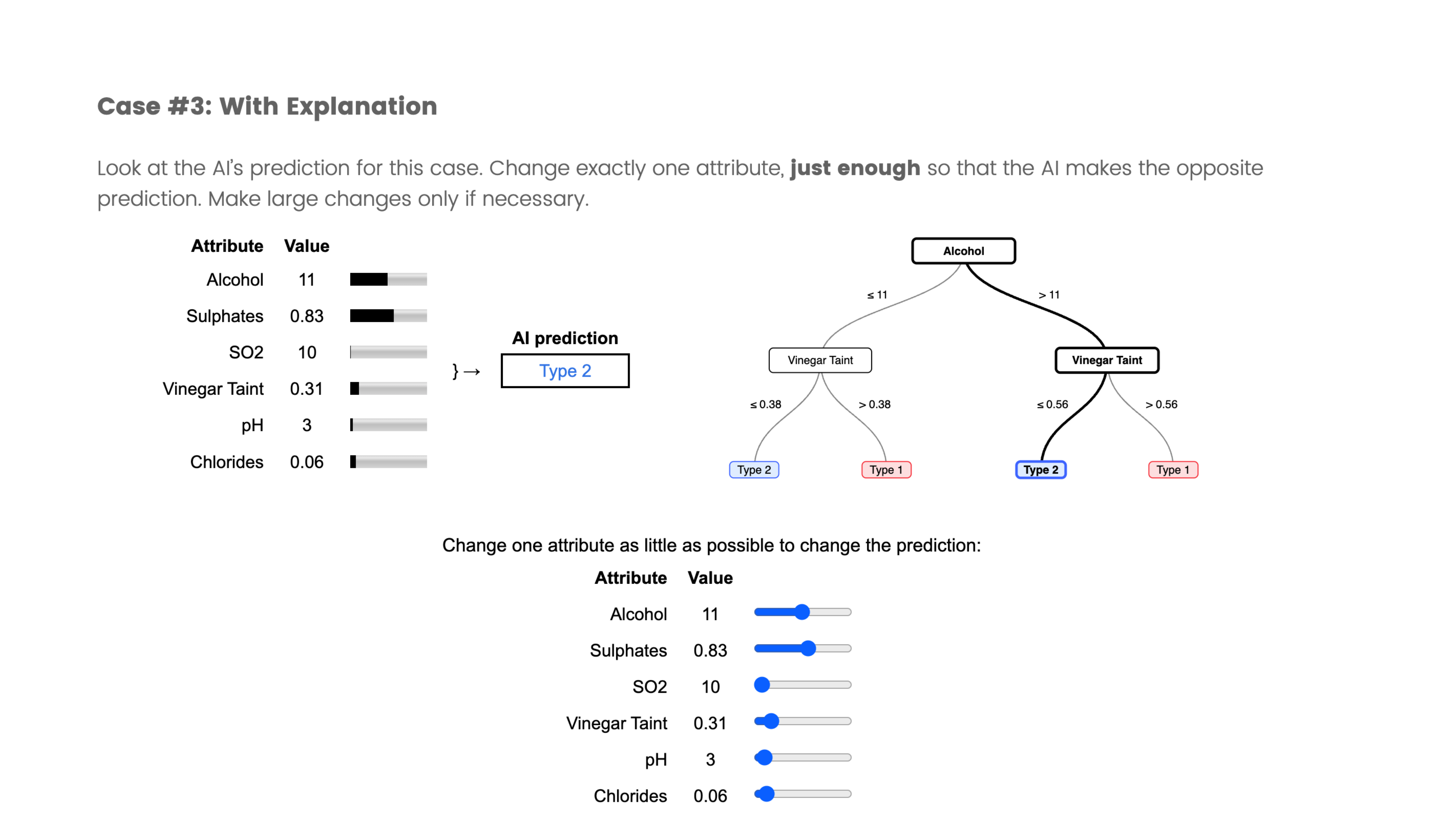}
    \caption{
        Change an attribute value with the Rule-based XAI (No AI feedback).
    }
    \Description{A counterfactual task screen with a rule-based explanation showing attribute values, a decision tree, and sliders for adjusting a single attribute, without feedback on whether the prediction changed.}
    \label{fig:UI-DT-CF}
\end{figure}

\end{document}